%% file: camera_ready.tex
\documentclass{article}

\usepackage{arxiv}

\usepackage[utf8]{inputenc} % allow utf-8 input
\usepackage[T1]{fontenc}    % use 8-bit T1 fonts
\usepackage{hyperref}       % hyperlinks
\usepackage{url}            % simple URL typesetting
\usepackage{booktabs}       % professional-quality tables
\usepackage{amsfonts}       % blackboard math symbols
\usepackage{nicefrac}       % compact symbols for 1/2, etc.
\usepackage{microtype}      % microtypography
\usepackage{lipsum}
\usepackage{graphicx}
\graphicspath{ {./images/} }

\usepackage[utf8]{inputenc} % allow utf-8 input
\usepackage[T1]{fontenc}    % use 8-bit T1 fonts
\usepackage{hyperref}       % hyperlinks
\usepackage{url}            % simple URL typesetting
\usepackage{booktabs}       % professional-quality tables
\usepackage{amsfonts}       % blackboard math symbols
\usepackage{nicefrac}       % compact symbols for 1/2, etc.
\usepackage{microtype}      % microtypography

\usepackage{hyperref}       % hyperlinks
\usepackage{url}            % simple URL typesetting
\usepackage{booktabs}       % professional-quality tables
\usepackage{amsfonts}       % blackboard math symbols
\usepackage{nicefrac}       % compact symbols for 1/2, etc.
\usepackage{microtype}      % microtypography
\usepackage{lipsum}
\usepackage{enumitem}
\usepackage{filecontents}
\usepackage{comment}
\usepackage{enumerate}
\usepackage{amsfonts}
\usepackage{amssymb}
\usepackage{verbatim}
\usepackage{amsthm}
\usepackage{setspace}
\usepackage{mathrsfs}
\usepackage{tikz}
\usepackage{soul}
\usetikzlibrary{arrows,shapes}
\newcommand\figref{Figure~\ref}
\usepackage{graphicx}
\usepackage{subcaption}

\usepackage{etoolbox}

\usepackage{todonotes}
\usepackage{bm} 
\usepackage{float}
\usepackage{graphicx}
\usepackage{tikz}
\usetikzlibrary{shapes.geometric}
\usetikzlibrary{arrows}
\usetikzlibrary{arrows.meta,calc,decorations.markings,math,arrows.meta}

\usepackage{amsmath}

\usepackage{bbm}
\usepackage{wrapfig}
\usepackage{enumitem}
\usepackage[stable]{footmisc}

\newtheorem{definition}{Definition}
\newtheorem{assumption}{Assumption}
\newtheorem{lemma}{Lemma}

\newtheorem{theorem}{Theorem}

\usepackage{array}
\newcolumntype{L}[1]{>{\raggedright\let\newline\\\arraybackslash\hspace{0pt}}m{#1}}
\newcolumntype{C}[1]{>{\centering\let\newline\\\arraybackslash\hspace{0pt}}m{#1}}
\newcolumntype{R}[1]{>{\raggedleft\let\newline\\\arraybackslash\hspace{0pt}}m{#1}}

\definecolor{green}{rgb}{0,0,0}
\definecolor{red}{rgb}{0,0,0}
\definecolor{blue}{rgb}{0,0,0}
\definecolor{purple}{rgb}{0,0,0}
\definecolor{orange}{rgb}{0,0,0}

\title{Byzantine Resilient Distributed Multi-Task Learning}
\author{%
  Jiani Li, Waseem Abbas, and Xenofon Koutsoukos
\\
  Department of Electrical Engineering and Computer Science\\
  Vanderbilt University, Nashville, TN, USA \\
  \texttt{\{jiani.li, waseem.abbas, xenofon.koutsoukos\}@vanderbilt.edu} \\
}

\begin{document}

\maketitle

\begin{abstract}
Distributed multi-task learning provides significant advantages in multi-agent networks with heterogeneous data sources where agents aim to learn distinct but correlated models simultaneously. However, distributed algorithms for learning \textcolor{blue}{relatedness} among tasks are not resilient in the presence of Byzantine agents. In this paper, we present an approach for Byzantine resilient distributed multi-task learning. We propose an efficient online weight assignment rule by measuring the accumulated loss using an agent's data and \textcolor{red}{its neighbors' models}. A small accumulated loss indicates a large similarity between the two tasks. \textcolor{green}{In order to} ensure the Byzantine resilience of the aggregation at a normal agent, we introduce a step for filtering out larger losses. We analyze the approach \textcolor{red}{for convex models} and show that normal agents converge resiliently towards \textcolor{purple}{the global minimum}.
% Further, an agent's learning performance using the proposed weight assignment rule is guaranteed to be at least as good as \textcolor{red}{in the} non-cooperative case as measured by the expected regret. 
\textcolor{orange}{Further, aggregation with the proposed weight assignment rule always results in an improved expected regret than the non-cooperative case.}
Finally, we demonstrate the approach using three case studies, \textcolor{red}{including} regression and classification \textcolor{red}{problems}, and \textcolor{green}{show} that our method \textcolor{red}{exhibits} good empirical performance for non-convex models, such as convolutional neural networks.
\end{abstract}

\input{introduction}

\input{related}

\input{problem}

\input{aggregation}

\input{resilience}

\input{evaluation_new}

\section{Conclusion}
In this paper, we \textcolor{green}{propose} an efficient online weight adjustment rule for learning the similarities among agents in distributed multi-task networks \textcolor{red}{with} an arbitrary number of Byzantine agents. 
We \textcolor{green}{argue} \textcolor{black}{that a} widely used approach of measuring the similarities based on the distance between \textcolor{red}{two agents'} model parameters is vulnerable to Byzantine attacks. To cope with such vulnerabilities, we \textcolor{green}{propose} to measure similarities based on the (accumulated) loss using an agent's data and \textcolor{red}{its neighbors' models}. \textcolor{black}{A small loss \textcolor{green}{indicates} a large similarity between the agents}. 
\textcolor{black}{To eliminate the}  influence of Byzantine agents, \textcolor{black}{a normal agent}   \textcolor{green}{filters} out the information from neighbors \textcolor{black}{whose losses \textcolor{green}{are} larger than the agent's own loss.}  With filtering, 
\textcolor{orange}{aggregation using the loss-based weight adjustment rule results in an improved expected regret than the non-cooperative case and guarantees that}
% \textcolor{purple}{promotes the convergence} and
each normal agent converges resiliently towards \textcolor{purple}{the global minimum}.
% with an improved expected regret than the non-cooperative case.
The experiment results \textcolor{green}{validate} the effectiveness of our approach.
% We believe our method can be generalized to other fields such as meta learning, transfer learning, and clustering.
% \textcolor{red}{(It might be a good idea to write one or two sentences regarding some future extension.)}

\section*{Broader Impact}
% including its ethical aspects and future societal consequences. Authors should take care to discuss both positive and negative outcomes.

% The privacy and security issue in distributed learning has been extensively considered.
\textcolor{black}{The}   problem of Byzantine resilient aggregation of distributed machine learning models has been actively studied in recent years; \textcolor{black}{however,} the \textcolor{black}{issue} of Byzantine resilient distributed  learning  in  multi-task networks has  received  much  less attention. 
% However, few work tries to address the cyber-vulnerabilties of distributed MTL, 
It is a general intuition that  MTL is robust and resilient to cyber-attacks since it can identify attackers by measuring similarities between neighbors. \textcolor{black}{In this paper, we have shown that}  some \textcolor{black}{commonly used} similarity \textcolor{black}{measures}  are not \textcolor{black}{resilient against certain attacks}.
\textcolor{black}{With an increase in} data heterogeneity, 
we hope this work could \textcolor{black}{highlight}  the security and privacy concerns in designing distributed MTL frameworks.

\section*{Acknowledgments and Disclosure of Funding}
\textcolor{red}{
This work is supported in part by the NSA Lablet (H98230-18-D-0010). Any opinions, findings, and conclusions or recommendations expressed in this material are those of the author(s) and do not necessarily reflect the views of NSA.
}

% \section*{References}

% References follow the acknowledgments. Use unnumbered first-level heading for
% the references. Any choice of citation style is acceptable as long as you are
% consistent. It is permissible to reduce the font size to \verb+small+ (9 point)
% when listing the references.
% {\bf Note that the Reference section does not count towards the eight pages of content that are allowed.}
% \medskip

% \small

% [1] Alexander, J.A.\ \& Mozer, M.C.\ (1995) Template-based algorithms for
% connectionist rule extraction. In G.\ Tesauro, D.S.\ Touretzky and T.K.\ Leen
% (eds.), {\it Advances in Neural Information Processing Systems 7},
% pp.\ 609--616. Cambridge, MA: MIT Press.

% [2] Bower, J.M.\ \& Beeman, D.\ (1995) {\it The Book of GENESIS: Exploring
%   Realistic Neural Models with the GEneral NEural SImulation System.}  New York:
% TELOS/Springer--Verlag.

% [3] Hasselmo, M.E., Schnell, E.\ \& Barkai, E.\ (1995) Dynamics of learning and
% recall at excitatory recurrent synapses and cholinergic modulation in rat
% hippocampal region CA3. {\it Journal of Neuroscience} {\bf 15}(7):5249-5262.

\small
\bibliographystyle{unsrt}
\bibliography{references}

\input{appendix}

\end{document}

%% file: introduction.tex
\section{Introduction}

Distributed machine learning models are gaining much attention recently as they improve the learning capabilities of agents distributed within a network with no central entity. In a distributed multi-agent system, agents interact with each other to improve their learning capabilities by leveraging the shared information via exchanging either data or models.
%(They can exchange data or models. For privacy concerns, we consider exchanging models in this paper). 
In particular, agents that do not have enough data to build refined models or agents that \textcolor{black}{have}  limited computational capabilities, benefit most from such cooperation. 
Distributed learning also addresses the single point of failure problem as well as scalability issues and is naturally suited to mobile phones, autonomous vehicles, drones, healthcare, smart cities, and many other applications \cite{DBLP:journals/corr/KonecnyMYRSB16, DBLP:conf/esann/AnguitaGOPR13, DBLP:journals/corr/abs-1812-11750, DBLP:journals/corr/abs-1907-09173}.
In networks with heterogeneous data sources, it is natural to consider the multi-task learning (MTL) framework, where agents aim to learn distinct but correlated models simultaneously \cite{DBLP:conf/nips/SmithCST17}. Typically, prior knowledge of the relationships among models is assumed in MTL. The relationships among agents can be promoted via several methods, such as mean regularization, clustered regularization, low-rank and sparse structures regularization  \cite{DBLP:conf/kdd/EvgeniouP04, DBLP:conf/nips/ZhouCY11, DBLP:conf/kdd/ChenZY11}. However, in real-world applications, such relationships are unknown beforehand and need to be estimated online from data. Learning similarities among tasks to promote effective cooperation is a primary consideration in MTL. There has been extensive work for learning the relationship matrix \emph{centrally} by optimizing a global convex regularized function \cite{DBLP:conf/nips/JacobBV08, DBLP:conf/uai/ZhangY10, DBLP:journals/jmlr/SahaRDV11}. In contrast, this paper focuses on computationally efficient \emph{distributed} learning of the relationship among agents that does not require optimizing a relationship matrix centrally
%optimizing a centralized relationship matrix 
\cite{6232902,  7060710,  DBLP:conf/nips/MurugesanLCY16,DBLP:conf/nips/MurugesanC17}.
%-------------------------------------

Although \textcolor{red}{the} distributed approach \textcolor{red}{to} learning and promoting similarities among neighbors from online data has many advantages, it is not resilient \textcolor{green}{to Byzantine agents}. 
Fault-tolerance for MTL \textcolor{black}{is} discussed in \cite{DBLP:conf/nips/SmithCST17}, focusing on dropped nodes that occasionally stop sending information to their neighbors.
In \cite{DBLP:LiWK20}, the relationship promoted by measuring the quadratic distance between two model parameters for distributed MTL is shown to be vulnerable to gradient-based attacks, 
and a Byzantine resilient distributed MTL algorithm is proposed for regression problems to 
cope with such attacks. 
\textcolor{black}{The proposed} algorithm relies on a user-defined parameter $F$ to filter out \textcolor{black}{information from $F$ neighbors in the aggregation step} and is resilient to $F$ Byzantine neighbors, but requires exponential time with respect to the number of \textcolor{purple}{neighbors}. 
%And as we discuss in this paper, any relationship promoted by measuring the distance between two model parameters can be easily attacked using simple attack strategies.

In this paper, 
% we study the online weight adjustment rules used to measure the similarities among agents. We \textcolor{blue}{note} that measuring similarities based on the distance between model vectors of two agents are vulnerable to cyber-attacks. 
% As such, 
we propose an \emph{online weight adjustment rule} for MTL that is  guaranteed to  achieve resilient \textcolor{purple}{convergence} for every normal agent using the rule.
% , in the presence of arbitrary number of Byzantine agents.
Compared to \cite{DBLP:LiWK20}, the proposed method is suited for both regression and classification problems, is resilient to an arbitrary number of Byzantine \textcolor{purple}{neighbors} (without the need to select a pre-defined parameter $F$ bounding the number of Byzantine \textcolor{purple}{neighbors}), and has linear time complexity.
% \todo[inline]{This is a very strong statement. We have to be careful here. We should also mention the (1) limitation of the approach, and/or (2) specify the cost of such a high resilience (may be in detrioration of accuracy) etc.} 
To the best of our knowledge, this is the first solution that aims to address the Byzantine resilient cooperation in distributed MTL networks via a resilient similarity promoting method.
\textcolor{black}{We} note that the proposed rule is not limited to the multi-task setting but can also be used for general distributed machine learning and federated learning systems to achieve resilient consensus.  We list our contributions below.
% \todo[inline]{It would be good to write a sentence in which we compare the proposed solution to the best available solution of distributed MTL. For instance, if there is no solution that consider the distributed online computation of relations while solving MT problem, we can mention that, "To the best of our knowledge, this is the first solution ...". Other wise just compare it with the current state of art.}

\begin{itemize}[leftmargin=*]
    \item We propose an efficient Byzantine resilient online weight adjustment rule for distributed MTL. We measure similarities among  \textcolor{red}{agents} based on the accumulated loss of 
    an agent’s data and the models of its neighbors.
    In each iteration, a normal agent computes the weights assigned to its neighbors in time that is linear in  the size of its neighborhood and the dimension of the data.
    % The weights can be computed in $\mathcal{O}(|\mathcal{N}_k|(d_x+d_y))$ time for each agent at each iteration where $d_x$ and $d_y$ are the dimensions of the online data and $|\mathcal{N}_k|$ is the neighborhood size of $k$.
    %\textcolor{blue}{in networks that is} based on the similarity \textcolor{red}{measured via loss of other agents' model vectors on ego agent's data.}\todo{Not very clear.} \textcolor{blue}{To implement the rule, each node performs $O(|\mathcal{N}_k|d)$ computations in every iteration, where $\mathcal{N}_k$ is the neighborhood set of $k$ and $d$ is the dimension of models.}\todo[inline]{I have combined the first two contributions.}
    
    \item We show that 
    % \textcolor{purple}{the aggregation step using the proposed weight assignment rule promotes convergence of normal agents as measured by the expected regret and normal agents converge resiliently towards the global minimum.}
    \textcolor{orange}{aggregation with the proposed weight assignment rule always results in an improved expected regret than the non-cooperative case, and normal agents converge  resiliently  towards the global minimum.}
    % We show that using the proposed rule, normal agents with convex models converge resiliently towards \textcolor{purple}{the global minimum} and
    % with an improved learning performance compared to the non-cooperative case \textcolor{black}{as} measured by the expected regret at convergence. 
    Even when all the neighbors are Byzantine, a normal agent can still resiliently converge to  \textcolor{purple}{the global minimum bounded by}  the same expected regret as without \textcolor{black}{any} cooperation \textcolor{black}{with other agents}, achieving resilience to \textcolor{black}{an} arbitrary number of Byzantine agents.
    
    %     with an expected regret 
    % % $    \lim_{i \rightarrow \infty}  R_{k}^{\text{\rm(coop)}}(i) = \frac{1}{|\mathcal{N}^{\leq}_k|} \frac{\mu L}{2m(2 - \mu L)} \sum_{l \in \mathcal{N}^{\leq}_k} \sigma_l^2$, that is 
    % guaranteed to be smaller than the non-cooperative case if at least one of its neighbors .
    % % where    $\lim_{i \rightarrow \infty}  R_{k}^{\text{\rm(ncop)}}(i) = \frac{\mu L}{2m(2 - \mu L)}  \sigma_k^2$ since $\sigma_l^2 \leq \sigma_k^2$ is true for $l \in \mathcal{N}_k^{\leq}$. 
    % Even when all its neighbors are Byzantine, 
    
    % \todo[inline]{We can also mention specific "technical" results here. It will be good to have a strong and technical list f contributions.}
    \item We conduct three experiments for both regression and classification problems \textcolor{red}{and demonstrate that our approach yields} %to validate our method and show that it exhibits g
    good empirical performance for non-convex models, such as convolutional neural networks.
\end{itemize}

%% file: related.tex
\section{Related Work}
% \todo[inline]{I think we can combine Sections 3 and 4 into one section with the heading "Background". First subsection in that section could be "Related Work" (which is current Section 2), and the second subsection could be "Distributed Multi-task Learning" (which is currently Section 3 with the heading Preliminaries).}

\textcolor{blue}{
\textbf{Multi-Task Learning.}
MTL  deals with the problem of learning multiple related tasks simultaneously to improve the
% \textcolor{red}{overall} %generalization 
\textcolor{red}{generalization}
performance of the models learned by each task with the help of the other auxiliary tasks \cite{DBLP:journals/ml/Caruana97, DBLP:journals/corr/Ruder17a}.
\textcolor{red}{The extensive literature in MTL can be broadly categorized into two categories based on how the data is collected.}
%While there is an extensive literature in MTL, most approaches can be broadly categorized based on how the data is collected.
The \emph{centralized} approach assumes the data is collected beforehand at a centralized entity. Many successful MTL applications with deep networks, such as in natural language processing and computer vision, fall into this category \cite{DBLP:conf/nips/LongCWY17, DBLP:conf/cvpr/MisraSGH16, DBLP:conf/emnlp/HashimotoXTS17,  DBLP:conf/cvpr/KendallGC18}.  
This approach usually learns multiple objectives from a shared representation by sharing layers and splitting architecture in the deep networks. 
On the other hand, the \emph{distributed} approach assumes data is collected separately by each task in a distributed manner. This approach is naturally suited to model distributed learning in multi-agent systems such as mobile phones, autonomous vehicles, and smart cities \cite{DBLP:conf/esann/AnguitaGOPR13, DBLP:journals/corr/abs-1812-11750, DBLP:journals/corr/abs-1907-09173}.
We focus on distributed MTL  in this paper.
}

\textbf{Relationship Learning in MTL.}
\textcolor{red}{Although it is often  assumed that a clustered, sparse, or low-rank structure among tasks is known \textit{a priori} \cite{DBLP:conf/kdd/EvgeniouP04, DBLP:conf/nips/ZhouCY11, DBLP:conf/kdd/ChenZY11}, such information} \textcolor{blue}{may not be available in many real-world applications.} 
Learning the \textcolor{blue}{relatedness} among tasks online from data to promote effective cooperation is a principle approach in MTL \textcolor{blue}{when the relationships among tasks are not known \textit{a priori}.} 
There has been extensive work in online relationship learning that can be broadly categorized into \textcolor{black}{centralized and distributed methods.}
The first group assumes that a centralized server collects the task models and utilizes a convex formulation of the regularized MTL optimization problem over the relationship matrix,  which is learned by \textcolor{black}{solving} the convex optimization problem \cite{DBLP:conf/nips/JacobBV08, DBLP:conf/uai/ZhangY10, DBLP:journals/jmlr/SahaRDV11}.
The second group \textcolor{black}{relies on a distributed architecture in which} agents learn relationships with their neighbors based on the similarities of their models and \textcolor{black}{accordingly} adjust weights assigned to neighbors \cite{6232902,  7060710,  DBLP:conf/nips/MurugesanLCY16,DBLP:conf/nips/MurugesanC17}.
Typical similarity metrics, such as $\mathcal{H}$ divergence \cite{10.5555/2946645.2946704, Konstantinov2019RobustLF, Shui2019APA} and Wasserstein distance \cite{Shui2019APA, NIPS2018_7913}, can be used in MTL \textcolor{red}{in the same way they are used in} domain adaptation,  transfer learning, and adversarial learning. However, such metrics are mainly designed for measuring the divergence in data distributions and are not suitable for online relationship learning \textcolor{black}{due to} efficiency and privacy concerns in data sharing.

% Due to security and privacy concerns, agents usually share their model vectors rather than sharing data directly. 
% By performing optimization using local data and averaging the model vectors of neighbors iteratively, agents will converge to a global optimal solution of consensus.
% Though as the most popular cooperative mechanism, averaging is susceptible to cyber-attacks and require high system integrity \cite{McMahan2016FederatedLO, McMahan2016CommunicationEfficientLO, Li2020On}. Non-cooperative or Byzantine agents
% sharing malicious message can easily disrupt the convergence
% of the algorithm and may even drive normal agents to a malicious point of convergence \cite{DBLP:conf/nips/BlanchardMGS17, DBLP:LiWK20} (See \figref{fig: average under attack}). 
% As such, various Byzantine resilient cooperation algorithms are studied to guarantee resilient distributed consensus of model vectors.

\textbf{Resilient Aggregation in Distributed ML.} 
\textcolor{black}{Inspired by the resilient consensus algorithms in multi-agent networks \textcolor{blue}{\cite{DBLP:journals/pieee/Olfati-SaberFM07, leblanc2013resilient}}, various resilient aggregation rules have been adapted in distributed ML, including the coordinate-wise trimmed mean \cite{DBLP:conf/icml/YinCRB18}, the coordinate-wise median \cite{DBLP:conf/icml/YinCRB18, DBLP:journals/corr/abs-1906-01736, DBLP:journals/corr/abs-1909-04532}, the geometric median \cite{Chen:2017:DSM:3175501.3154503, krishana2019}, and the Krum algorithm \cite{DBLP:conf/nips/BlanchardMGS17}.} 
% Byzantine resilient aggregation in distributed learning has been actively studied with the goal of  \textcolor{black}{achieving} consensus among the networked agents in the presence of Byzantine agents.
% Many methods are inspired by the resilient consensus algorithms, such as the coordinate-wise trimmed mean \cite{DBLP:conf/icml/YinCRB18}, the coordinate-wise median \cite{DBLP:conf/icml/YinCRB18, DBLP:journals/corr/abs-1906-01736, DBLP:journals/corr/abs-1909-04532}, the geometric median \cite{Chen:2017:DSM:3175501.3154503, krishana2019}, and the Krum algorithm \cite{DBLP:conf/nips/BlanchardMGS17}, \textcolor{black}{have been proposed}. 
However, studies have shown \textcolor{black}{that these rules} are not resilient \textcolor{black}{against} certain attacks
\cite{DBLP:conf/nips/BaruchBG19, DBLP:conf/uai/XieKG19, DBLP:journals/corr/abs-1911-11815}.
% \cite{centerpoint_RSS, DBLP:conf/nips/BaruchBG19, DBLP:conf/uai/XieKG19, DBLP:journals/corr/abs-1911-11815}.
%The \textcolor{black}{main} reason is that these rules cannot ensure that in the presence of Byzantine agents, the aggregation result always lies in the convex hull of normal {agents'} models. Such a condition, which is \textcolor{black}{often called the \emph{safety} condition, is necessary for the resilient convergence of models using these aggregation rules.} %usually referred to as the \emph{safety} condition in resilient vector consensus and is the key for the resilient convergence of models \cite{centerpoint_RSS}.
\textcolor{blue}{The centerpoint based aggregation rule \cite{centerpoint_RSS} has been proposed recently that guarantees resilient distributed learning to Byzantine attacks.
% \textcolor{black}{however, the centerpoint-based} method is limited to low-dimensional problems \textcolor{black}{due to the complexity of computing a centerpoint in higher dimensions.} %since it is computational expensive for computing  a centerpoint in high dimensions.
However, s}ince each agent fits a distinct model in MTL, consensus-based resilient aggregation rules are not directly  applicable to MTL.

%% file: problem.tex
\section{Distributed Multi-Task Learning}\label{sec: distributed MTL}
\textbf{Notation.}
In this paper, 
$|A|$  denotes the cardinality of a set $A$, $\|\cdot\|$ denotes the $\ell_2$ norm, $\text{Tr}(\cdot)$ denotes the trace of a matrix, and $\mathbb{E}_{\xi}[\cdot]$ denotes the expected value 
% \st{taken with respect to the distribution generated by}
\textcolor{black}{of} a random variable $\xi$. If the context is clear, $\mathbb{E}[\cdot]$ is used.
% And $\mathbb{E}[\cdot]$ denotes the expected value taken with respect to the joint distribution of all random variables $\xi^1, \xi^2, \ldots, \xi^{i-1}$, i.e.
% $
%     \mathbb{E}  \left[\cdot \right] = \mathbb{E}_{\xi^1}   \mathbb{E}_{\xi^2} \ldots  \mathbb{E}_{\xi^{i-1}}       
%     \left[\cdot \right].
% $
% \textbf{Distributed Multi-Task Learning.}\quad
% 
% Our objective is then to design weight assignment rule that approximately solves \eqref{eq: optimize weight} for each agent $k$, which then approximately solves  \eqref{eq: cost function each agent}.
% Note that if our method can work for multi-task setting, then it is sure to be resilient to the single-task setting where all the agents estimate the same parameter of interest.
%We consider the model of peer-to-peer distributed multi-task  learning (MTL) where networked agents learn distinct and correlated models   cooperatively. %by aggregating neighbors' models to construct an improved model.

\textbf{Background.}
Consider a network of \textcolor{purple}{$n$} agents\footnote{Each agent is modeled as a separate task, thus, the terms \emph{agent} and \emph{task} are used interchangeably.} modeled by an \emph{undirected graph}
$\mathcal{G}=(\mathcal{V},\mathcal{E})$, where $\mathcal{V}$ represents agents and $\mathcal{E}$ represents interactions between agents. 
A bi-directional edge $(l,k) \in \mathcal{E}$ means that agents $k$ and $l$ can exchange information with each other. \textcolor{black}{Since each agent also has its own information, we have} $(k,k) \in \mathcal{E}, \forall k \in \mathcal{V}$. The \emph{neighborhood} of $k$ is the set  $\mathcal{N}_k =\{l\in \mathcal{V}| (l,k)\in\mathcal{E}\}$. 
% Each agent $k$ has data \textcolor{red}{$\left\{(x_k^i, y_k^i) \right\}_{i \in \mathcal{S}_k}$} sampled randomly from the distribution generated by the random variable $\xi_k$, where $x_k^i \in \mathbb{R}^{d_x}$, $y_k^i \in \mathbb{R}^{d_y}$, 
% % \st{${d_x}$ and ${d_y}$ are the dimensions of $x_k^i$ and $y_k^i$} 
% and \textcolor{red}{$\mathcal{S}_k$ is the sample set}. 
\textcolor{red}{
Each agent $k$ has data \textcolor{red}{$\left\{(x_k^i, y_k^i) \right\}$} sampled randomly from the distribution generated by the random variable $\xi_k$, where $x_k^i \in \mathbb{R}^{d_x}$, $y_k^i \in \mathbb{R}^{d_y}$.}
% \st{${d_x}$ and ${d_y}$ are the dimensions of $x_k^i$ and $y_k^i$} 
%We consider a convex \emph{prediction function} (model) $f(x_k^i, \theta_k) = \theta_k^\top x_k^i$, where $\theta_k \in \mathbb{R}^{d_x}$ is the model parameter.
We use $\ell(\theta_k; \xi_k)$ to denote a convex \emph{loss function} associated with the prediction function parameterized by $\theta_k$ for agent $k$. 
% and $\ell_k(f_{\theta_k}(x_k^i), y_k^i)$ denotes the loss using $\theta_k$ for $(x_k^i, y_k^i)$.
% For simplicity, we use  $\ell_{k}^i(\theta_k) \triangleq  \ell_k(f_{\theta_k}(x_k^i), y_k^i) $  to denote the loss of agent $k$ at iteration $i$. And similarly, $ \ell_{k}^i(\theta_l) \triangleq  \ell_k(f_{\theta_l}(x_k^i), y_k^i)$ denotes the loss 
% of a neighbor's model $\theta_l$ using agent $k$'s data at iteration $i$.
\textcolor{black}{MTL is concerned with fitting}  separate models ${\theta}_k$ to the data for agent $k$ via the \emph{expected risk function} 
% $
%     r_k(\theta_k) = \frac{1}{n_k}
%      \sum_{i=1}^{n_k}  \ell_k(f_{\theta_k}(x_k^i), y_k^i) 
%     % = \frac{1}{n_k}
%     % \sum_{i=1}^{n_k} \ell_k^i({\theta_k})
% $ or  
$
    % r_k(\theta_k) = \mathbb{E}
    % \left[\ell_k(f_{\theta_k}(x_k^i), y_k^i) \right]
    r_k(\theta_k) = \mathbb{E}
    \left[ \ell(\theta_k; \xi_k) \right]
$.
We use $\theta_k^*$ to denote the global minimum of the convex function  $r_k(\theta_k)$.
% Let $D_k \sim \varphi_k (x,y)$
% Then the empirical risk function of $k$ can approximate the true risk arbitrarily well if the sample size is large enough, which is given by
% $
%     r_k(\theta_k) %=\sum_{(x_k^i,y_k^i) \in D_k} \ell_k({\theta_k}^\top x_k^i, y_k^i) 
%     = \frac{1}{n_k}
%     \sum_{i=1}^{n_k} \ell_k({\theta_k}^\top x_k^i, y_k^i).
% $
The model parameters can be optimized via the following objective function:
% \begin{equation}\label{eq: cost function MTL}
% \min_{[\theta_1, \ldots, \theta_m]} \left\{  \sum_{k=1}^m \sum_{i=1}^{n_k} \ell_{k}^i(\theta_k) \right\},
% \end{equation}
% \vspace{-0.2cm}
\begin{equation}\label{eq: cost function MTL}
\min_{\Theta} \left\{  \sum_{k=1}^{\textcolor{purple}{n}}  r_{k}(\theta_k) + \eta \mathcal{R}(\Theta, \Omega) \right\},
%, \text{ subject to } \Theta \in \Omega
\end{equation}
where $\Theta = [\theta_1, \ldots, \theta_{\textcolor{purple}{n}}]\in \mathbb{R}^{d_x \times \textcolor{purple}{n}}$,  $\mathcal{R}(\cdot)$ is a convex  regularization  function  promoting the relationships among the agents, and $\Omega \in \mathbb{R}^{\textcolor{purple}{n} \times \textcolor{purple}{n}}$ models the relationships among the agents that can be \textcolor{black}{assigned} \textit{a priori} or can be estimated from data. 
% The problem \eqref{eq: cost function MTL} can then be decomposed into $m$ separate optimization problems, one for each agent $k$:
% \begin{equation}\label{eq: cost function each agent}
% \min_{\theta_k}  r_k(\theta_k).
% \end{equation}
An example of the regularizer takes the  form of 
% \begin{equation*}\label{eq: quadratic regularizer}
    % \mathcal{R}(\Theta, \Omega) %= \Theta^\top \mathcal{L} \Theta = 
    % =\frac{1}{2} \sum_{k=1}^m \sum_{l \in \mathcal{N}_k}  c_{lk} \|\theta_k -\theta_l\|^2,
   $\mathcal{R}(\Theta, \Omega) = \lambda_1 \text{Tr}(\Theta\Omega\Theta^\top) + \lambda_2 \text{Tr}(\Theta \Theta^\top),$
% \end{equation*}
where $\lambda_1, \lambda_2$ are non-negative parameters. 
\textcolor{black}{In a \emph{centralized} setting, where  a centralized server optimizes the relationship matrix by collecting the models of agents, an optimal solution}
$\Omega = \frac{(\Theta^\top \Theta)^{\frac{1}{2}}}{\text{Tr}\left((\Theta^\top \Theta)\right)^{\frac{1}{2}}}$ is proposed in 
\cite{DBLP:conf/uai/ZhangY10}  for learning the structure of clustered MTL using the above regularizer. 
% \begin{equation}\label{eq: quadratic regularizer}
%     \mathcal{R}(\Theta, \Omega) %= \Theta^\top \mathcal{L} \Theta = 
%     =\frac{1}{2} \sum_{k=1}^m \sum_{l \in \mathcal{N}_k}  c_{lk} \|\theta_k -\theta_l\|^2,
% \end{equation}
% where $c_{lk}$ reflects the strength of the relation between $k$ and $l$.
% In addition to the quadratic regularizer $ \|\theta_k -\theta_l\|^2$ in \eqref{eq: quadratic regularizer}, other typical choices of the regularization term include the $\ell_1$-norm reguarizer $\|\theta_k - \theta_l\|$ and the $\ell_2$-norm reguarizer $\|\theta_k - \theta_l\|_2$.
% $r_k$ is also convex and has a unique minimizer $\theta_k^*$ with $r_k^* = r_k(\theta_k^*)$.
\textcolor{black}{In the \emph{distributed}  case}, the task relationships $\Omega$ are not learned centrally and we can use the \emph{adapt-then-combine (ATC) diffusion} algorithm \cite{DBLP:journals/spm/NassifVRCS20} as a projection-based distributed solution of \eqref{eq: cost function MTL}:
    \begin{align}
  &  {\hat{\theta}}_{k,i} = {\theta}_{k,i-1} - \mu_{k}  \nabla \ell({\theta_{k,i-1}}; \xi_k^{i-1}), \hspace{-3cm} &  \text{(adaptation)} \label{eq: adapt}  \\
   & \theta_{k,i} = \sum_{l \in \mathcal{N}_k} a_{lk} {\hat{\theta}}_{l,i}, \text{subject to} \sum_{l \in \mathcal{N}_k} a_{lk} = 1, a_{lk}\geq0, a_{lk}=0 \text{ if } l\not\in \mathcal{N}_k,  & \text{(combination)} \label{eq: combine}
    \end{align}
% \begin{equation}
% \label{eq:adapt}
% {\hat{\theta}}_{k,i} = {\theta}_{k,i-1} + \mu_k  \nabla_{\theta_k} \ell_k({\theta_k}^\top x_k^i, y_k^i)  \hspace{0.2in} \text{(adaptation)}
% \end{equation}
% \begin{equation}
% \label{eq:combine}
% \theta_{k,i} = \sum_{l \in \mathcal{N}_k} a_{lk} {\hat{\theta}}_{l,i}  \; \hspace{1.12in} \text{(combination)}
% \end{equation}
where $\mathcal{N}_k$ is the neighborhood of agent $k$,  $\mu_{k}$ is the step size, and 
$a_{lk}$ denotes the weight assigned by agent $k$ to $l$, 
% that satisfies the following constraints:
% %\begin{equation}\label{eq: weight assignment}
% $
% \sum_{l \in \mathcal{N}_k} a_{lk} = 1, a_{lk}\geq0, a_{lk}=0 \text{ if } l\not\in \mathcal{N}_k
% $, 
which should accurately reflect the similarity relationships among agents\footnote{$\mu_{k}$ and $a_{lk}$ can be time-dependent, but when context allows, we write $\mu_{k,i}$ as $\mu_{k}$ and $a_{lk}(i)$ as $a_{lk}$ for simplicity.}.
$\nabla \ell({\theta_{k,i-1}}; \xi_k^{i-1})$ is the gradient using the instantaneous realization $ \xi_k^{i-1}$ of the random variable $\xi_k$.
\textcolor{black}{At each iteration $i$, agent $k$}  minimizes the individual risk using stochastic gradient descent (SGD) given local data followed by a combination step that aggregates neighboring models \textcolor{black}{according to the weights assigned to them.}
The weights $\{a_{lk}\}$ are free parameters selected by the designer \textcolor{black}{and they serve the same purpose} as $\Omega$ in a centralized formulation. Thus, there is no need to \textcolor{black}{design}  $\Omega$ \textcolor{black}{in the case of distributed MTL that utilizes ATC diffusion algorithm for aggregation} \cite{6845334}.

%\end{equation}

% Put \eqref{eq:combine} into \eqref{eq: cost function each agent}, 
% and use the current loss $\ell_k({\theta_k}^\top x_k^i, y_k^i)$ to approximate the true empirical risk  $r_k(\theta_k)$, we then have goal \eqref{eq: cost function each agent} as 
% \begin{equation}\label{eq: optimize weight}
% \begin{aligned}
%   & \min_{a_{1k}, \ldots, a_{|\mathcal{N}_k| k}}  \ell_k({ \sum_{l \in \mathcal{N}_k} a_{lk} {\hat{\theta}}_{l,i}}^\top x_k^i, y_k^i)  \\
%      s.t. \; & \sum_{l \in \mathcal{N}_k} a_{lk} = 1,    \quad a_{lk}\geq0, \quad a_{lk}=0 \text{ if } l\not\in \mathcal{N}_k.
% \end{aligned}
% \end{equation}
%Typically when all the agents share the common objective model, the combination step aggregates the models by averaging and the algorithm can converge with a better learning performance than without such aggregation \cite{}.  
%However, averaging is vulnerable to Byzantine attacks and is not applicable to MTL \cite{}.
% However, in the MTL setting, average-based aggregation will lead to a common model of convergence yet agents share distinct models, rendering it not useful. %MTL usually assumes a structure of clustering that can be captured by the design of the weight matrix.

\textbf{Online Weight Adjustment Rules.}
%It is natural to assume a cluster structure of the MTL network where agents in the same cluster have similar parameter vectors to each other. 
Without knowing the relationships \textit{a priori}, one can assume the existence of similarities among agents and can learn these similarities online from data. 
\textcolor{black}{The approach is based on the distance between the model parameters of agents, where a  small distance indicates a large similarity \cite{6232902, 7060710, 9005223, CHEN2018107}. A common approach to learning similarities between two agents online is given by }
%Below is a common approach \cite{6232902, 7060710, 9005223, CHEN2018107} in learning the similarities online based on the quadratic distance between the model parameters of two agents where a small distance indicates a large similarity
% \vspace{-0.2cm}
\begin{equation}\label{eq: quadratic distance weight}
    a_{lk}(i) = \frac{\|\tilde{\theta}_k^* - \hat{\theta}_{l,i}\|^{-2}}{\sum_{p \in \mathcal{N}_k}\|\tilde{\theta}_k^* - \hat{\theta}_{p,i}\|^{-2}},
\end{equation}
where 
% $\|\cdot\|$ denotes the $\ell_2$ norm, and 
$\tilde{\theta}_k^*$ is an approximation of $\theta_k^*$ since $\theta_k^*$ is unknown. Examples include using the current model $\tilde{\theta_k^*} = \theta_{k,i-1}$, and one-step ahead approximation $\tilde{\theta_k^*} = \hat{\theta}_{k,i} + \mu_k \nabla \ell(\hat{\theta}_{k,i}; \xi_k^{i-1})$.
Although the $\ell_2$ norm is widely used, this formulation of weights can be generalized to \textcolor{black}{$\ell_p$}  norm as well. 
% The smoothing method $
%   \mathbb{E} [ \varphi_{lk}^i] = (1 - \nu_k) \mathbb{E} [  \varphi_{lk}^{i-1}] + \nu_k \mathbb{E} [\ell_{k}({{\hat{\theta}}_{l,i}^{\text{(coop)}} }; \xi_k)],
% $ has also been used in the place of $\|\tilde{\theta}_k^* - \hat{\theta}_{l,i}\|^{2}$.

% \textcolor{red}{data similarities not online method, not privacy, h divergence, no-efficiency}

\section{Problem Formulation}
Byzantine agents can send \textcolor{purple}{arbitrary}  different  \textcolor{black}{information} to different neighbors usually with a malicious goal of  \textcolor{black}{disrupting} the network's convergence.
% by \textcolor{black}{increasing}  the expected risk.% defined in Section \ref{sec: distributed MTL}. 
%The issue of dropped nodes that stop exchanging information with other agents occasionally has been discussed in \cite{DBLP:conf/nips/SmithCST17}. 
%\textcolor{black}{Here}, 
% We assume a synchronous network \textcolor{black}{in which} Byzantine agents send information to their neighbors in \textcolor{red}{each} iteration.
% \footnote{Note that to address the issue of dropped nodes, one can allocate time for each iteration and cooperate only with neighbors sending information within the allocated time period.}. % XK: This is not clear, can you rephrase
% \textcolor{red}{(The next sentence is not clear. Do we need to include it here? Do we even need to mention the issue of dropped nodes? I think we can just remove it, even in the previous sentence)} 
It has been shown in \cite{DBLP:LiWK20} that normal agents \textcolor{red}{assigning} weights according to \eqref{eq: quadratic distance weight} are vulnerable to Byzantine agents. 
\textcolor{purple}{Particularly,
by sending $\| \hat{\theta}_{b,i} -  \tilde{\theta_k^*} \| \ll \| \hat{\theta}_{k,i} - \tilde{\theta_k^*} \|$, a Byzantine agent $b$ can gain a large weight from $k$ and continuously drive its normal neighbor $k$ towards a desired malicious point.
}

%{In the presence of Byzantine agents, the vulnerabilities of weights deriving from \eqref{eq: quadratic distance weight} have been reported by \cite{DBLP:LiWK20}}.
% \textcolor{black}{This result can be stated in the context of this paper as follows:}

% \begin{lemma}\footnote{All the Proofs are given in Appendix \ref{app: Assumptions and Theoretical Results}.}
% One Byzantine agent can lead any normal agent $k$ using the online weight adjustment rules deriving from \eqref{eq: quadratic distance weight} with $\ell_p$ norm moving away from $\theta_k^*$ if $\tilde{\theta_k^*} \neq \theta_k^*$. 
% \end{lemma}

% \begin{lemma}
% \textcolor{black}{If a normal agent $k$ \textcolor{purple}{running the ATC diffusion algorithm} adapts weights according to \eqref{eq: quadratic distance weight}, then a single Byzantine agent can 
% }
% % lead $k$ away from $\theta_k^*$ if $\tilde{\theta_k^*} \neq \theta_k^*$.} 
% \end{lemma}

% \vspace{-0.5cm}

% \begin{proof}
% By sending $\| \hat{\theta}_{b,i} -  \tilde{\theta_k^*} \| \ll \| \hat{\theta}_{k,i} - \tilde{\theta_k^*} \|$ and $\| \hat{\theta}_{b,i} -  \theta_k^* \| > \| \textcolor{purple}{\hat{\theta}_{b,i-1}} - \theta_k^*\|$, a Byzantine agent $b$ can gain a large weight from $k$ by the first condition and make $\theta_{k,i}$ move  away from $\theta_k^*$ by the second condition. (The same strategy can be generalized to $\ell_p$ norm.) 
% \end{proof}

% \vspace{-0.3cm}

To address the vulnerabilities of the online weight adjustment rules derived from \eqref{eq: quadratic distance weight}, 
this paper aims to design an efficient resilient online  weight assignment rule in the presence of Byzantine agents for MTL.
% As a common setup of the analysis of the SGD algorithm, 
% we use $\mathbb{E}_{\varphi_k}[\cdot]$ to denote the expected value taken with respect to the distribution $\varphi_k$.
Let the \emph{expected regret} $\textcolor{purple}{\mathbb{E} [r_k(\theta_{k,i}) - r_k(\theta_k^*)]}$ \textcolor{black}{be}   the value of the expected difference between the risk of $\theta_{k,i}$ and the optimal decision $\theta_k^*$.
% , i.e., 
% %  $ R_k (i) =  \mathbb{E} [r_k(\theta_{k,i}) - r_k(\theta_k^*)]$.
% $\textcolor{purple}{R_k (\theta_{k,i})} =  \mathbb{E} [r_k(\theta_{k,i}) - r_k(\theta_k^*)]$.
% As a baseline, we consider the case when every normal agent runs the SGD algorithm without cooperation, i.e., $\theta_{k,i}^{\text{(ncop)}} =  {\hat{\theta}}_{k,i}^{\text{(ncop)}}$, followed by \eqref{eq: adapt}.
% We also consider the cooperative case when $\theta_{k,i}^{\text{(coop)}} = \sum_{l \in \mathcal{N}_k} a_{lk} {\hat{\theta}}_{l,i}^{\text{(coop)}}$ as indicated in \eqref{eq: combine}, followed by \eqref{eq: adapt}.
% The respective expected \textcolor{red}{regrets} for the two methods \textcolor{red}{are} given as 
% $$R_{k}^{\text{(ncop)}}(i) =  \mathbb{E} [r_k(\theta_{k,i}^{\text{(ncop)}}) - r_k(\theta_k^*)] \; \text{and} \; R_{k}^{\text{(coop)}}(i) =  \mathbb{E} [r_k(\theta_{k,i}^{\text{(coop)}}) - r_k(\theta_k^*)].
% $$
\textcolor{purple}{We aim to design} weights $A_k  = [a_{1k}, \ldots, a_{nk}] \in \mathbb{R}^{1\times n}$ for a normal agent $k$ that satisfy the following conditions:
% \begin{itemize}[leftmargin=*]
%[rightmargin=\dimexpr\linewidth-12cm-\leftmargin\relax]
%     \item compute the weights for every normal  agent $k$ as a solution of
%     \begin{equation}\label{eq: optimize weight}
% \begin{aligned}
%   & \min_{A_k} \| \theta_{k,i} - \theta_k^* \|^2,  
%   \text{ subject to }  \sum_{l \in \mathcal{N}_k}  a_{lk} = 1,    a_{lk}\geq0,  a_{lk}=0 \text{ if } l\not\in \mathcal{N}_k,
% \end{aligned}
% \end{equation}
% \item show that at each iteration, the combination step makes the model move closer to the minimizer for each normal agent $k$, i.e., 
% $$\mathbb{E} [r_k(\theta_{k,i}) - r_k^*] \leq \mathbb{E} [r_k(\hat{\theta}_{k,i}) - r_k^*],$$
% which implies each normal agent $k$ converges towards the minimizer with improved expected empirical risk using the computed weights than without such cooperation. 

\textcolor{purple}{
\textbf{Resilient Convergence.} 
\textcolor{black}{It must be guaranteed}  that using the computed weights $A_k$, every normal agent $k$ resiliently converges to  $\theta_k^*$, even in the presence of Byzantine neighbors.}

\textcolor{purple}{
\textbf{Improved Learning Performance.} 
Cooperation among agents is meaningful only when it improves the learning performance. 
Hence, it is important to guarantee that for every normal agent, the combination step using the computed weights $A_k$ always results in an  improved expected regret, even in the presence of Byzantine agents, i.e.,
% \begin{equation}\label{eq: resilient convergence  with improved learning error}
%     % \mathbb{E} [r_k(\theta_{k,i \rightarrow \infty}) - r_k^*] \leq \chi_k, \forall k
%     % \lim_{i \rightarrow \infty} \sup R_{k, \text{ATC}}(i) \leq  \lim_{i \rightarrow \infty}  R_{k, \text{SGD}}(i)
%     \lim_{i \rightarrow \infty} \sup R_{k}^{\text{(coop)}}(i) \leq  \lim_{i \rightarrow \infty}  \sup R_{k}^{\text{(ncop)}}(i), \forall k \in \mathcal{N}^+.
% \end{equation}
\textcolor{purple}{
\begin{equation}\label{eq: resilient convergence  with improved learning error}
    % \mathbb{E} [r_k(\theta_{k,i \rightarrow \infty}) - r_k^*] \leq \chi_k, \forall k
    % \lim_{i \rightarrow \infty} \sup R_{k, \text{ATC}}(i) \leq  \lim_{i \rightarrow \infty}  R_{k, \text{SGD}}(i)
    % \lim_{i \rightarrow \infty}  R_{k}^{\text{(coop)}}(i) \leq  \lim_{i \rightarrow \infty}   R_{k}^{\text{(ncop)}}(i), \forall k \in \mathcal{N}^+, i \in \mathbb{N}
        % R_{k}(\theta_{k,i}) \leq   R_{k}(\hat{\theta}_{k,i})
        \mathbb{E} [r_k(\theta_{k,i}) - r_k(\theta_k^*)] \leq 
        \mathbb{E} [r_k(\hat{\theta}_{k,i}) - r_k(\theta_k^*)]
        , \forall k \in \mathcal{N}^+, i \in \mathbb{N}
\end{equation}
}
}
\begin{comment}
\textbf{Resilient Convergence.} 
\textcolor{black}{It must be guaranteed}  that using the computed weights $A_k  = [a_{1k}, \ldots, a_{mk}] \in \mathbb{R}^{1\times m}$, every normal agent $k$ resiliently converges to the true target $\theta_k^*$
, i.e.,
\begin{equation}\label{eq: resilient convergence}
    \lim_{i \rightarrow \infty} \theta_{k,i}^{\text{(coop)}} = \theta_k^*, \forall k \in \mathcal{N}^+,
\end{equation}
where $\mathcal{N}^+$ denotes the set of normal agents in the network.
\textbf{Improved Expected Regret w.r.t. Non-Cooperation.} 
\textcolor{black}{Cooperation among agents is meaningful only when it results in improving the learning performance.}
% \st{We note that only when the cooperation steps really improve the learning performance, the cooperation is meaningful.}
Hence, it is important to guarantee that using the computed weights $A_k$, a normal agent $k$ obtains an improved expected regret  \textcolor{black}{as compared to} using the SGD algorithm without cooperation, even in the presence of Byzantine agents, i.e.,
\begin{equation}\label{eq: resilient convergence  with improved learning error}
    % \mathbb{E} [r_k(\theta_{k,i \rightarrow \infty}) - r_k^*] \leq \chi_k, \forall k
    % \lim_{i \rightarrow \infty} \sup R_{k, \text{ATC}}(i) \leq  \lim_{i \rightarrow \infty}  R_{k, \text{SGD}}(i)
    \lim_{i \rightarrow \infty} \sup R_{k}^{\text{(coop)}}(i) \leq  \lim_{i \rightarrow \infty}  \sup R_{k}^{\text{(ncop)}}(i), \forall k \in \mathcal{N}^+.
\end{equation}
\end{comment}

\vspace{-0.4cm}
\textbf{Computational Efficiency.} 
% \st{Learn an efficient weight assignment rule $A_k$  for each normal agent $k$, such that the computation of $A_k$ in each iteration takes linear time 
% with respect to both $|\mathcal{N}_k|$, and the summation of $d_x$ and $d_y$, where $A_k = [a_{1k}, \ldots, a_{mk}] \in \mathbb{R}^{1\times m}$.}
\textcolor{black}{
At each iteration, a normal agent $k$ needs to compute the weights $A_k$ in time that is linear in the size of the neighborhood of $k$ and the dimension of the data, i.e., in $O(|\mathcal{N}_k|(d_x + d_y))$ time.}

% \end{itemize}
% \textcolor{black}{If we assume that} every normal agent converges using the SGD algorithm without cooperation, condition \eqref{eq: resilient convergence  with improved learning error} directly implies \eqref{eq: resilient convergence}. Therefore, we focus on condition \eqref{eq: resilient convergence  with improved learning error} and the computational efficiency.
% Similarly, we consider optimizing \eqref{eq: cost function MTL} over the weights with fixed neighboring parameters in the combination step as follows,
% % \begin{equation}\label{eq: optimize weight}
% % \begin{aligned}
% %   & \min_{a_{1k}, \ldots, a_{|\mathcal{N}_k| k}}  \ell_{k}^i({\sum_{l \in \mathcal{N}_k} a_{lk} {\hat{\theta}}_{l,i}})  \\
% %      s.t. \;  \sum_{l \in \mathcal{N}_k} & a_{lk} = 1,    a_{lk}\geq0,  a_{lk}=0 \text{ if } l\not\in \mathcal{N}_k.
% % \end{aligned}
% % \end{equation}
% \begin{equation}\label{eq: optimize weight}
% \begin{aligned}
%   & \min_{a_{1k}, \ldots, a_{|\mathcal{N}_k| k}} 
%  r_{k}({\sum_{l \in \mathcal{N}_k} a_{lk} {\hat{\theta}}_{l,i}})  \\
%      s.t.  \; & \sum_{l \in \mathcal{N}_k}  a_{lk} = 1,    a_{lk}\geq0,  a_{lk}=0 \text{ if } l\not\in \mathcal{N}_k.
% \end{aligned}
% \end{equation}

%% file: aggregation.tex
\section{Loss-based Online Weight Adjustment}\label{sec: loss-based weight}
% We aim to optimize the weights for every normal  agent $k$ as a solution of
%     \begin{equation}\label{eq: optimize weight}
% \begin{aligned}
%   & \min_{A_k} \| \theta_{k,i} - \theta_k^* \|^2,  
%   \text{ subject to }  \sum_{l \in \mathcal{N}_k}  A_k \mathbbm{1} = 1,    a_{lk}\geq0,  a_{lk}=0 \text{ if } l\not\in \mathcal{N}_k,
% \end{aligned}
% \end{equation}
% \subsection{Weight Optimization\footnote{For better readability, we omit the superscript in this section. For instance, $\theta_{k,i}  $ are simply written as $\theta_{k,i}$}}

% In this section,  $\theta_{k,i} $ are simply written as $\theta_{k,i}$ for better readability.

\subsection{Weight Optimization} 
% \textbf{Weight Optimization.} 
We follow a typical approach of learning the optimal weight adjustment rule \cite{6232902, 7060710, 9005223, CHEN2018107} \textcolor{black}{in which the goal}  is to minimize the quadratic distance between the aggregated model $\theta_{k,i}$ and the true model $\theta_k^*$ over the weights, i.e., $\min_{A_k} \|\theta_{k,i}  - \theta_k^*\|^2$. Using \eqref{eq: combine}, we get an equivalent problem:
\begin{equation*}\label{eq: original minimization problem}
    \min_{A_k} \left\| \sum_{l \in \mathcal{N}_k} a_{lk} \hat{\theta}_{l,i}  - \theta_k^* \right\|^2,   \text{ subject to }  \sum_{l \in \mathcal{N}_k}  a_{lk} = 1,    a_{lk}\geq0,  a_{lk}=0 \text{ if } l\not\in \mathcal{N}_k,
\end{equation*}
%where 
% $A_k = [a_{1k}, \ldots, a_{mk}] \in \mathbb{R}^{1 \times m}$ and 
% $\mathbbm{1}$ denotes an $N \times 1$ column vector with all its entries equal to one.
% We can express $\| \sum_{l \in \mathcal{N}_k} a_{lk} \hat{\theta}_{l,i}  - \theta_k^*\|^2$ as 
where
% \begin{equation*}\label{eq: sum of psi - goal theta}
$
    \left\| \sum_{l \in \mathcal{N}_k} a_{lk} \hat{\theta}_{l,i}  - \theta_k^* \right\|^2 
    % =     \left\| \sum_{l \in \mathcal{N}_k} a_{lk} (\hat{\theta}_{l,i}  - \theta_k^*) \right\|^2 
    = \sum_{l \in \mathcal{N}_k} \sum_{p \in \mathcal{N}_k}  a_{lk} a_{pk} (\hat{\theta}_{l,i}  - \theta_k^*)^\top (\hat{\theta}_{p,i}  - \theta_k^*).
$
% \end{equation*}
As in a typical approximation approach, we consider 
% only the diagonal entries of the matrix form of \eqref{eq: sum of psi - goal theta}, and get the approximation
% In order to make the problem tractable, we   omit the co-effect by two agents (which is equivalent to only consider the diagonal entries of the matrix form of \eqref{eq: sum of psi - goal theta}), and get the approximation
\begin{equation}\label{eq: approximation weighted theta}
    \left\| \sum_{l \in \mathcal{N}_k} a_{lk} \hat{\theta}_{l,i}  - \theta_k^* \right\|^2 \approx  \sum_{l \in \mathcal{N}_k}   a_{lk}^2  \left\|\hat{\theta}_{l,i}  - \theta_k^*\right\|^2.
\end{equation}
The weight assignment rule \eqref{eq: quadratic distance weight} is an optimal solution of \eqref{eq: approximation weighted theta} using \textcolor{red}{the} approximation of $\theta_k^*$, which as we discuss above, can be easily attacked.
To \textcolor{black}{avoid} the use of the  distance between model parameters as a similarity measure, we introduce a \emph{resilient} counterpart, which is the \emph{accumulated loss} (or risk). \textcolor{black}{Assume} risk functions $r_k$ to be $m$-strongly convex\footnote{Details of the assumptions are given in Appendix \ref{app: assumptions}.}, \textcolor{black}{then} it holds that 
$$
    r_k(\hat{\theta}_{l,i} ) - r_k(\theta_k^*)  \geq \langle \nabla r_k(\theta_k^*),  \hat{\theta}_{l,i} - \theta_k^* \rangle + \frac{m}{2} \|\hat{\theta}_{l,i}  - \theta_k^*\|^2,
$$
where $r_k(\hat{\theta}_{l,i} ) = \mathbb{E}
    \left[ \ell(\hat{\theta}_{l,i} ; \xi_k) \right]$.
\textcolor{black}{Since}  $\nabla r_k(\theta_k^*) = 0$, we obtain 
\vspace{-0.2cm}
\begin{equation}\label{eq: upper bound of theta distance}
    \|\hat{\theta}_{l,i}  - \theta_k^*\|^2 \leq \frac{2}{m} \left( r_k(\hat{\theta}_{l,i} ) - r_k(\theta_k^*) \right).
\end{equation}
Instead of directly minimizing the right \textcolor{black}{side}  of \eqref{eq: approximation weighted theta}, we consider minimizing its upper bound given \textcolor{black}{in}~\eqref{eq: upper bound of theta distance}. \textcolor{black}{Later in Section \ref{sec: resilient convergence analysis}, we show that this alternate approach} \textcolor{red}{facilitates the} resilient distributed MTL, which cannot be achieved by minimizing the distance between models directly.
Hence, \textcolor{black}{by} combining \eqref{eq: approximation weighted theta} and \eqref{eq: upper bound of theta distance}, we consider \textcolor{black}{the following} minimization problem: 
\begin{equation*}
    \min_{A_k} \sum_{l \in \mathcal{N}_k}   a_{lk}^2 \left( r_k(\hat{\theta}_{l,i} ) - r_k(\theta_k^*) \right)   \text{ subject to }  \sum_{l \in \mathcal{N}_k}  a_{lk} = 1,    a_{lk}\geq0,  a_{lk}=0 \text{ if } l\not\in \mathcal{N}_k.
\end{equation*}
This optimization problem indicates that if a neighbor $l$'s model has a small regret on agent $k$'s data distribution, then it should be assigned a large weight.
Since $\theta_k^*$ is unknown, one can 
use $r_k(\theta_{k,i} )$ to approximate $r_k(\theta_k^*)$. 
%If one use such approximation, cases could arise when some agent has a smaller risk than $k$ itself and therefore $r_k(\theta_{l,i}) - r_k(\theta_k^*)$ could be negative. One could consider only cooperate with agents that have a smaller risk than itself
Alternatively, since $r_k(\theta_k^*)$ is small compared to $r_k(\theta_{l,i} )$, we could simply assume $r_k(\theta_k^*) = 0$ and consider the \textcolor{black}{following} minimization problem:  
\begin{equation}\label{eq: final minimization problem}
    \min_{A_k} \sum_{l \in \mathcal{N}_k}   a_{lk}^2  r_k(\hat{\theta}_{l,i} )   \text{ subject to }  \sum_{l \in \mathcal{N}_k}  a_{lk} = 1,    a_{lk}\geq0,  a_{lk}=0 \text{ if } l\not\in \mathcal{N}_k.
\end{equation}
Using the Lagrangian relaxation, we obtain the optimal solution\footnote{Detailed solution is given in Appendix \ref{app: optimal solution of weights}.} of \eqref{eq: final minimization problem} as
\begin{equation}\label{eq: optimal weight rule}
    a_{lk}(i) = \frac{{r_k({{\hat{\theta}}_{l,i}  })}^{-1}}{\sum_{p \in \mathcal{N}_k} {r_k({{\hat{\theta}}_{p,i} })}^{-1}}.
    % \hspace{1cm}  a_{lk}(i) = \frac{{ \varphi_{lk}^i}^{-1}}{\sum_{p \in \mathcal{N}_k} { \varphi_{pk}^i}^{-1}},
\end{equation}
% The optimal weight assignment rule meets the intuition. 
% It assigns a large weight to the neighbor who has a small loss on the ego agent's data since this implies the neighbor is constructing its model from a similar underlying distribution.
% Although there are other task similarity measurements in the literature, we find this loss based weight assignment rule is an easy approach to illustrate the problem with provable mathematical guarantees.
% Since $r_k( {\hat{\theta}}_{l,i})$ is not available during learning, we can approximate it using the smoothing method \cite{6232902}. And assign iterative weights as
% \begin{equation}\label{eq: approximate weight}
%     a_{lk}(i) = \frac{{ \varphi_{lk}^i}^{-1}}{\sum_{p \in \mathcal{N}_k} { \varphi_{pk}^i}^{-1}}.
% \end{equation}
We can approximate $r_k({{\hat{\theta}}_{l,i}  })$ using the exponential moving average 
$ \varphi_{lk}^i = (1 - \nu_k)  \varphi_{lk}^{i-1} + \nu_k \ell({{\hat{\theta}}_{l,i}  }; \xi_k),
$
where $\nu_k$ is the forgetting factor. 
%and 
%$\varphi_{lk}^{-1}$ can be initialized to be zero or a uniform positive constant for all $l$.
Given
$
   \mathbb{E} [ \varphi_{lk}^i] = (1 - \nu_k) \mathbb{E} [  \varphi_{lk}^{i-1}] + \nu_k \mathbb{E} [\ell({{\hat{\theta}}_{l,i}  }; \xi_k)],
$
we obtain $\lim_{i \rightarrow \infty} \mathbb{E} [ \varphi_{lk}^i] = \lim_{i \rightarrow \infty}  \mathbb{E} [\ell({{\hat{\theta}}_{l,i}  }; \xi_k)] =  \lim_{i \rightarrow \infty}  r_k(\hat{\theta}_{l,i} )$, \textcolor{black}{ which means}  $ \varphi_{lk}^i$ converges \textcolor{black}{(in expectation) to} $ \lim_{i \rightarrow \infty}  r_k(\hat{\theta}_{l,i} )$.
Hence, we can use $\varphi_{lk}^i$ to approximate
$r_k({{\hat{\theta}}_{l,i} })$.
Note that in addition to the smoothing methods, one can use the average batch loss to approximate  ${r_k({{\hat{\theta}}_{l,i} )}}$ when using the (mini-) batch gradient descent in \textcolor{black}{the place}  of SGD  for adaptation.

\subsection{Filtering for Resilience}
% \textbf{Filtering for Resilience.}  
Let $\mathcal{N}_k^+$ denote the set of $k$'s normal neighbors with $|\mathcal{N}_k^+| \geq 1$. 
We assume there are  $q$ Byzantine \textcolor{black}{neighbors in}   the set $\mathcal{B} = \mathcal{N}_k \backslash \mathcal{N}_k^+$.
In the following, we examine the resilience of the cooperation using \eqref{eq: optimal weight rule} in the presence of Byzantine agents.
% \begin{discussion}\footnote{Full discussion is given in the Appendix.}
% It is not sufficient to guarantee that a normal agent $k$ using \eqref{eq: optimal weight rule} for cooperation can converge resiliently to its unique minimizer $\theta_k^*$ with improved expected regret w.r.t. SGD.
% \end{discussion}
\begin{lemma}\label{lemma: expected regret for cooperation}
\footnote{
All proofs are given in Appendix \ref{app: Assumptions and Theoretical Results}; appendices can be found in the supplementary material.
}
\textcolor{purple}{The following condition holds for the combination step \eqref{eq: combine}}
using weights \eqref{eq: optimal weight rule}:
\begin{equation*}
% \label{eq: expected regret as sum of neighbor regret}
\begin{aligned}
\mathbb{E} \left[r_k(\theta_{k,i}  ) -r_k(\theta_{k}^*) \right] \leq  \frac{1}{ |\mathcal{N}_k|} \sum_{l \in \mathcal{N}_k} \mathbb{E} \left[r_k  \left(\hat{\theta}_{l,i}  \right) -r_k(\theta_{k}^*)\right].
\end{aligned}
\end{equation*}
\end{lemma}
\vspace{-0.2cm}
Since $l$ can be a Byzantine agent, it is possible that $\mathbb{E} \left[r_k  \left(\hat{\theta}_{l,i}  \right) -r_k(\theta_{k}^*)\right]$ is a large value. \textcolor{black}{Consequently, we cannot compute a useful upper bound on}  the value \textcolor{black}{of} $\mathbb{E} \left[r_k(\theta_{k,i}  ) -r_k(\theta_{k}^*) \right]$ given Lemma \ref{lemma: expected regret for cooperation} and cannot provide further convergence guarantees.
To facilitate the resilient cooperation, we consider a modification of \eqref{eq: optimal weight rule} \textcolor{black}{as follows}.  
\begin{equation}\label{eq: filtering weight}
     a_{lk}(i) =  \begin{cases} 
    \frac{{r_k({{\hat{\theta}}_{l,i} })}^{-1}}{\sum_{p \in \mathcal{N}^{\leq}_k} {r_k({{\hat{\theta}}_{p,i} })}^{-1}}, &\hspace{-0.2cm} \text{if } r_k({{\hat{\theta}}_{l,i} }) \leq r_k({{\hat{\theta}}_{k,i} }), \\
    0,  &\hspace{-0.2cm} \text{otherwise,}
    \end{cases}
    % a_{lk}(i) = 
    % \begin{cases}
    % \frac{{ \varphi_{lk}^i}^{-1}}{\sum_{p \in \mathcal{N}^{\leq}_k} { \varphi_{pk}^i}^{-1}},  &\hspace{-0.2cm} \text{if }  \varphi_{lk}^i \leq  \varphi_{kk}^i \\
    % 0,  &\hspace{-0.2cm} \text{otherwise. }
    % \end{cases}
\end{equation}
where $\mathcal{N}^{\leq}_k$ denotes the set of  neighbors  with $r_k({{\hat{\theta}}_{l,i} }) \leq r_k({{\hat{\theta}}_{k,i} })$.
% and $ \varphi_{lk}^i \leq  \varphi_{kk}^i$, for the two formulations respectively.
This implies that the cooperation filters out the information coming from the  neighbors incurring a larger risk 
% or $ \varphi$ value 
 and cooperate only with the remaining neighbors. In the next section, we show how this modification \textcolor{purple}{benefits learning and} guarantees the resilient convergence of MTL.
 
%  with an improved learning performance \textcolor{black}{as} measured by the expected regret w.r.t. the non-cooperative case.

\subsection{Computational Complexity} 
% \textbf{Computational Efficiency.} 
% It takes  $\mathcal{O}(d_x)$ time to compute the predicted value $f_{l}(x_k^i)$ using the model ${\hat{\theta}}_{l,i} $ and data $x_k^i$.
% \textcolor{black}{Similarly,} 
% It takes $\mathcal{O}(d_y)$ time to compute the loss from $f_{l}(x_k^i)$ and $y_k^i$. \textcolor{black}{ Hence,}  i
It takes $\mathcal{O}(d_x + d_y)$ time to compute $\ell({{\hat{\theta}}_{l,i}  }; \xi_k^i)$.
\textcolor{black}{Using the exponential moving average method for approximating $r_k({{\hat{\theta}}_{l,i}  })$,} for a normal agent $k$, at each iteration $i$, the total time for computing $A_k(i)$ with the proposed rule \eqref{eq: filtering weight} \textcolor{black}{is}  $\mathcal{O}(|\mathcal{N}_k|(d_x+d_y))$.

%% file: resilience.tex
\section{Byzantine Resilient Convergence Analysis}\label{sec: resilient convergence analysis}
We make the following \textcolor{purple}{general assumptions for the convergence of SGD \cite{DBLP:journals/siamrev/BottouCN18} to derive our results}.
\textcolor{purple}{
\begin{assumption}
For every normal agent $k$, the risk function $r_k(\cdot)$ is \emph{$m$-strongly convex} and has \emph{$L$-Lipschitz continuous gradient}.\footnote{Details of the assumptions about the risk functions are given in Appendix \ref{app: assumptions}.}
\end{assumption}
\begin{assumption}
For every normal agent $k$, the stochastic gradient $\nabla  \ell({\theta_{k,i}}; \xi_{k}^i)$ is an unbiased estimate of  $\nabla r_k(\theta_{k,i})$, i.e., $\mathbb{E} [\nabla  \ell({\theta_{k,i}}; \xi_k^i)] =  \nabla r_k ({\theta_{k,i}})$, for all $i \in \mathbb{N}$.
\end{assumption}
\begin{assumption}
For every normal agent $k$, there exists $c_k \geq 1$, such that for all $i \in \mathbb{N}$, $\mathbb{E}[\|\nabla  \ell({\theta_{k,i}}; \xi_{k}^i)\|_2^2] \leq \sigma_k^2 + c_k\|\nabla r_k(\theta_{k,i})\|_2^2$.
\end{assumption}
% \begin{assumption}
% The step-sizes $\mu_{k,i}$ used by every normal agent $k$ satisfies $\sum_{i=1}^{\infty} \mu_{k,i} = \infty, \sum_{i=1}^\infty \mu_{k,i}^2 \leq \infty$.
% \end{assumption}
}

Given these assumptions, \textcolor{purple}{the convergence of a normal agent running SGD is guaranteed with appropriate step size \cite{DBLP:journals/siamrev/BottouCN18}. Using the proposed rule \eqref{eq: filtering weight}, under these assumptions, we further guarantee the convergence of the normal agents running the ATC diffusion algorithm in Theorem \ref{theorem1}}.

\begin{theorem}\label{theorem1}
\textcolor{purple}{
A normal agent $k$ which runs the ATC diffusion  algorithm using the loss-based weights \eqref{eq: filtering weight} 
converges towards $\theta_{k}^*$ with $\lim_{i \rightarrow \infty }\mathbb{E}
           \left[r_k \left({\theta}_{k,i} \right) - r_k(\theta_k^*) \right] \leq  \frac{\mu_k L \sigma_k^2}{2 m}$, for fixed stepsize $\mu_k \in (0, \frac{1}{Lc_k}]$, in the presence of an arbitrary number of  Byzantine neighbors. Further, it holds that 
% \textcolor{purple}{
% A normal agent $k$ which runs the ATC diffusion  algorithm using the loss-based weights \eqref{eq: filtering weight} 
% converges in the presence of an arbitrary number of  Byzantine neighbors.
% % for fixed stepsize $\mu \in (0, \frac{2}{L}]$,
% And the expected regret at the convergence point satisfies
%If a normal agent $k$ runs the cooperative SGD algorithm using the loss-based weights \eqref{eq: filtering weight}, then for fixed stepsizes $\mu_k = \mu \in (0, \frac{2}{L}], \forall k \in \mathcal{N}^+$,
%it converges in the presence of arbitrary number of  Byzantine neighbors, and the expected regret at the convergence point satisfies
% \vspace{-0.1cm}
% \begin{equation*}\label{eq: loss-based convergence regret}
%  \textcolor{purple}{
%     \lim_{i \rightarrow \infty} R_{k}^{\text{\rm(coop)}}(i) 
%     =
%     \lim_{i \rightarrow \infty}  \mathbb{E}  \left[r_k \left({\theta}_{k,i}^{\text{\rm(coop)}}  \right) - r_k^* \right] 
%     \leq
%     \frac{\mu L}{2m(2 - \mu L)} \frac{1}{|\mathcal{N}^{\leq}_k|}  \sum_{l \in \mathcal{N}^{\leq}_k} \sigma_l^2.
%     }
% \end{equation*}
% \textcolor{black}{Furthermore, in the presence of arbitrary number of Byzantine neighbors, we have}
% $$
% \textcolor{purple}{
% % \label{eq: expected regret as sum of neighbor regret}
% \textcolor{black}{\lim_{i \rightarrow \infty}  R_{k}^{\text{\rm(coop)}}(i) \leq \lim_{i \rightarrow \infty} R_{k}^{\text{\rm(ncop)}}(i).}
% }
% $$
$
\textcolor{purple}{
% R_{k}(\theta_{k,i}) \leq   R_{k}(\hat{\theta}_{k,i})
 \mathbb{E} [r_k(\theta_{k,i}) - r_k(\theta_k^*)] \leq 
        \mathbb{E} [r_k(\hat{\theta}_{k,i}) - r_k(\theta_k^*)]
, \forall k \in \mathcal{N}^+, i \in \mathbb{N}.
}
$
}
\end{theorem}
% \vspace{-0.5cm}
\textcolor{black}{Theorem \ref{theorem1}
indicates that cooperation using weights \textcolor{red}{in} \eqref{eq: filtering weight} is always at least as good as the non-cooperative case, as measured by the expected regret, which satisfies the conditions lsited in \textcolor{purple}{Section 4}. Note that even when all the neighbors of a normal agent are Byzantine, one can still guarantee that \textcolor{red}{the agent's} learning performance \textcolor{red}{as a result of} cooperation with neighbors using \eqref{eq: filtering weight} will be same as the non-cooperative case.
}

\textcolor{blue}{
\textbf{Discussion.}
We assume convex models \textcolor{red}{to carry out the analysis, which is typical in \textcolor{green}{the literature}. However,} %in our analysis, which are typically assumed in the ML literature for convergence analysis.
%Although the analysis is based on convex models, 
the intuition behind \textcolor{red}{the approach} \textcolor{green}{is} --- \emph{to measure the relatedness of a neighbor to itself, a normal agent \textcolor{red}{evaluates} the loss of the neighbor using the neighbor's model parameters and its own data, and \textcolor{red}{cuts} down the cooperation if \textcolor{red}{this loss is larger than the agent's own loss }}%the loss of the neighbor is larger than the loss of itself}
 --- \textcolor{red}{and the same idea should also apply} to non-convex models.
In the next section, we also evaluate our methods on non-convex models, such as CNNs, which generates \textcolor{red}{experimental results similar to those produced by convex models.}% consistent experimental results to those of the convex models. 
}

%% file: evaluation_new.tex
\section{Evaluation}
In this section, we evaluate the resilience of the proposed online weight adjustment rule \eqref{eq: filtering weight} with the smoothing  method discussed in Section \ref{sec: loss-based weight},  
and compare it with the non-cooperative case, the  average weights ($a_{lk} = \frac{1}{|\mathcal{N}_k|}$), and the quadratic distance-based weights \eqref{eq: quadratic distance weight} (with $\tilde{\theta}_k^* = \theta_{k,i-1}$ and use the same smoothing method $ \phi_{lk}^i = (1 - \nu_k)  \phi_{lk}^{i-1} + \nu_k \|\tilde{\theta}_k^* - \hat{\theta}_{l,i}\|^{2}
$ in the place of $\|\tilde{\theta}_k^* - \hat{\theta}_{l,i}\|^{2}$, with the same forgetting factor $\nu_k$ used for \eqref{eq: filtering weight}). 
\textcolor{black}{We use} three  distributed MTL case studies, including the  regression and classification problems, with and without the presence of Byzantine agents. 
Although the convergence analysis in Section \ref{sec: resilient convergence analysis} is based on convex models and SGD, we show \textcolor{black}{empirically} that the weight assignment rule \eqref{eq: filtering weight}  performs  well for non-convex models, such as CNNs and mini-batch gradient descent. 
\textcolor{blue}{
Our code is available at  \url{https://github.com/JianiLi/resilientDistributedMTL}.
}

\begin{figure}[ht]
    \centering
\begin{minipage}{0.245\textwidth}
\begin{subfigure}{1\textwidth}
  \centering
  % include first image
  \includegraphics[width=1\linewidth, trim=1.5cm 1cm 1.5cm -0.2cm]{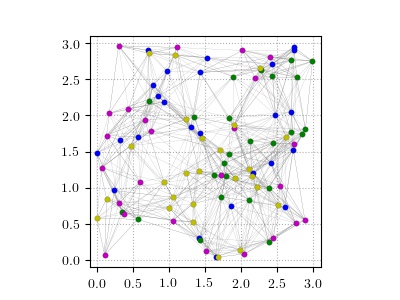} 
     \vspace{0.1cm}
  \caption{Network topology}
  \label{fig: target localization network}
\end{subfigure}
%   \label{fig:multi-task}
\end{minipage}
\begin{minipage}{0.745\textwidth}
\begin{minipage}{1\textwidth}
\centering
\vspace{-1.5cm}
    \includegraphics[width=0.33\linewidth, angle=270, trim=5cm 10cm 5cm 10cm]{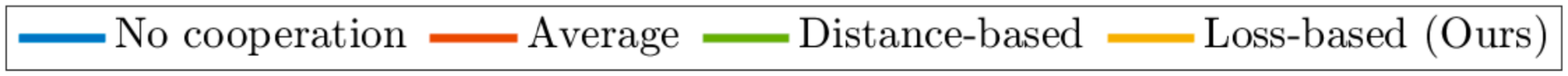}
\vspace{-1.4cm}
\end{minipage}\\
\vspace{0cm}
\begin{minipage}{1\textwidth}
\begin{subfigure}{.325\textwidth}
  \centering
  % include first image
  \vspace{0.2cm}
  \includegraphics[width=1\linewidth, trim=1cm 0cm 0cm 1cm]{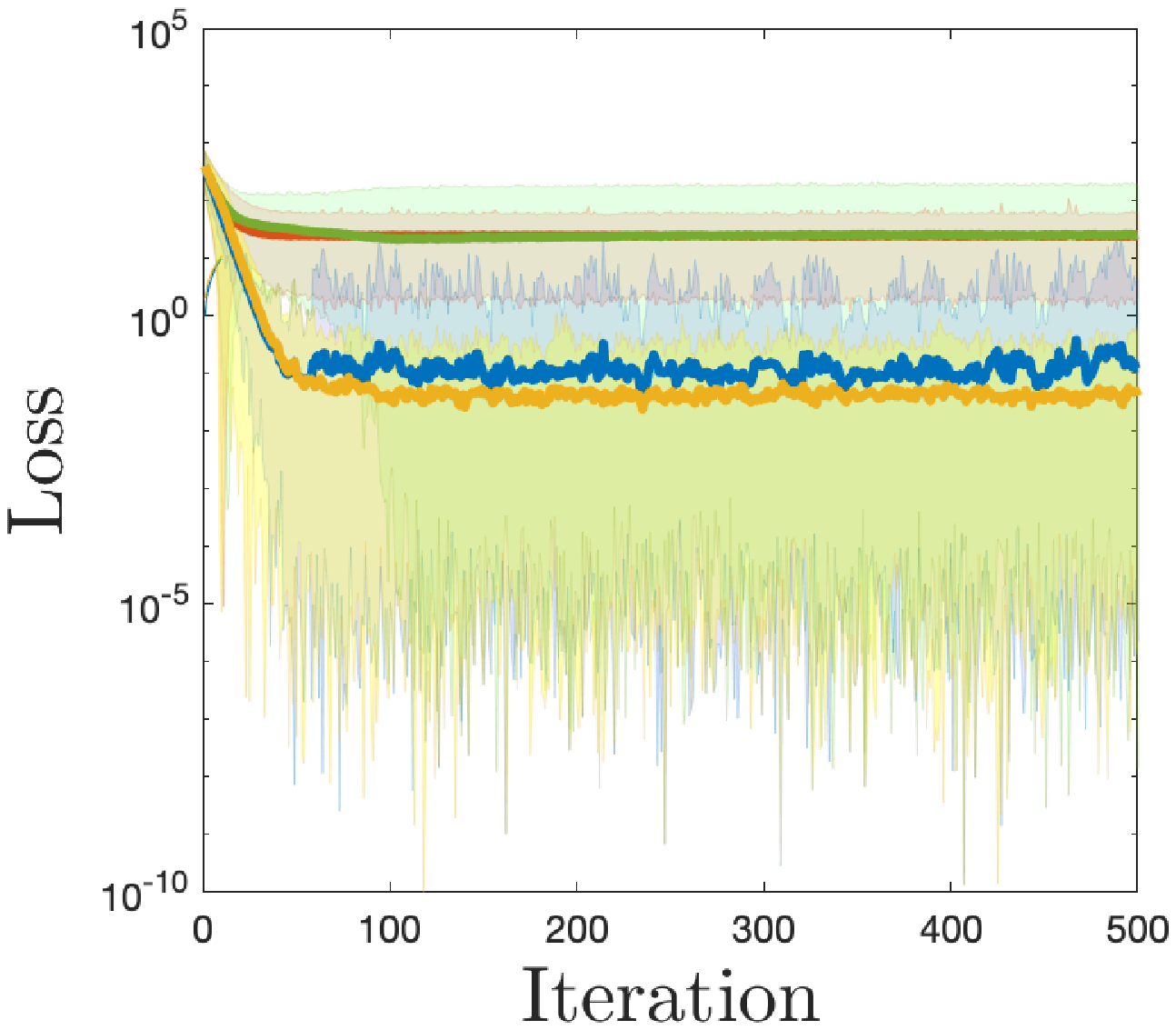}  
  \caption{No attack}
  \label{fig: TL, no attack}
\end{subfigure}
\begin{subfigure}{.325\textwidth}
  \centering
  % include second image
    \vspace{0.2cm}
  \includegraphics[width=1\linewidth, trim=1cm 0cm 0cm 1cm]{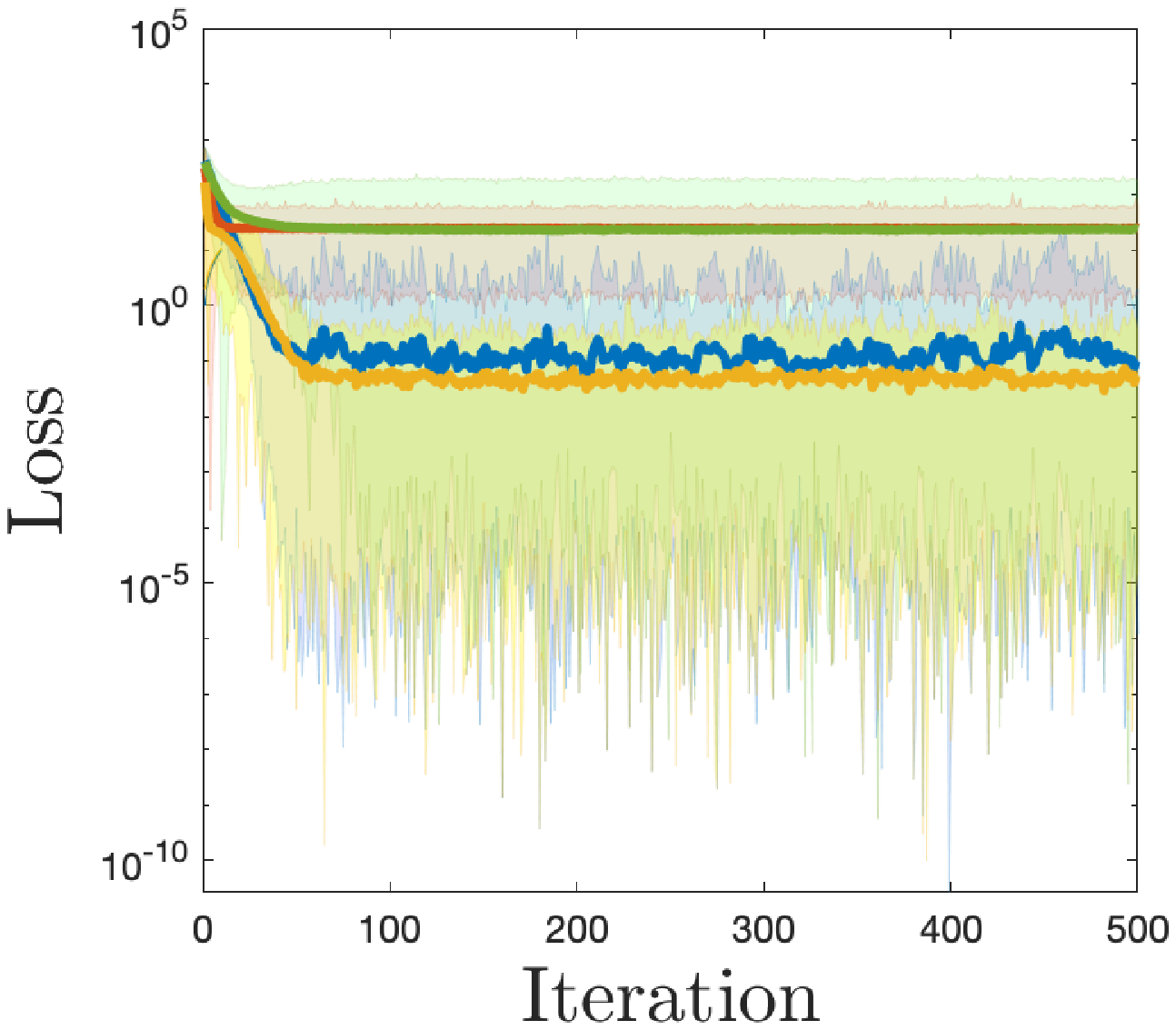}  
  \caption{20 Byzantine agents}
%   \label{fig:Byzantine}
\end{subfigure}
\begin{subfigure}{.325\textwidth}
  \centering
    \vspace{0.2cm}
  % include second image
  \includegraphics[width=1\linewidth, trim=1cm 0cm 0cm 1cm]{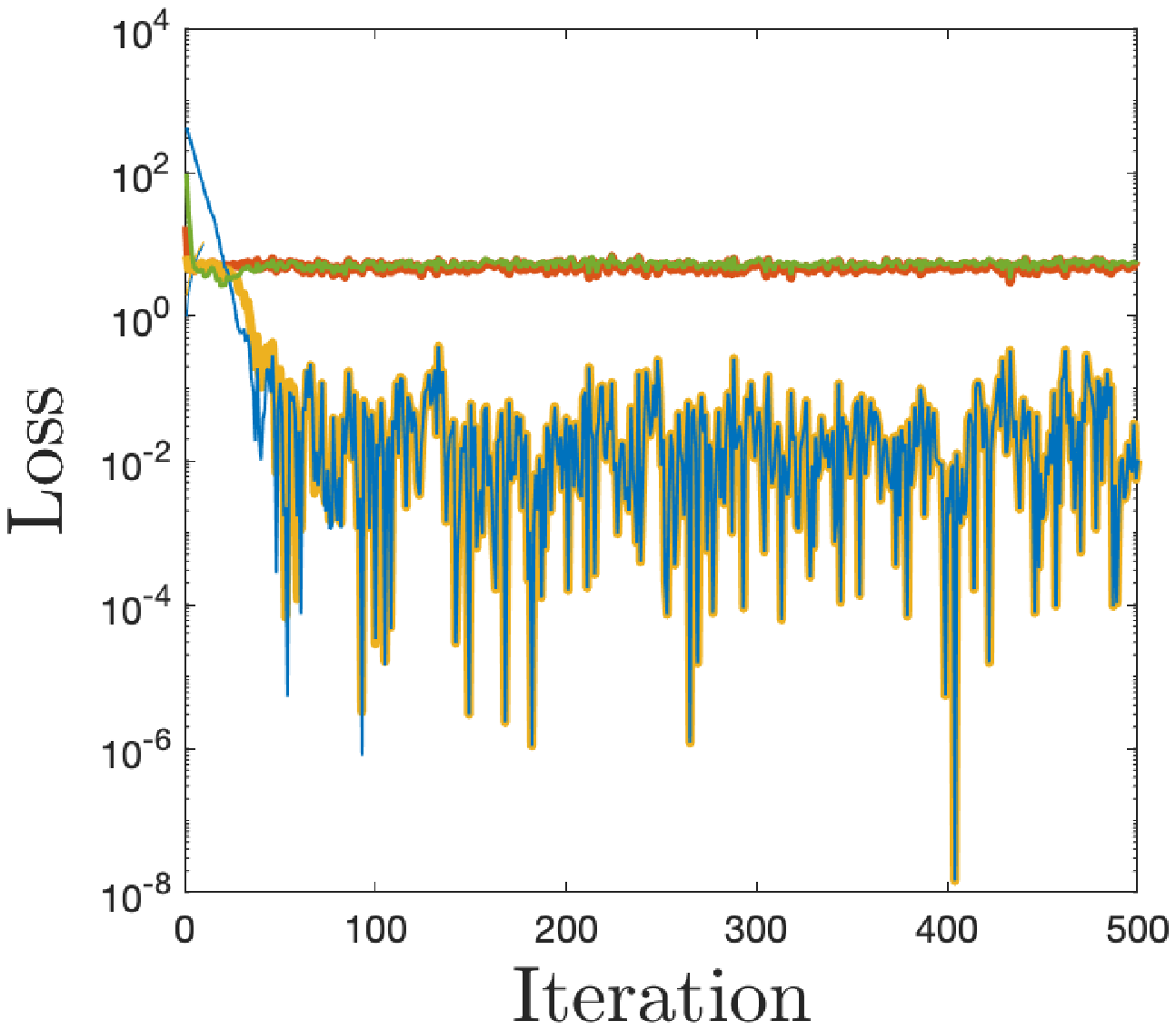}  
  \caption{99 Byzantine agents}
   \label{fig: TL, 99 attackers}
\end{subfigure}
\end{minipage}%
\end{minipage}%
\caption{Target Localization: network topology and  loss of streaming data for normal agents.}
\label{fig: Target Localization: Network topology and average Loss for streaming data}
\end{figure}

% \subsection{Human Activity Recognition}\label{sec: human action recognition}
% \vspace{-0.2cm}

\begin{figure}[ht]
\centering
% \vspace{0.1cm}
\begin{subfigure}{0.495\textwidth}
  \centering
  \begin{minipage}{.48\textwidth}
  \centering
    \includegraphics[width=1\linewidth, trim=1cm 0cm 0cm 1cm]{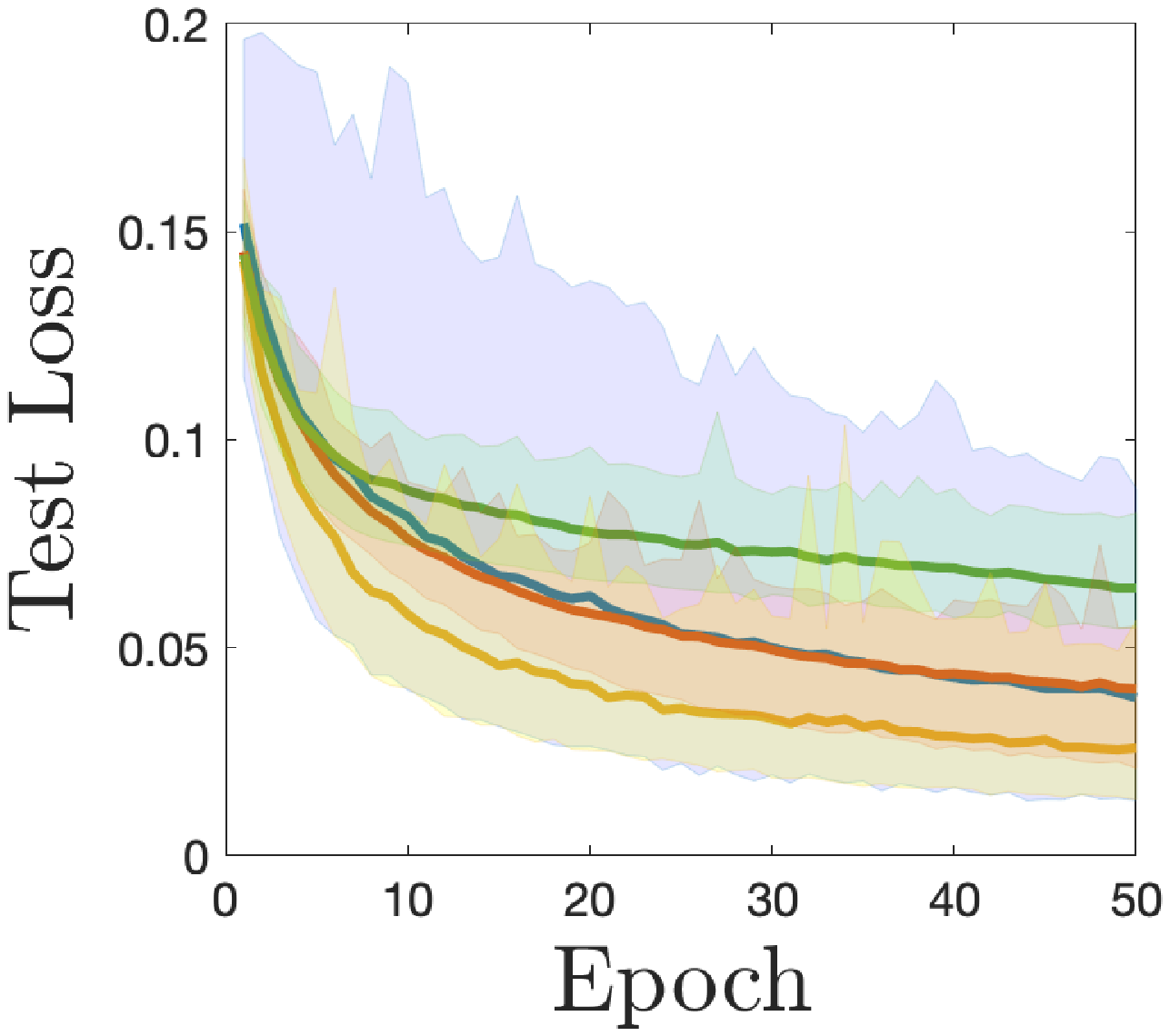}  
  \end{minipage}
    \begin{minipage}{.48\textwidth}
  \centering
    \includegraphics[width=1\linewidth, trim=1cm 0cm 0cm 1cm]{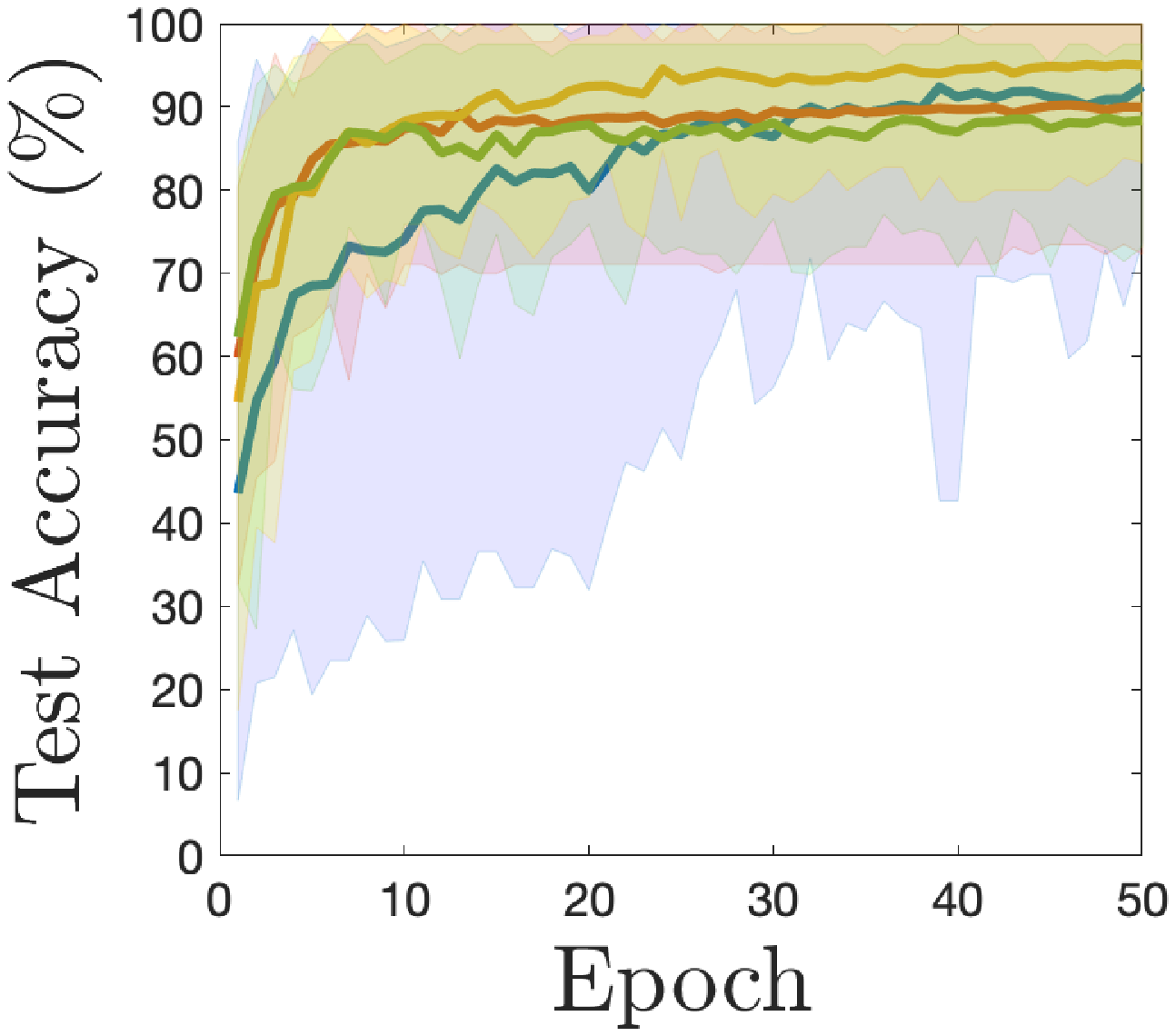}  
  \end{minipage}
  \vspace{-0.2cm}
\caption{No attack}
\end{subfigure}
\begin{subfigure}{0.495\textwidth}
  \centering
  \begin{minipage}{.48\textwidth}
  \centering
    \includegraphics[width=1\linewidth, trim=1cm 0cm 0cm 1cm]{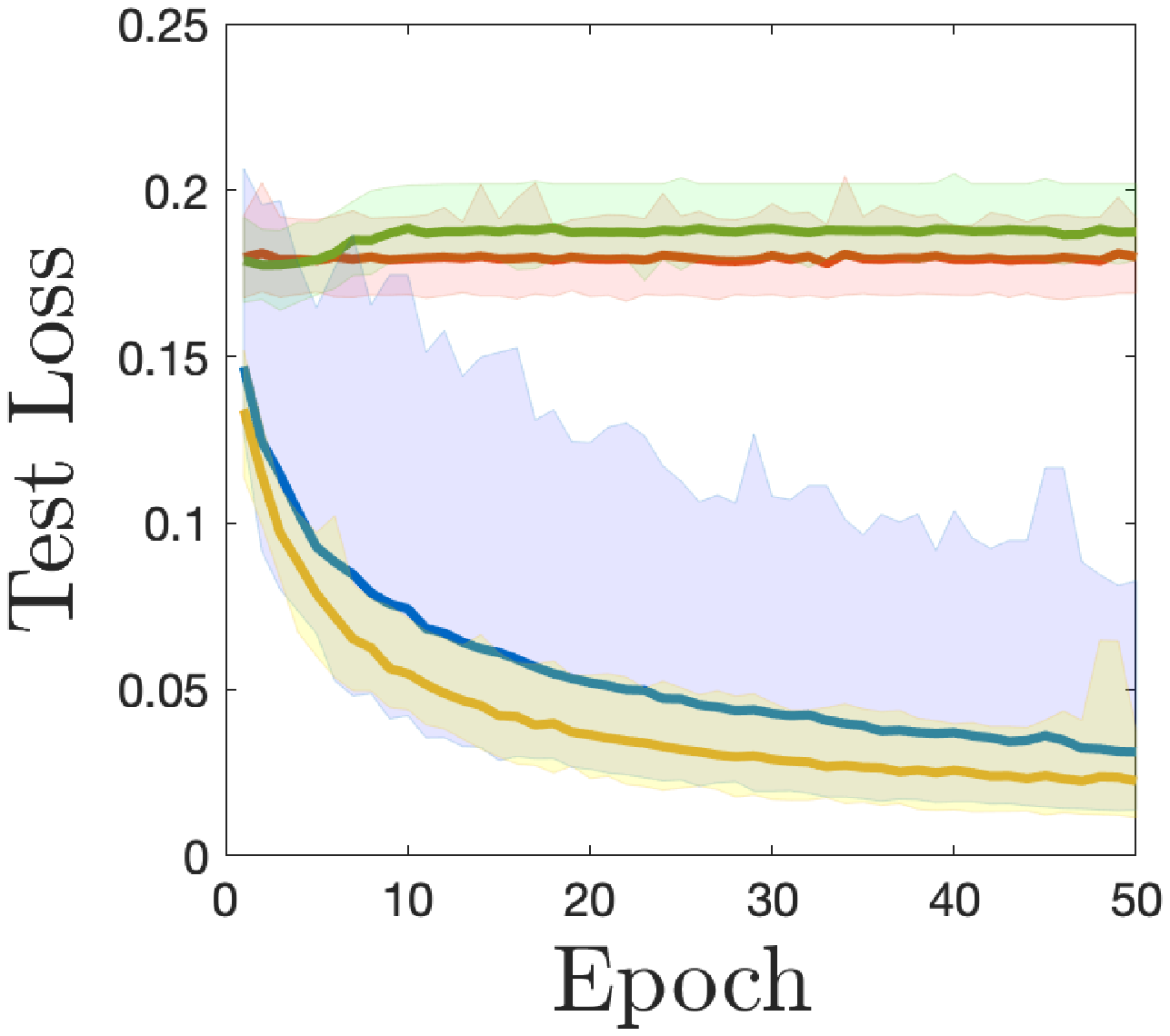}  
  \end{minipage}
  % include first image
  \begin{minipage}{.48\textwidth}
  \centering
  % include first image
  \includegraphics[width=1\linewidth, trim=1cm 0cm 0cm 1cm]{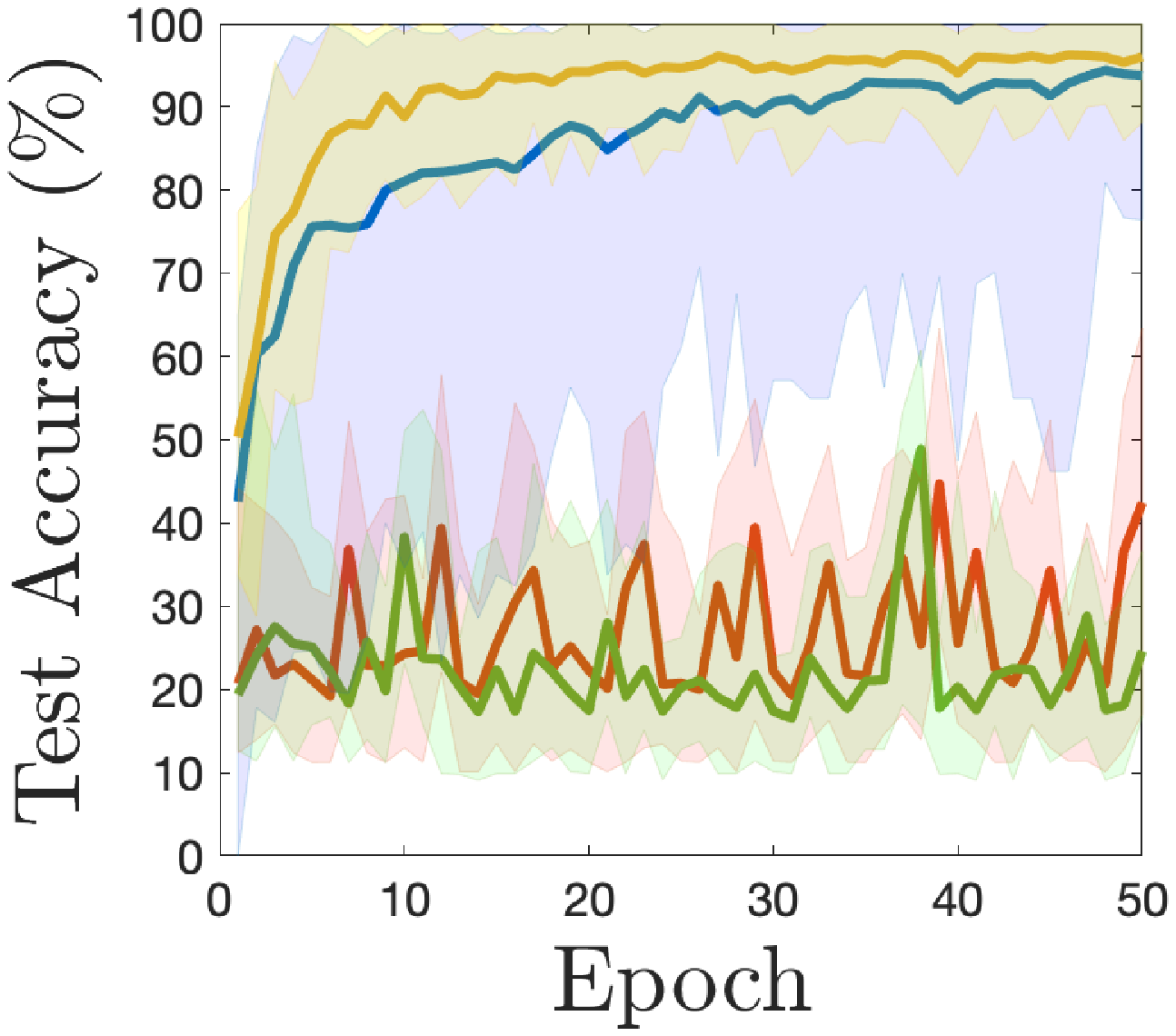}  
%   \label{fig:multi-task}
\end{minipage}
  \vspace{-0.2cm}
\caption{10 Byzantine agents}
\end{subfigure}
% \begin{subfigure}{0.325\textwidth}
%   \centering
%   \begin{minipage}{.48\textwidth}
%   \centering
%     \includegraphics[width=1\linewidth, trim=1cm 0cm 0cm 1cm]{figure/HAR_range_loss_test_attack_99.eps}  
%   \end{minipage}
%   % include first image
%   \begin{minipage}{.48\textwidth}
%   \centering
%   % include first image
%   \includegraphics[width=1\linewidth, trim=1cm 0cm 0cm 1cm]{figure/HAR_range_acc_test_attack_99.eps}  
% %   \label{fig:multi-task}
% \end{minipage}
% \caption{29 Byzantine agents}
% \end{subfigure}
\caption{Human Action Recognition: average testing loss and  accuracy for normal agents.}
\label{fig: har test}
\end{figure}

\vspace{0.3cm}

\subsection{Datasets and Simulation Setups}
\begin{itemize}[leftmargin=*]
\item \textbf{Target Localization}: Target localization is a widely-studied linear regression problem \cite{DBLP:conf/icassp/ChenRS14}.
The task is to estimate the location of the target by minimizing the squared error loss of  noisy streaming sensor data.
We consider a network of 100 agents with four targets as shown in \figref{fig: target localization network}. 
Agents in the same color share the same target, \textcolor{black}{however,}  they do not know this group information beforehand.

    \item \textbf{Human Activity Recognition\footnote{\url{https://archive.ics.uci.edu/ml/datasets/human+activity+recognition+using+smartphones}}}:
    Mobile phone sensor data (accelerometer and gyroscope) is collected from 30 individuals   performing one of six activities: \{walking, walking-upstairs, walking-downstairs, sitting, standing, lying-down\}. 
The goal is to predict the activities performed using 561-length feature vectors for each instance generated by the processed sensor signals \cite{DBLP:conf/esann/AnguitaGOPR13}.
We model each individual as a separate task and use a complete graph to model the network topology.
We use linear model as the prediction function with cross-entropy-loss.

 \item \textbf{Digit Classification}:
We consider a network of ten agents  performing digit classification.
Five of the ten agents \textcolor{black}{have}   access to the MNIST dataset\footnote{\url{http://yann.lecun.com/exdb/mnist}}  \cite{lecun-mnisthandwrittendigit-2010} (group 1) and the other five \textcolor{black}{have}  access to the synthetic  dataset\footnote{\url{https://www.kaggle.com/prasunroy/synthetic-digits}} (group 2) that is composed by generated images of digits embedded on random backgrounds  \cite{roy2018effects}.
All the images are preprocessed to be $28 \times 28 $ grayscale images.
We model each agent as a separate task and use a complete graph to  model the network topology.
\textcolor{black}{An agent does not} know which of its neighbors are performing the same task as \textcolor{black}{the agent} itself.
We use a CNN model of the same architecture for each agent and cross-entropy-loss.
% We classify the digits $0-9$ using the preprocessed images.
% \figref{fig: Examples of the Digit classification dataset} in Appendix are some preprocessed examples of the two datasets.
% We refer to the 5 agents for the MNIST dataset as group 1 and the other 5 agents for the synthetic digits dataset as group 2. 
% (or which of the neighbors are access to the same dataset as itself).
%Since some of the agents are access to limited training data, they are prone to overfitting or learn slowly if without cooperation.
% and use mini-batch gradient descent with batch size of $64$ and step-sizes (learning rate) $\mu_k = 0.001$ and forgetting factor $\nu_k = 0.05$ for all the normal agents.
\end{itemize}

\subsection{Results\footnote{Simulation details and supplementary results are given in Appendix \ref{app: Simulation Details and Supplementary Results}.}}
\begin{figure}[H]
\centering
% \begin{minipage}{1\textwidth}
% \centering
%     \includegraphics[width=0.7\linewidth, trim=0cm -0.4cm -1cm 0cm]{figure/TL_legend_new.jpg}
% \end{minipage}\\
% \vspace{0.1cm}
\begin{subfigure}{0.495\textwidth}
  \centering
  \begin{minipage}{.48\textwidth}
  \centering
    \includegraphics[width=1\linewidth, trim=1cm 0cm 0cm 1cm]{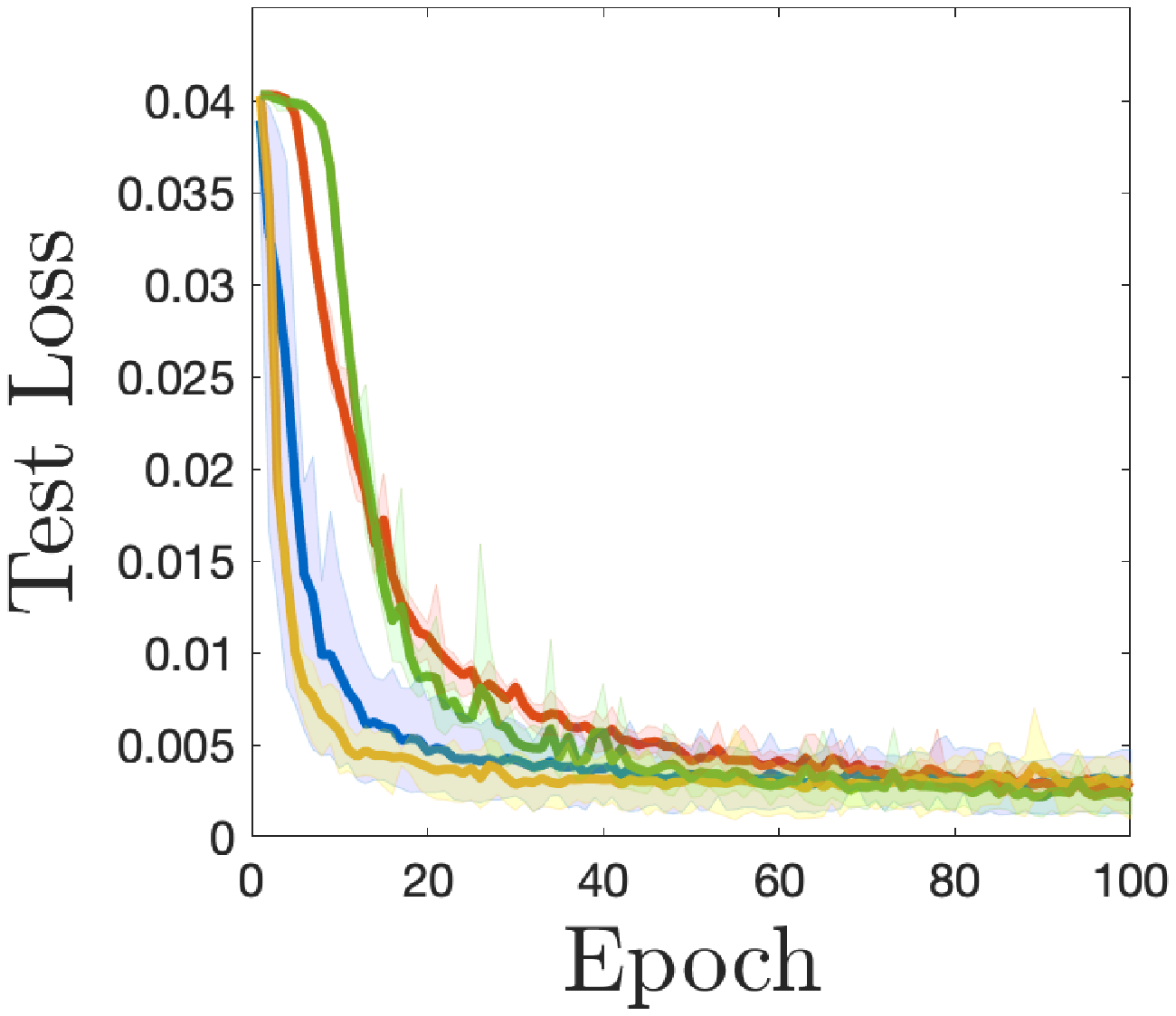}  
  \end{minipage}
  % include first image
  \begin{minipage}{.48\textwidth}
  \centering
  % include first image
  \includegraphics[width=1\linewidth, trim=1cm 0cm 0cm 1cm]{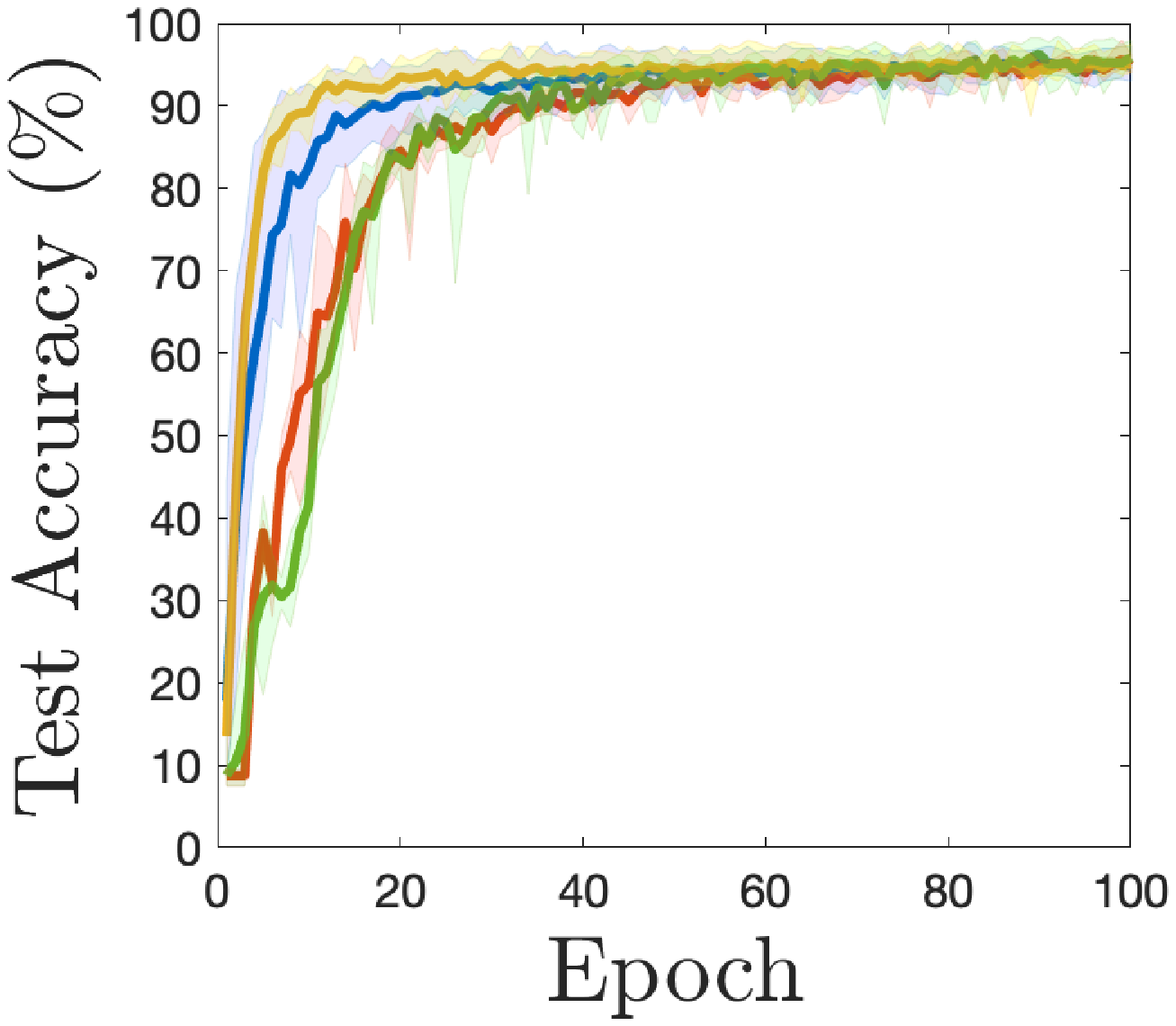}  
%   \label{fig:multi-task}
\end{minipage}
\vspace{-0.2cm}
\caption{No attack}
\end{subfigure}
\begin{subfigure}{0.495\textwidth}
  \centering
  \begin{minipage}{.48\textwidth}
  \centering
    \includegraphics[width=1\linewidth, trim=1cm 0cm 0cm 1cm]{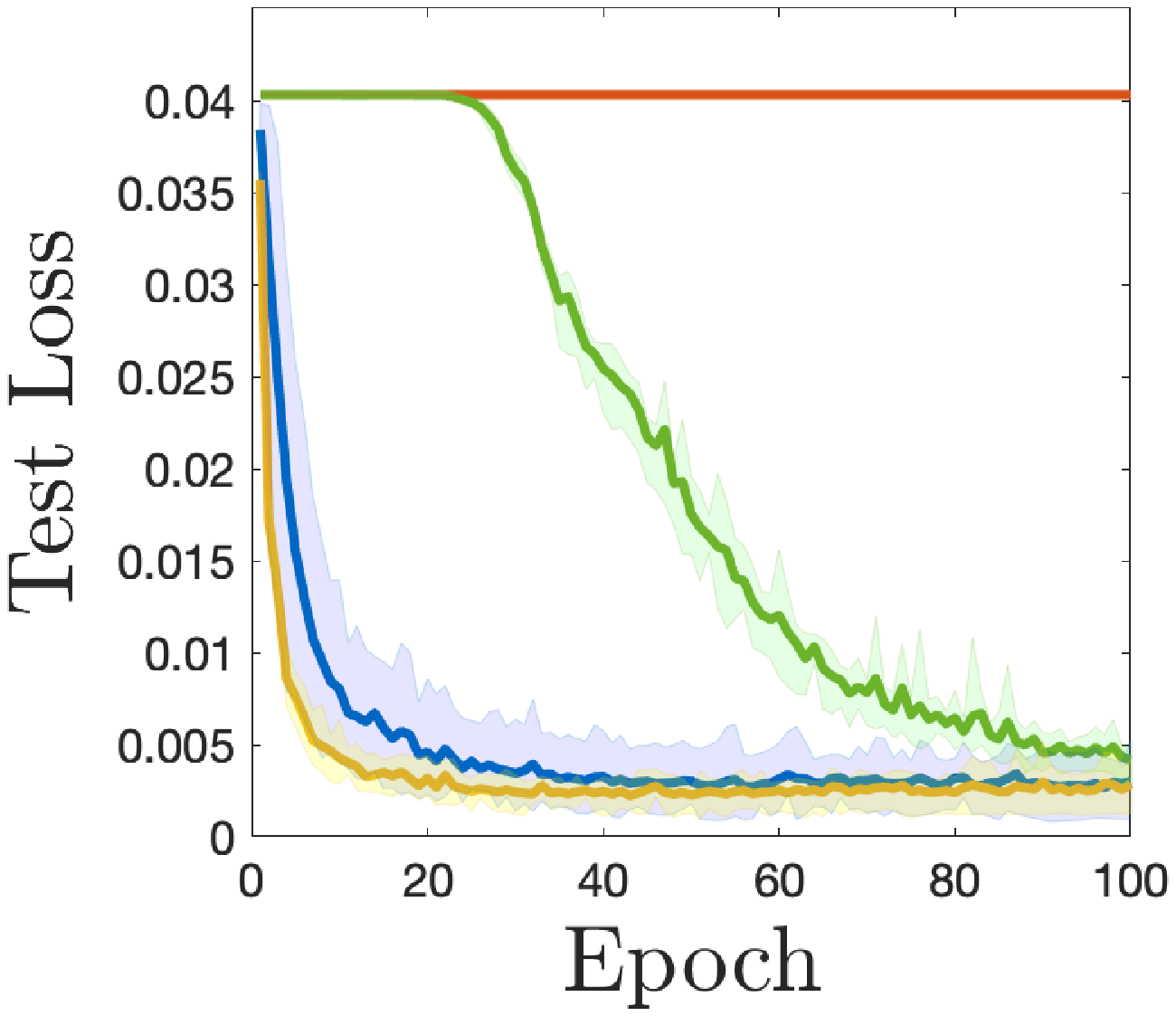}  
  \end{minipage}
  % include first image
  \begin{minipage}{.48\textwidth}
  \centering
  % include first image
  \includegraphics[width=1\linewidth, trim=1cm 0cm 0cm 1cm]{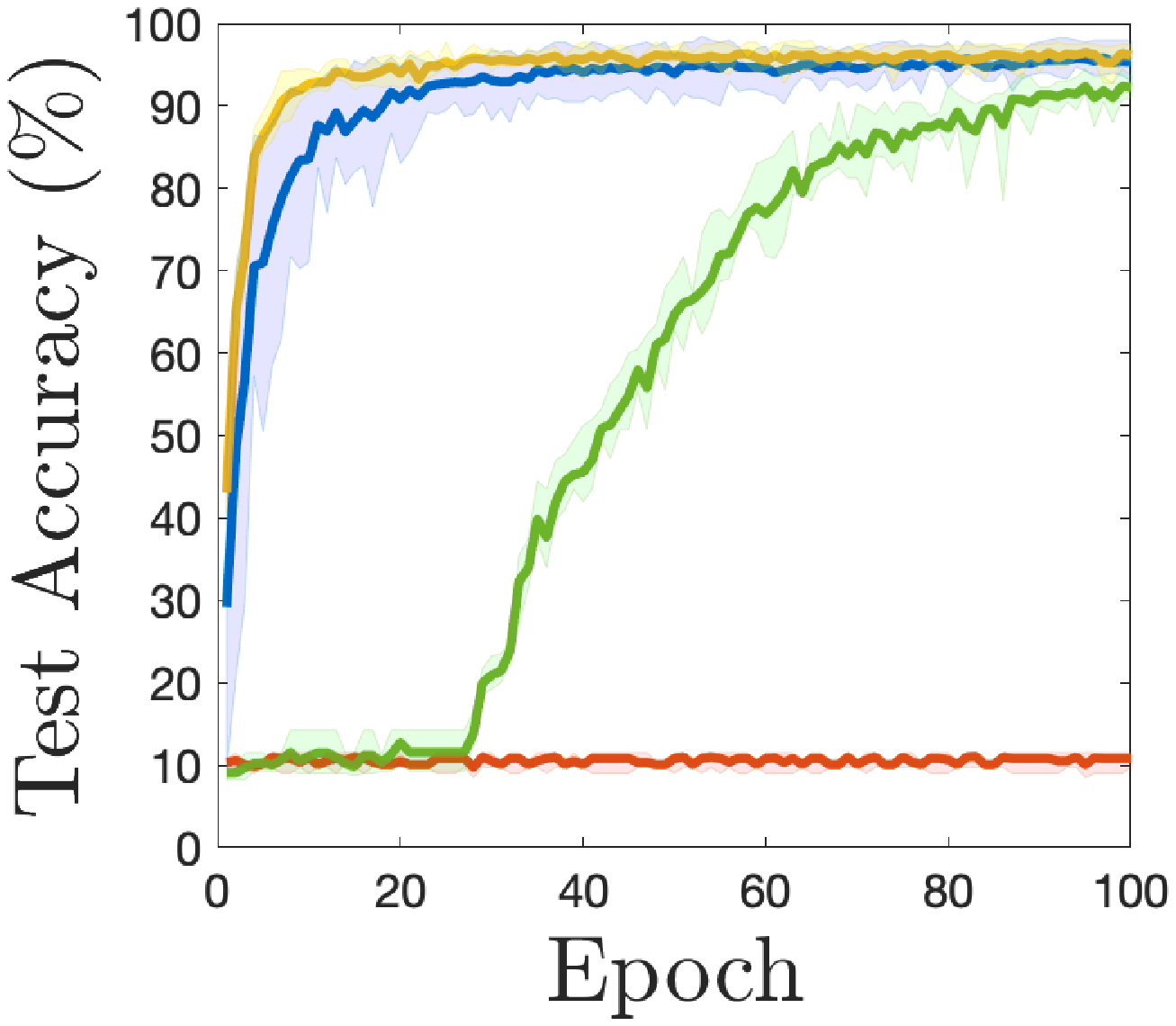}  
%   \label{fig:multi-task}
\end{minipage}
\vspace{-0.2cm}
\caption{2 Byzantine agents}
\end{subfigure}
% \begin{subfigure}{0.325\textwidth}
%   \centering
%   \begin{minipage}{.48\textwidth}
%   \centering
%     \includegraphics[width=1\linewidth, trim=1cm 0cm 0cm 1cm]{figure/mnist_range_loss_test_attack_8.eps}  
%   \end{minipage}
%   % include first image
%   \begin{minipage}{.48\textwidth}
%   \centering
%   % include first image
%   \includegraphics[width=1\linewidth, trim=1cm 0cm 0cm 1cm]{figure/mnist_range_acc_test_attack_8.eps}  
% %   \label{fig:multi-task}
% \end{minipage}
% \caption{8 Byzantine agents}
% \end{subfigure}
\caption{Digit Classification: average testing loss and  accuracy for normal agents in group 1.}
\label{fig: MNist test}
\end{figure}

\begin{figure}[H]
\centering
% \begin{minipage}{1\textwidth}
% \centering
%     \includegraphics[width=0.7\linewidth, trim=0cm -0.4cm -1cm 0cm]{figure/TL_legend_new.jpg}
% \end{minipage}\\
% \vspace{0.1cm}
\begin{subfigure}{0.495\textwidth}
  \centering
  \begin{minipage}{.48\textwidth}
  \centering
    \includegraphics[width=1\linewidth, trim=1cm 0cm 0cm 1cm]{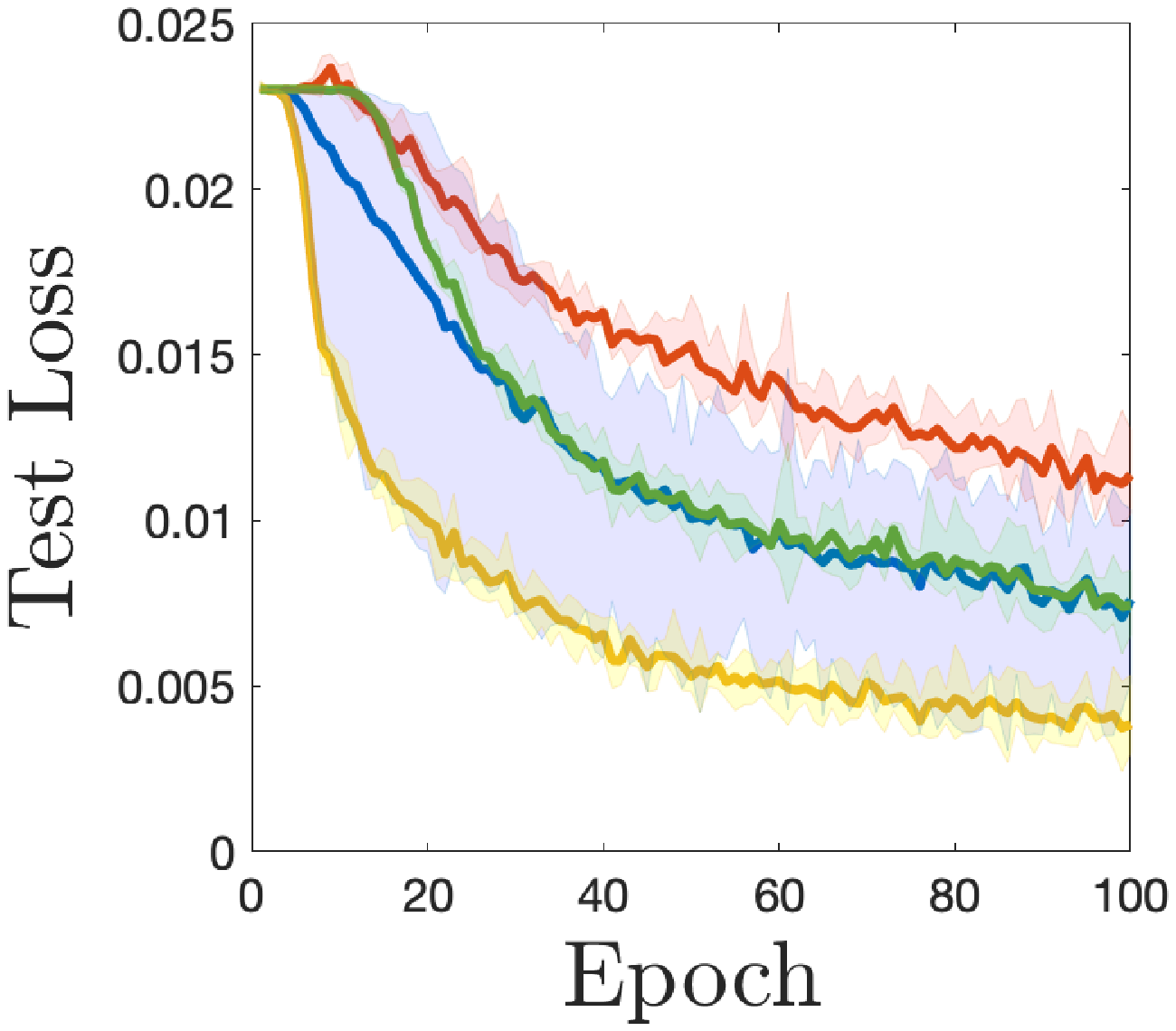}  
  \end{minipage}
  % include first image
  \begin{minipage}{.48\textwidth}
  \centering
  % include first image
  \includegraphics[width=1\linewidth, trim=1cm 0cm 0cm 1cm]{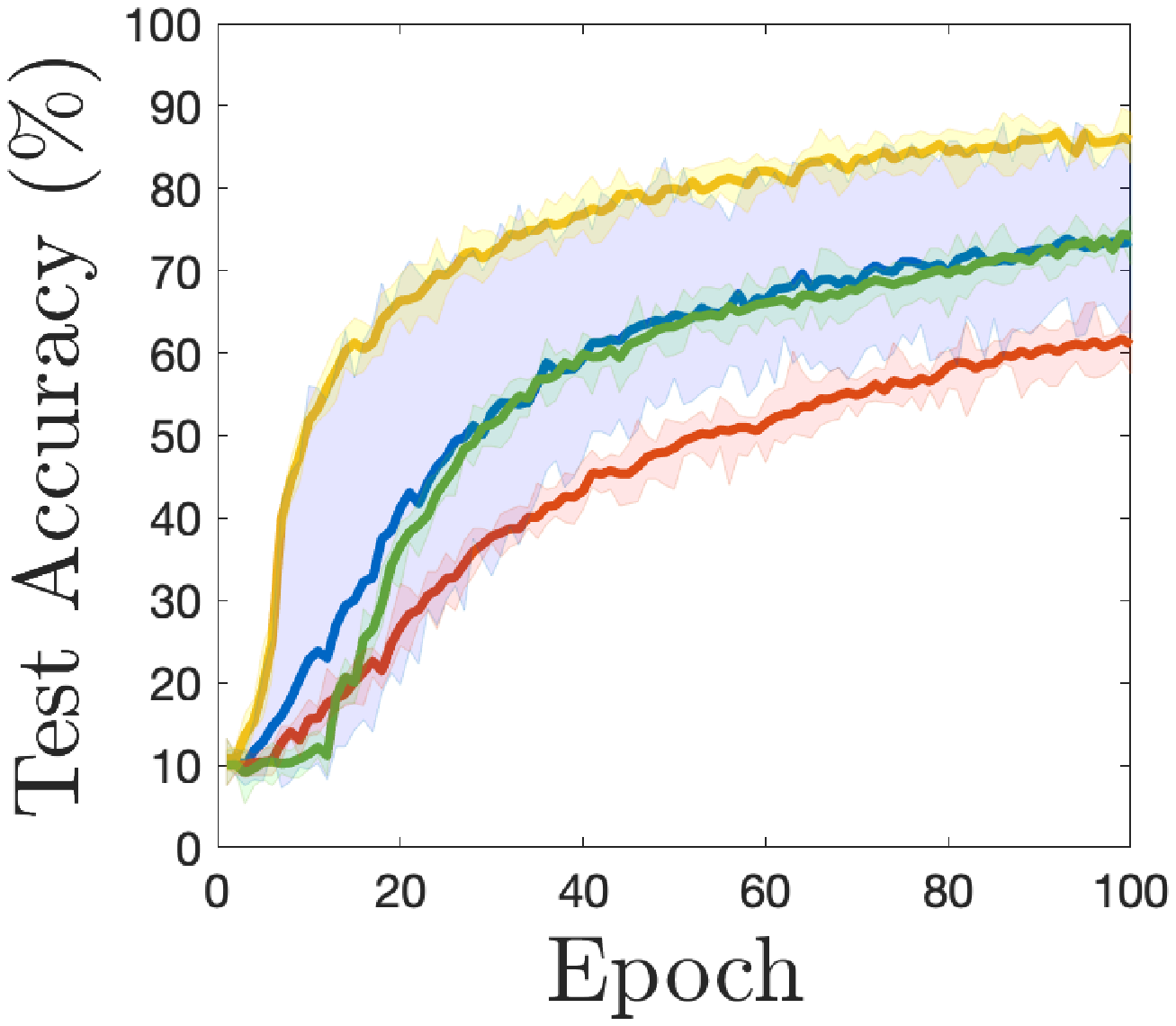}  
%   \label{fig:multi-task}
\end{minipage}
\vspace{-0.2cm}
\caption{No attack}
\end{subfigure}
\begin{subfigure}{0.495\textwidth}
  \centering
  \begin{minipage}{.48\textwidth}
  \centering
    \includegraphics[width=1\linewidth, trim=1cm 0cm 0cm 1cm]{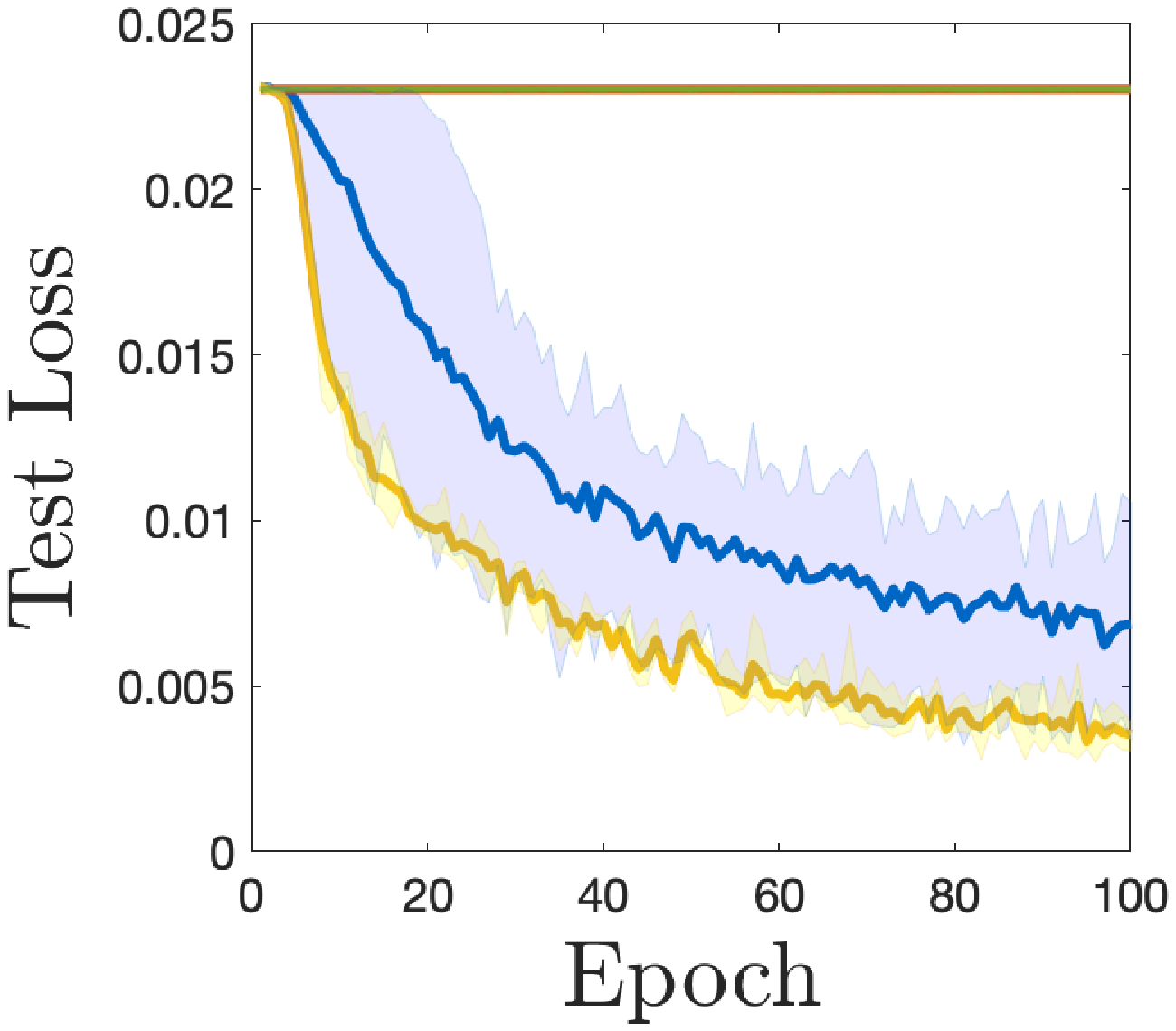}  
  \end{minipage}
  % include first image
  \begin{minipage}{.48\textwidth}
  \centering
  % include first image
  \includegraphics[width=1\linewidth, trim=1cm 0cm 0cm 1cm]{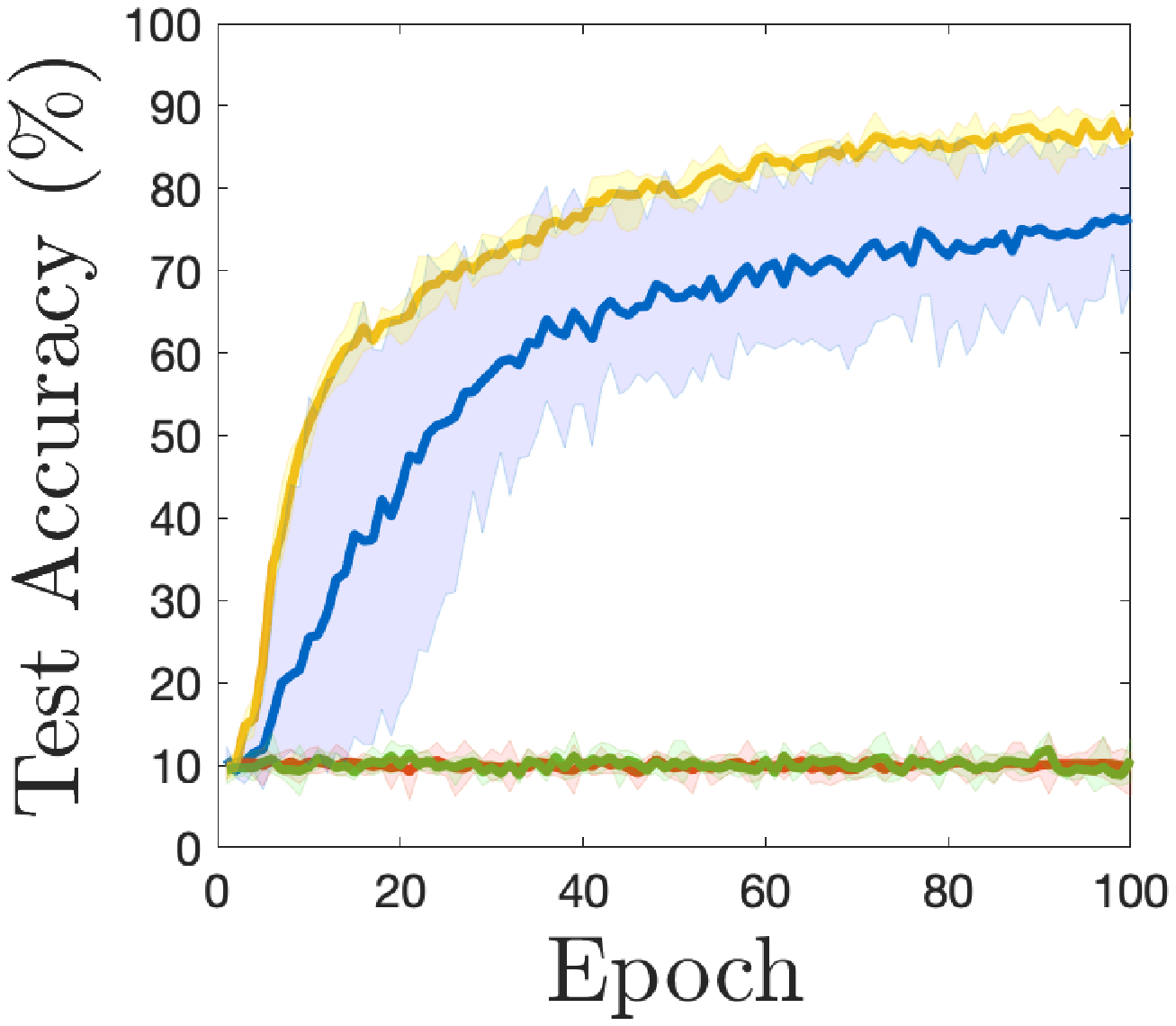}  
%   \label{fig:multi-task}
\end{minipage}
\vspace{-0.2cm}
\caption{2 Byzantine agents}
\end{subfigure}
% \begin{subfigure}{0.325\textwidth}
%   \centering
%   \begin{minipage}{.48\textwidth}
%   \centering
%     \includegraphics[width=1\linewidth, trim=1cm 0cm 0cm 1cm]{figure/synthetic_range_loss_test_attack_8.eps}  
%   \end{minipage}
%   % include first image
%   \begin{minipage}{.48\textwidth}
%   \centering
%   % include first image
%   \includegraphics[width=1\linewidth, trim=1cm 0cm 0cm 1cm]{figure/synthetic_range_acc_test_attack_8.eps}  
% %   \label{fig:multi-task}
% \end{minipage}
% \caption{8 Byzantine agents}
% \end{subfigure}
\caption{Digit Classification: average testing loss and  accuracy for normal agents in group 2.}
\label{fig: synthetic test}
\end{figure}

We plot the mean and range of the average loss of every normal agent for the target localization problem in \figref{fig: TL, no attack}--d. \textcolor{black}{Similarly, we}  
plot the mean and range of the average testing loss and classification accuracy of every normal agent for human action recognition in  \figref{fig: har test}, and for digit classification in \figref{fig: MNist test} (for group 1) and \figref{fig: synthetic test} (for group 2).
\textcolor{red}{At each iteration, Byzantine agents}
%\textcolor{blue}{Byzantine agents are designed to continuously}
\textcolor{blue}{send random values \textcolor{red}{(for each dimension)} from the interval  $[15,16]$ for target localization, and $[0,0.1]$ for the other two case studies.
}

In all of the examples, we find that the loss-based weight assignment rule \eqref{eq: filtering weight} outperforms all the other rules \textcolor{red}{and} the non-cooperative case, with respect to the mean and range of the average loss and accuracy with and without the presence of Byzantine agents. 
Hence, our simulations validate the results indicated by \eqref{eq: resilient convergence  with improved learning error} and imply that the loss-based weights  \eqref{eq: filtering weight} have accurately learned the relationship among agents. Moreover, normal agents having a large regret in their estimation benefit \textcolor{black}{from} cooperating with other agents having a small regret. 
We also consider the extreme case \textcolor{black}{in which there is only one normal agent in the network, and all the other agents are Byzantine.} 
In such \textcolor{black}{a case}, the loss-based weight assignment rule \eqref{eq: filtering weight} has the same performance as the non-cooperative case, thus, showing that it is resilient to an arbitrary number of Byzantine agents.

\vspace{-0.6cm}

%% file: appendix.tex
\newpage
\appendix 

\section{Assumptions and Theoretical Results}\label{app: Assumptions and Theoretical Results}

\subsection{Assumptions of  risk functions}\label{app: assumptions}
% \textbf{\Large A. Assumption of the loss functions}
\begin{definition}
{($L$-Lipschitz continuous gradient). } A differentiable convex function $f$ is said to have an $L$-Lipschitz continuous gradient, if there exists
a constant $L > 0$, such that
\begin{equation*}
   \| \nabla f(x) - \nabla f(y) \| \leq L \|x - y\|, \forall x, y.
\end{equation*}
If $f$ has an $L$-Lipschitz continuous gradient, then it holds that
\begin{equation*}
    f(y) \leq f(x) +  \langle \nabla f(x),  y-x\rangle + \frac{L}{2} \|y-x\|^2, \forall x, y.
\end{equation*}
\end{definition}
\begin{definition}
{($m$-strongly convex). } A differentiable convex function $f$ is said to be $m$-strongly convex if there exists a constant $m > 0$, such that 
\begin{equation*}
    f(y) \geq f(x)  +  \langle \nabla f(x),  y-x\rangle + \frac{m}{2} \|y-x\|^2, \forall x, y.
\end{equation*}
\end{definition}
% If $f$ is $m$-strongly convex and $x^* = \arg \min_{x \in \mathbb{R}^d} f(x)$, then it holds that
% \begin{equation*}
%     \|\nabla f(x)\|^2  \geq 2m (f(x) - f(x^*)), \forall x.
% \end{equation*}
If $f$ is $m$-strongly convex and has an $L$-Lipschitz continuous gradient, then it is obvious that $m \leq L$.

% \subsection{Proof of Lemma 1}
% \begin{proof}
% By sending $\| \hat{\theta}_{b,i} -  \tilde{\theta_k^*} \| \ll \| \hat{\theta}_{k,i} - \tilde{\theta_k^*} \|$ and $\| \hat{\theta}_{b,i} -  \theta_k^* \| > \| \hat{\theta}_{k,i} - \theta_k^*\|$, a Byzantine agent $b$ can gain a large weight from $k$ by the first condition and make $\theta_{k,i}$ move  away from $\theta_k^*$ by the second condition. (The same strategy can be generalized to $\ell_p$ norm.) 
% \end{proof}

\subsection{Optimal solution of equation \eqref{eq: final minimization problem}}\label{app: optimal solution of weights}
Let $\lambda$ be the Lagrange multiplier. We  define the Lagrangian of \eqref{eq: final minimization problem} given the constraints \textcolor{red}{on} the weights as
% \begin{equation*}
%   \mathcal{L}(a_{1,k},\ldots, a_{|\mathcal{N}_k|k}, \lambda) =  \sum_{l \in \mathcal{N}_k}  a^2_{lk}  \ell_{k}^i( {\hat{\theta}}_{l,i})  + \lambda (1 - \sum_{l \in \mathcal{N}_k}  a_{lk})
% \end{equation*}
\begin{equation*}
  \mathcal{L}(a_{lk}, \lambda) =  \sum_{l \in \mathcal{N}_k}  a^2_{lk}  r_{k}( {\hat{\theta}}_{l,i}  )  + \lambda (1 - \sum_{l \in \mathcal{N}_k}  a_{lk}).
\end{equation*}
Set $\nabla_{a_{lk}, \lambda} \mathcal{L}(a_{lk}, \lambda)$ $=$ $\left(\frac{\partial \mathcal{L}}{\partial a_{lk}}, \frac{\partial \mathcal{L}}{\partial \lambda} \right) = 0$, i.e.,
% \begin{equation*}
%     \begin{cases}
%     2 a_{lk}  \ell_{k}^i( {\hat{\theta}}_{l,i}) - \lambda = 0, \forall l \in \mathcal{N}_k \\
%     1 - \sum_{l \in \mathcal{N}_k}  a_{lk} = 0
%     \end{cases}
%     % \Leftrightarrow
%     %     \begin{cases}
%     % a_{lk} = \frac{\lambda}{\ell_{k}^i( {\hat{\theta}}_{l,i})}, \forall l \in \mathcal{N}_k  \\
%     % \sum_{l \in \mathcal{N}_k}  a_{lk} = 1
%     % \end{cases}
% \end{equation*}
\begin{equation*}
    \begin{cases}
    2 a_{lk}  r_{k}( {\hat{\theta}}_{l,i}  ) - \lambda = 0, \forall l \in \mathcal{N}_k, \\
    1 - \sum_{l \in \mathcal{N}_k}  a_{lk} = 0.
    \end{cases}
    % \Leftrightarrow
    %     \begin{cases}
    % a_{lk} = \frac{\lambda}{\ell_{k}^i( {\hat{\theta}}_{l,i})}, \forall l \in \mathcal{N}_k  \\
    % \sum_{l \in \mathcal{N}_k}  a_{lk} = 1
    % \end{cases}
\end{equation*}
% \begin{equation*}
%     2 a_{lk}  r_{k}( {\hat{\theta}}_{l,i}  ) - \lambda = 0, \forall l \in \mathcal{N}_k, 
%     1 - \sum_{l \in \mathcal{N}_k}  a_{lk} = 0
%     % \Leftrightarrow
%     %     \begin{cases}
%     % a_{lk} = \frac{\lambda}{\ell_{k}^i( {\hat{\theta}}_{l,i})}, \forall l \in \mathcal{N}_k  \\
%     % \sum_{l \in \mathcal{N}_k}  a_{lk} = 1
%     % \end{cases}
% \end{equation*}
Thus,  $a_{lk} = \frac{\lambda}{r_{k}( {\hat{\theta}}_{l,i}  )}, \forall l \in \mathcal{N}_k$ and $\sum_{l \in \mathcal{N}_k}  a_{lk} = 1$.
We have $ \lambda \sum_{l \in \mathcal{N}_k}  \frac{1}{r_{k}( {\hat{\theta}}_{l,i}  )} = 1$ and hence  $\lambda = \frac{1}{\sum_{l \in \mathcal{N}_k} {r_{k}( {\hat{\theta}}_{l,i}  )}^{-1}}$, and  
$
    a_{lk} = \frac{{r_k({{\hat{\theta}}_{l,i}  })}^{-1}}{\sum_{p \in \mathcal{N}_k} {r_k({{\hat{\theta}}_{p,i}  })}^{-1}}
$
is \textcolor{red}{the} optimal solution of \eqref{eq: final minimization problem}.

\subsection{Proof of Lemma \ref{lemma: expected regret for cooperation}}
\begin{proof}
Given \eqref{eq: combine}, 
$
     r_k(\theta_{k,i}   )  = r_k\left(\sum_{l \in \mathcal{N}_k} a_{lk}(i)  \hat{\theta}_{l,i}    \right).
$
Using Jensen's inequality, we have
\begin{equation}\label{eq: expection r theta_j,i+1}
\begin{aligned}
r_k(\theta_{k,i}   ) \leq \sum_{l \in \mathcal{N}_k} a_{lk}(i)   r_k  \left(\hat{\theta}_{l,i}    \right).
\end{aligned}
\end{equation}
Subtracting $r_k(\theta_k^*)$ from both sides of \eqref{eq: expection r theta_j,i+1} and taking expectations over the joint distribution $\xi_k$, we obtain
\begin{equation}\label{eq: regret as a sum of neighbor regret t -> infty}
\begin{aligned}
\mathbb{E} \left[r_k(\theta_{k,i}  ) -r_k(\theta_k^*) \right] \leq & \sum_{l \in \mathcal{N}_k} \mathbb{E}[a_{lk}(i)]   \mathbb{E} \left[r_k  \left(\hat{\theta}_{l,i}   \right) -r_k(\theta_k^*)\right]\\
\leq &  \frac{\sum_{l \in \mathcal{N}_k}  {\mathbb{E} \left[r_k({{\hat{\theta}}_{l,i}  })\right]}^{-1} \mathbb{E} \left[r_k  \left(\hat{\theta}_{l,i}   \right) -r_k(\theta_k^*)\right]}{\sum_{p \in \mathcal{N}_k} {\mathbb{E} \left[r_k({{\hat{\theta}}_{p,i}}  ) \right]}^{-1}}.
\end{aligned}
\end{equation}
% Suppose at $t \rightarrow \infty$, $\mathbb{E}[a_{lk}(t)] = a_{lk}^*(t)$. Then at $i \geq t$, we have
% \begin{equation}\label{eq: regret as a sum of neighbor regret t -> infty}
% \begin{aligned}
% \mathbb{E} \left[r_k(\theta_{k,i}  ) -r_k^* \right] \leq  \frac{\sum_{l \in \mathcal{N}_k}  {r_k({{\hat{\theta}}_{l,i}  })}^{-1} \mathbb{E} \left[r_k  \left(\hat{\theta}_{l,i}   \right) -r_k^*\right]}{\sum_{p \in \mathcal{N}_k} {r_k({{\hat{\theta}}_{p,i}}  )}^{-1}}.
% \end{aligned}
% \end{equation}
% Since $l \in \mathcal{N}_k$,
% \begin{equation}
%     r_k\left(\hat{\theta}_{l,i}  \right) \leq  r_k\left(\hat{\theta}_{k,i}  \right), \text{ and } 
% r_k\left(\hat{\theta}_{l,i}  \right)^{-1}\geq  r_k\left(\hat{\theta}_{k,i}  \right)^{-1}, \forall l
% \end{equation}
\textcolor{purple}{We next prove the right-hand side of \eqref{eq: regret as a sum of neighbor regret t -> infty} is less than  $\frac{1}{ |\mathcal{N}_k|} \sum_{l \in \mathcal{N}_k}  \mathbb{E} \left[r_k(\hat{\theta}_{l,i}  ) -r_k(\theta_k^*) \right]$. }
For succinctness, we use $\chi_{l,i}$ to denote $\mathbb{E} \left[r_k\left(\hat{\theta}_{l,i}  \right) \right]^{-1}$, and $\Delta_{l,i}$ to denote $\mathbb{E} \left[r_k  \left(\hat{\theta}_{l,i}   \right) -r_k(\theta_k^*)\right]$.
And we aim to prove $ \frac{\sum_{l \in \mathcal{N}_k} \chi_{l,i} \Delta_{l,i}}{\sum_{p \in \mathcal{N}_k} \chi_{p,i}} \leq \frac{1}{ |\mathcal{N}_k|} \sum_{l \in \mathcal{N}_k}  \Delta_{l,i}$, or equivalently, $|\mathcal{N}_k| \sum_{l \in \mathcal{N}_k} \chi_{l,i} \Delta_{l,i} \leq  \sum_{p \in \mathcal{N}_k} \chi_{p,i} \sum_{l \in \mathcal{N}_k}  \Delta_{l,i}$.

\textcolor{purple}{When $|\mathcal{N}_k| = 1$, one can easily validate that this condition holds.}
When $|\mathcal{N}_k| \geq 2$, 
let $l_1^i$ be the one with the smallest risk $ r_k\left(\hat{\theta}_{l_1^i,i}  \right) = \min_{l \in \mathcal{N}_k} r_k\left(\hat{\theta}_{l,i}  \right) $ and $l_2^i$ be the one with the second smallest risk $ r_k\left(\hat{\theta}_{l_2^i,i}  \right) = \min_{l \in  \mathcal{N}_k \backslash l_1^i} r_k\left(\hat{\theta}_{l,i}  \right)$. 
Hence, $\chi_{l_1^i, i} \geq \chi_{l_2^i, i} \geq \chi_{l, i}$, and $\Delta_{l_1^i,i} \leq \Delta_{l_2^i,i} \leq \Delta_{l,i}$ for $l \in \mathcal{N}_k \backslash \{l_1^i,l_2^i\}$.
Thus,
\begin{equation*}
\begin{aligned}
  & |\mathcal{N}_k| \sum_{l \in \mathcal{N}_k} \chi_{l,i} \Delta_{l,i}-  \sum_{p \in \mathcal{N}_k} \chi_{p,i} \sum_{l \in \mathcal{N}_k}  \Delta_{l,i} \\
   =&  \sum_{l \in \mathcal{N}_k} \chi_{l,i}  \left( |\mathcal{N}_k| \Delta_{l,i} - \sum_{p \in \mathcal{N}_k}  \Delta_{p,i} \right)\\
   =& \chi_{l_1^i,i}  \left( \left(|\mathcal{N}_k| - 1\right) \Delta_{l_1^i,i} - \sum_{l \in \mathcal{N}_k \backslash l_1^i}  \Delta_{l,i} \right) + \sum_{l \in \mathcal{N}_k \backslash l_1^i,i} \chi_{l,i}  \left( |\mathcal{N}_k| \Delta_{l,i} - \sum_{p \in \mathcal{N}_k}  \Delta_{p,i} \right) \\
  \leq& \chi_{l_1^i,i}  \left( \left(|\mathcal{N}_k| - 1\right) \Delta_{l_1^i,i} - \sum_{l \in \mathcal{N}_k \backslash l_1^i}  \Delta_{l,i} \right) +  \chi_{l_2^i,i} \left(\sum_{l \in \mathcal{N}_k \backslash l_1^i}  |\mathcal{N}_k| \Delta_{l,i} - ( |\mathcal{N}_k| - 1) \sum_{p \in \mathcal{N}_k}  \Delta_{p,i}
  \right)\\
   =& \chi_{l_1^i,i}  \left( \left(|\mathcal{N}_k| - 1\right) \Delta_{l_1^i,i} - \sum_{l \in \mathcal{N}_k \backslash l_1^i}  \Delta_{l,i} \right) +  \chi_{l_2^i,i}  \left( \sum_{l \in \mathcal{N}_k  \backslash l_1^i } \Delta_{l,i} -   \left(|\mathcal{N}_k| - 1\right)  \Delta_{l_1^i,i} \right) \\
   =& \left(\chi_{l_1^i,i} - \chi_{l_2^i,i}\right)  \left( \left(|\mathcal{N}_k| - 1\right) \Delta_{l_1^i,i} - \sum_{l \in \mathcal{N}_k \backslash l_1^i}  \Delta_{l,i} \right)\\
   =& \left(\chi_{l_1^i,i} - \chi_{l_2^i,i}\right)  \left(  \sum_{l \in \mathcal{N}_k \backslash l_1^i} \left( \Delta_{l_1^i,i} - \Delta_{l,i} \right)\right) \leq 0.
\end{aligned}
\end{equation*}
Therefore, 
$ \frac{\sum_{l \in \mathcal{N}_k} \chi_{l,i} \Delta_{l,i}}{\sum_{p \in \mathcal{N}_k} \chi_{p,i}} \leq \frac{1}{ |\mathcal{N}_k|} \sum_{l \in \mathcal{N}_k}  \Delta_{l,i}$.
Put it back to \eqref{eq: regret as a sum of neighbor regret t -> infty}, we obtain 
\begin{equation*}
\begin{aligned}
\mathbb{E} \left[r_k(\theta_{k,i}  ) -r_k(\theta_k^*) \right] \leq  \frac{1}{ |\mathcal{N}_k|} \sum_{l \in \mathcal{N}_k} \mathbb{E} \left[r_k  \left(\hat{\theta}_{l,i}   \right) -r_k(\theta_k^*)\right],
\end{aligned}
\end{equation*}
which completes the proof.
\end{proof}

\subsection{Proof of Theorem 1}\label{app: proof of theorem 1}
\begin{proof}
Let $\mathbb{E}[\cdot]$ denote the expected value taken with respect to the joint distribution of all random
variables $\xi_k$ and $\xi_l$ for $l \in \mathcal{N}_k^{\leq}$, i.e.
% \begin{equation*}
%     \mathbb{E}  \left[ \cdot \right] = \mathbb{E}_{\xi_k}   \mathbb{E}_{\xi_{l,1}} \ldots  \mathbb{E}_{\xi_{l,{|\mathcal{N}_k^{\leq}|-1}}}       
%     \left[\cdot \right].
% \end{equation*}
\begin{equation*}
    \mathbb{E}  \left[ \cdot \right] = \mathbb{E}_{\xi_k}   \mathbb{E}_{\{\xi_{l}| l \in \mathcal{N}_k^{\leq} \}} 
    \left[\cdot \right].
\end{equation*}
Similar to the proof for Lemma \ref{lemma: expected regret for cooperation},
using $\mathcal{N}_k^{\leq}$ in the place of $\mathcal{N}_k$, 
with rule \eqref{eq: filtering weight}, we  obtain 
\begin{equation}\label{eq: expected regret smaller than sum of regret for filtering weights}
\begin{aligned}
\mathbb{E} \left[r_k(\theta_{k,i}) -r_k(\theta_k^*) \right] \leq  \frac{1}{ |\mathcal{N}_k^{\leq}|} \sum_{l \in \mathcal{N}_k} \mathbb{E} \left[r_k  \left(\hat{\theta}_{l,i} \right) -r_k(\theta_k^*)\right].
\end{aligned}
\end{equation}
For  every $l \in \mathcal{N}_k^{\leq}$, we have
$r_k(\hat{\theta}_{l,i}) \leq  r_k(\hat{\theta}_{k,i})$ and hence 
$\frac{1}{ |\mathcal{N}_k^{\leq}|} \sum_{l \in \mathcal{N}_k} \mathbb{E}\left[ \left(r_k(\hat{\theta}_{l,i}) - r_k(\theta_k^*)\right) \right] \leq \mathbb{E}\left[ \left(r_k(\hat{\theta}_{k,i}) - r_k(\theta_k^*)\right) \right]$.
Put it back to \eqref{eq: expected regret smaller than sum of regret for filtering weights}, we obtain
\begin{equation}\label{eq: cooperation with filtering reduce expected regret}
\begin{aligned}
\mathbb{E} \left[r_k\left(\theta_{k,i}\right) -r_k(\theta_k^*) \right] \leq  \frac{1}{ |\mathcal{N}_k^{\leq}|} \sum_{l \in \mathcal{N}_k^{\leq}} \mathbb{E} \left[r_k  \left(\hat{\theta}_{l,i} \right) -r_k(\theta_k^*)\right] 
\leq \mathbb{E} \left[r_k  \left(\hat{\theta}_{k,i} \right)-r_k(\theta_k^*)\right], \forall k \in \mathcal{N}^+, i \in \mathbb{N},
% = \mathbb{E} \left[r_k  \left({\theta}_{k,i}^{\text{\rm(ncop)}} \right)-r_k(\theta_k^*)\right]
\end{aligned}
\end{equation}
\textcolor{purple}{
which yields 
\eqref{eq: resilient convergence  with improved learning error}.}

\textcolor{purple}{
We next prove the convergence of the algorithm with the proposed weight assignment rule. Given Assumptions 1-3, we obtain from  \cite{DBLP:journals/siamrev/BottouCN18} that using constant step size $\mu_k \in (0, \frac{1}{Lc_k}]$, it holds that
\begin{equation*}
\begin{aligned}
            \mathbb{E}
           \left[r_k \left(\hat{\theta}_{k,i} \right) - r_k(\theta_k^*) \right] 
          -  \frac{\mu_k L \sigma_k^2}{2 m}  \leq  (1-\mu_k m) \left(   \mathbb{E}
           \left[r_k \left({\theta}_{k,i-1}\right) - r_k(\theta_k^*)\right]  - \frac{\mu_k L \sigma_k^2}{2 m} \right).
\end{aligned}
\end{equation*}
Combined with \eqref{eq: cooperation with filtering reduce expected regret}, we obtain
\begin{equation}\label{eq: iteration base}
\begin{aligned}
            \mathbb{E}
           \left[r_k \left({\theta}_{k,i} \right) - r_k(\theta_k^*) \right] 
          -  \frac{\mu_k L \sigma_k^2}{2 m}  \leq  (1-\mu_k m) \left(   \mathbb{E}
           \left[r_k \left({\theta}_{k,i-1}\right) - r_k(\theta_k^*)\right]  - \frac{\mu_k L \sigma_k^2}{2 m} \right).
\end{aligned}
\end{equation}
Given $\mu_k \in (0, \frac{1}{Lc_k}]$, with $c_k \geq 1$, $m \leq L$, it holds that $(1-\mu_k m) \in [0,1)$. 
Applying \eqref{eq: iteration base} repeatedly through iteration $i \in \mathbb{N}$, we obtain
\begin{equation*}
\begin{aligned}
            \mathbb{E}
           \left[r_k \left({\theta}_{k,i} \right) - r_k(\theta_k^*) \right] &\leq
           \frac{\mu_k L \sigma_k^2}{2 m} + (1-\mu_k m)^i \left(  
           r_k \left({\theta}_{k,0}\right) - r_k(\theta_k^*)  - \frac{\mu_k L \sigma_k^2}{2 m} \right) \\
           & \stackrel{i \rightarrow \infty}{\longrightarrow} \frac{\mu_k L \sigma_k^2}{2 m}.
\end{aligned}
\end{equation*}
This means $\theta_{k,i}$ converges towards $\theta_{k}^*$ with the expected regret bounded by $\frac{\mu_k L \sigma_k^2}{2 m}$.
}
\end{proof}

\section{Simulation Details and Supplementary Results}\label{app: Simulation Details and Supplementary Results}
\subsection{Simulation details of Target Localization}
% Target localization is a widely-studied linear regression example in distributed diffusion algorithms \cite{DBLP:conf/icassp/ChenRS14}. We consider a network of 100 agents in the network as shown in \figref{fig: target localization network}. 
% There are four targets in $\mathbb{R}^2$: $(10.84, 10.76), (20.42, 20.26), (20.51, 10.40), (10.78, 20.30)$.
% Agents are interested in estimating the location of one of the four targets and agents sharing the same target are indicated in the same color yet they do not know which agents are estimating the same target as themselves.
The four target locations in $\mathbb{R}^2$ are: $(10.84, 10.76), (20.42, 20.26), (20.51, 10.40), (10.78, 20.30)$. \textcolor{red}{Agents' locations} are indicated in \figref{fig: target localization network}.
An edge between two agents means \textcolor{red}{they} are neighbors.
At each iteration, \textcolor{red}{every} agent $k$ has a noisy observation (streaming data) of the distance $\bm{d}_{k}(i)$ and the unit direction vector $\bm{u}_{k,i}$ pointing from $x_k$ to its target based on built-in sensors.  Let $\theta_k \in \mathbb{R}^2$ denote the estimation of the target location for agent $k$, then the loss is computed as $\ell_k(\theta_{k,i}; \xi_k^i) = \|\bm{d}_{k}(i) - (\theta_k -x_k)^\top \bm{u}_{k,i}\|^2$, and the agent  estimates  $\theta_k$ using the SGD algorithm as well as the ATC diffusion  algorithm with different weight assignment rules. 
The distance measurement data has noise variance $\sigma_{d,k}^2 \in [0.1, 0.2]$, and the unit direction vector has additive white Guassian noise with diagnonal covariance matrices $R_{u,k} = \sigma^2_{u,k} I_2$, with  $\sigma^2_{u,k} \in [0.01,0.1]$ for different $k$.
We tune the step-sizes and forgetting factors from the interval $(0,1)$ and find the best empirical performance by setting them to be  $\mu_k = 0.1$  and   $\nu_k = 0.1$  for every normal agent $k$.
$\varphi_{lk}^{-1}$ and $\phi_{lk}^{-1}$ are initialized to be zero for all $l \in \mathcal{N}_k$.
Byzantine agents are designed to continuously send random values for each dimension from the interval   $[15,16]$ at each iteration.

\subsection{Simulation details and supplementary results of Human Action Recognition}
% Mobile phone sensor data (accelerometer and gyroscope)\footnote{https://archive.ics.uci.edu/ml/datasets/human+activity+recognition+using+smartphones} is collected from 30 individuals   performing one of six activities: \{walking, walking-upstairs, walking-downstairs, sitting, standing, lying-down\}. 
% The goal is to predict the activities they perform using the  561-length feature vectors for each instance generated by the processed sensor signals \cite{DBLP:conf/esann/AnguitaGOPR13}.
% We model each individual as a separate task and use a complete graph to model the network topology among the 30 agents.
% We randomly split the data into 75\% training and 25\% testing for each agent and use a linear model $f( x_k^i) = \theta_{k}^\top  x_k^i$ with cross-entropy-loss.
We randomly split the data into 75\% training and 25\% testing for each agent.
During training, ten of the \textcolor{red}{thirty} agents are randomly selected to have access to much less data (about ${\frac{1}{10}} {th}$) than the other agents at each epoch. \textcolor{red}{This is} to model the realistic scenario in which some of the agents may have less data samples and they may learn slowly than others. 
We use mini-batch gradient descent with batch size of 10.
We tune the step-sizes and forgetting factors from the interval $(0,1)$ and find the best empirical performance by setting them to be
$\mu_k = 0.01$ and  $\nu_k = 0.05$ for every normal agent $k$.
$\varphi_{lk}^{-1}$ and $\phi_{lk}^{-1}$ are initialized to be zero for all $l \in \mathcal{N}_k$.
Byzantine agents are designed to send a model with very small noisy elements for each dimension from the interval $[0,0.1]$  at each iteration.

\begin{figure}[H]
\centering
\vspace{0.1cm}
\begin{subfigure}{0.495\textwidth}
  \centering
  \begin{minipage}{.48\textwidth}
  \centering
    \includegraphics[width=1\linewidth, trim=1cm 0cm 0cm 1cm]{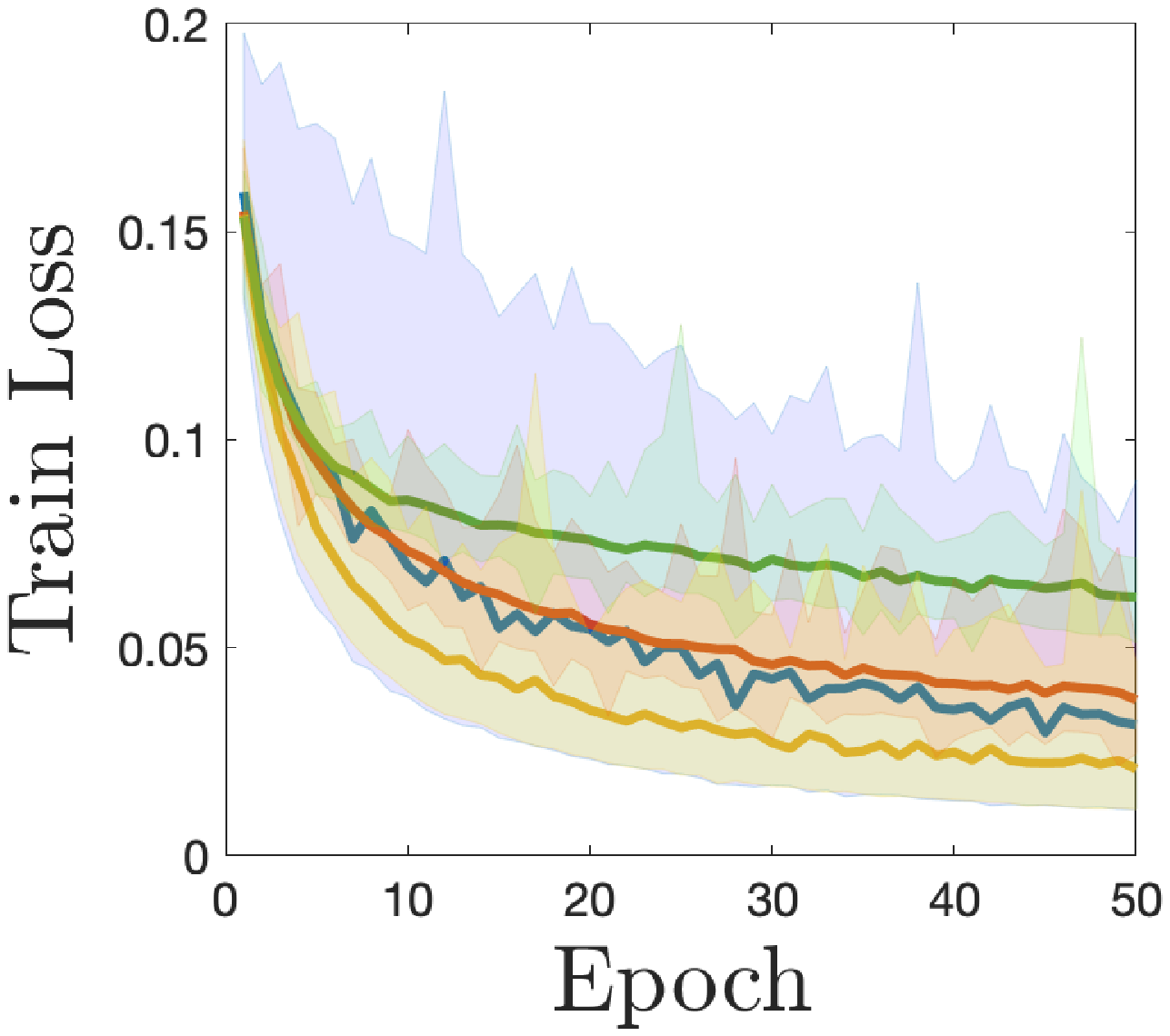}  
  \end{minipage}
    \begin{minipage}{.48\textwidth}
  \centering
    \includegraphics[width=1\linewidth, trim=1cm 0cm 0cm 1cm]{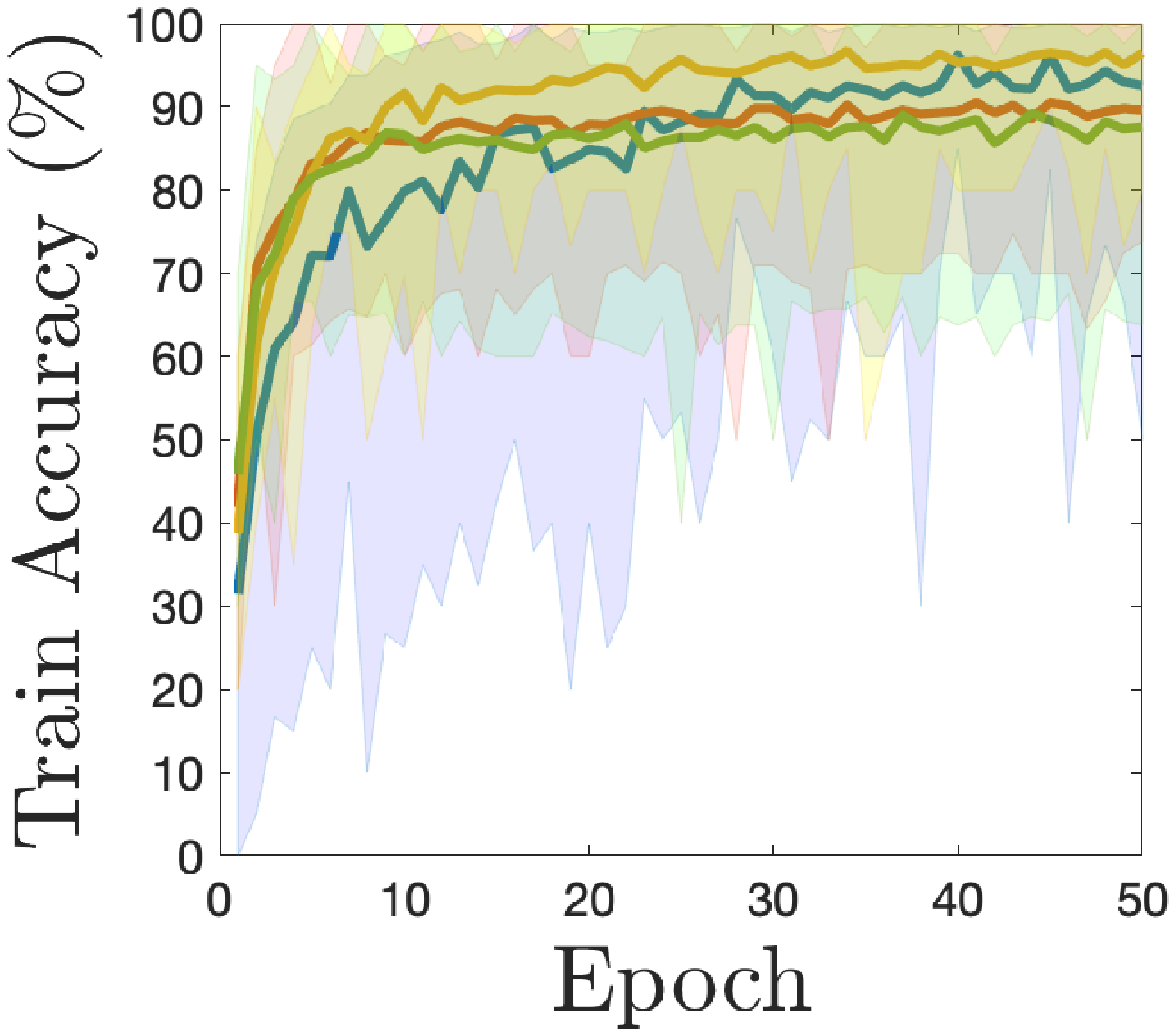}  
  \end{minipage}
\caption{No attack}
\end{subfigure}
\begin{subfigure}{0.495\textwidth}
  \centering
  \begin{minipage}{.48\textwidth}
  \centering
    \includegraphics[width=1\linewidth, trim=1cm 0cm 0cm 1cm]{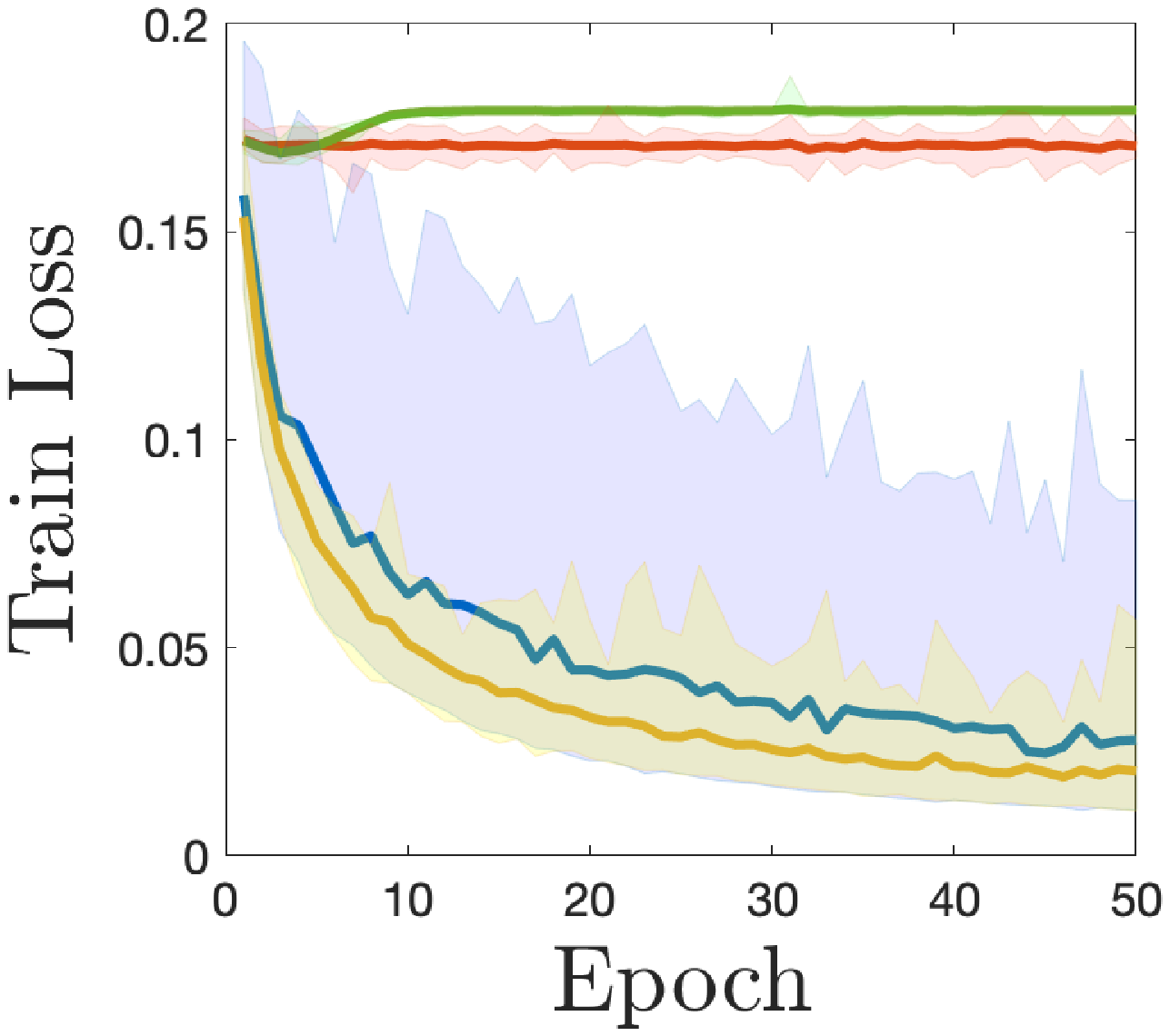}  
  \end{minipage}
  % include first image
  \begin{minipage}{.48\textwidth}
  \centering
  % include first image
  \includegraphics[width=1\linewidth, trim=1cm 0cm 0cm 1cm]{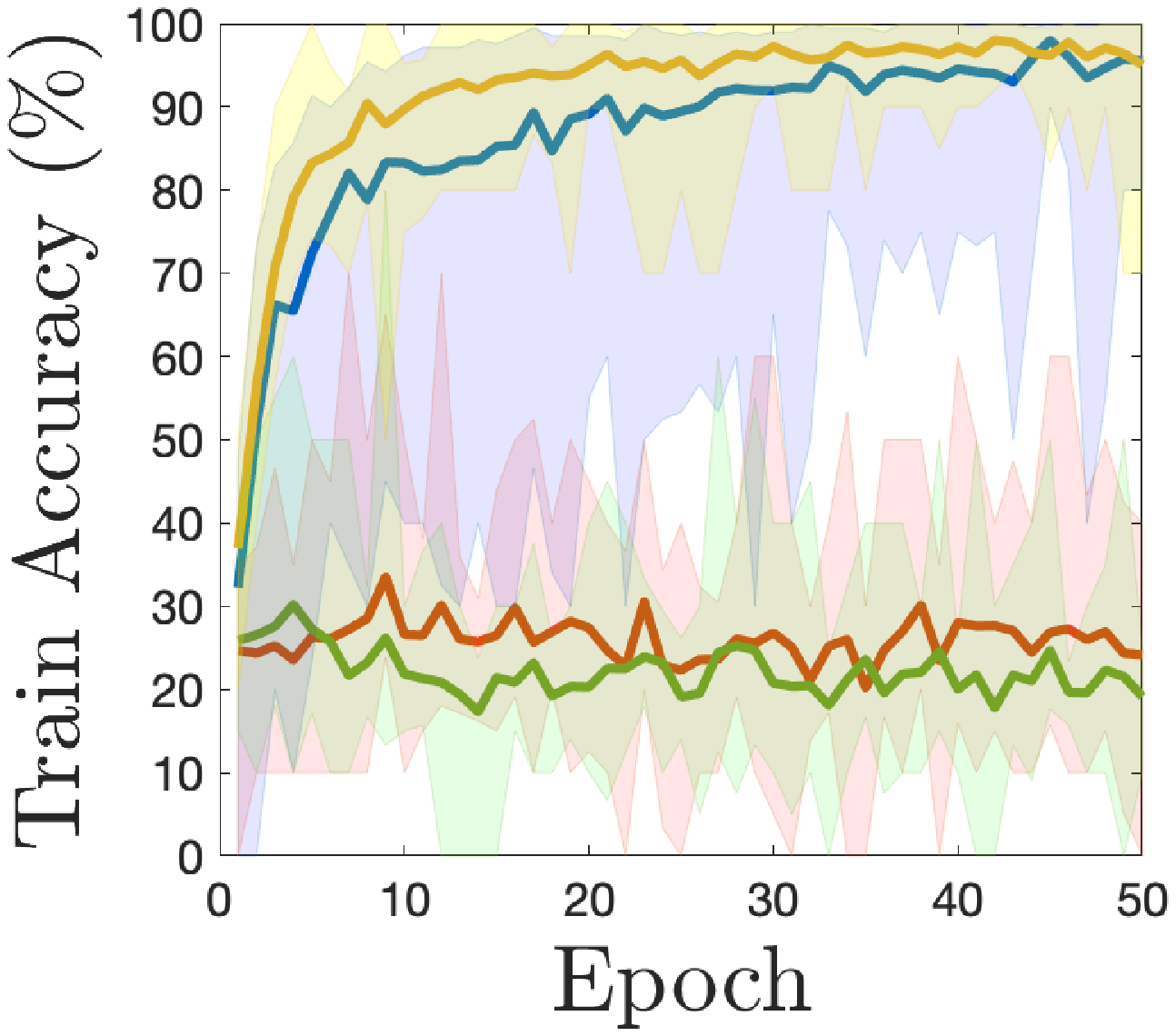}  
%   \label{fig:multi-task}
\end{minipage}
\caption{10 Byzantine agents}
\end{subfigure}
% \begin{subfigure}{0.325\textwidth}
%   \centering
%   \begin{minipage}{.48\textwidth}
%   \centering
%     \includegraphics[width=1\linewidth, trim=1cm 0cm 0cm 1cm]{figure/HAR_range_loss_train_attack_99.eps}  
%   \end{minipage}
%   % include first image
%   \begin{minipage}{.48\textwidth}
%   \centering
%   % include first image
%   \includegraphics[width=1\linewidth, trim=1cm 0cm 0cm 1cm]{figure/HAR_range_acc_train_attack_99.eps}  
% %   \label{fig:multi-task}
% \end{minipage}
% \caption{29 Byzantine agents}
% \end{subfigure}
\caption{Human Action Recognition: average training loss and  accuracy for normal agents.}
\label{fig: har train}
\end{figure}

\begin{figure}[ht]
\centering
% \vspace{0.1cm}
\begin{subfigure}{0.495\textwidth}
  \centering
  \begin{minipage}{.48\textwidth}
  \centering
    \includegraphics[width=1\linewidth, trim=1cm 0cm 0cm 1cm]{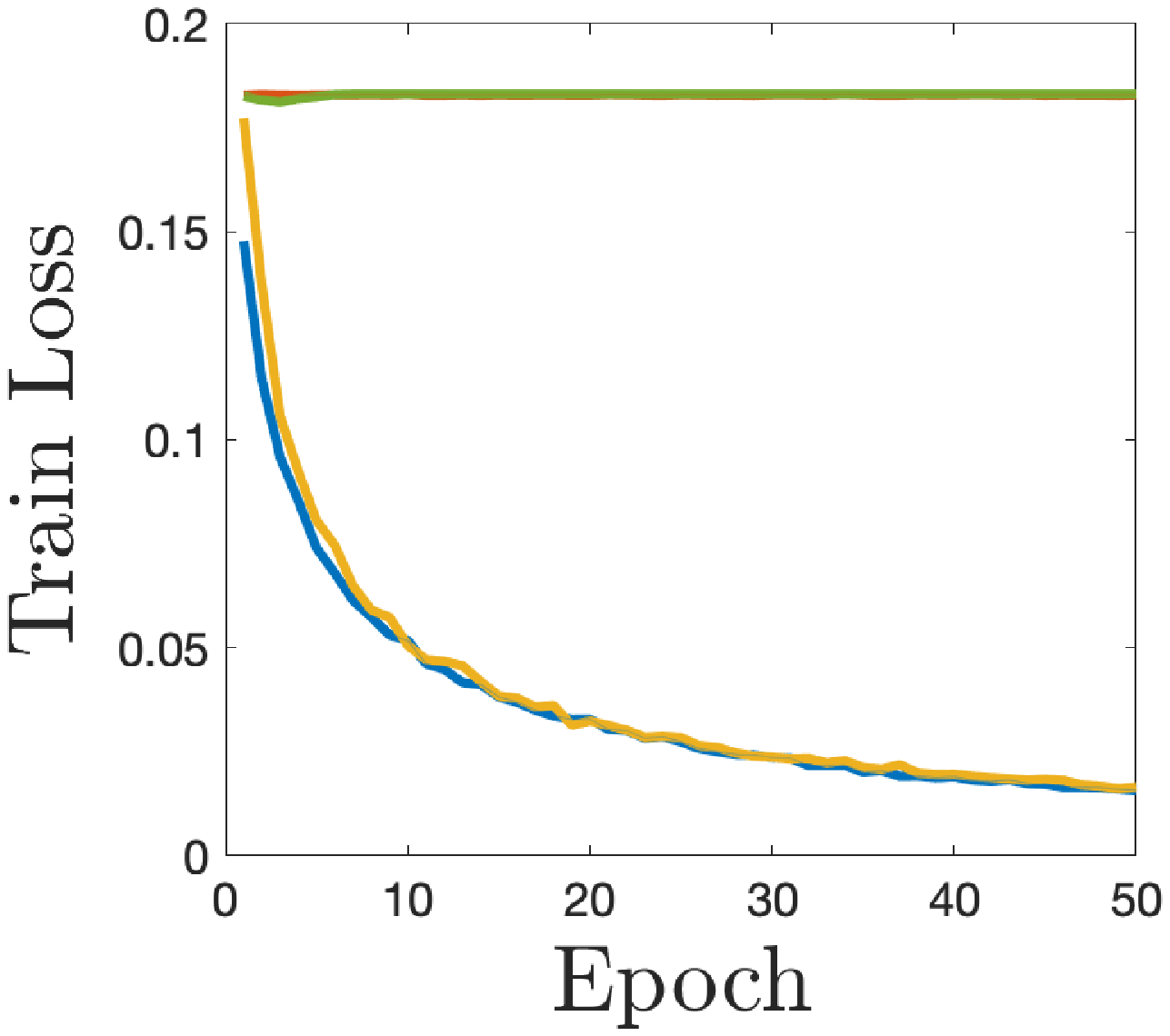}  
  \end{minipage}
  % include first image
  \begin{minipage}{.48\textwidth}
  \centering
  % include first image
  \includegraphics[width=1\linewidth, trim=1cm 0cm 0cm 1cm]{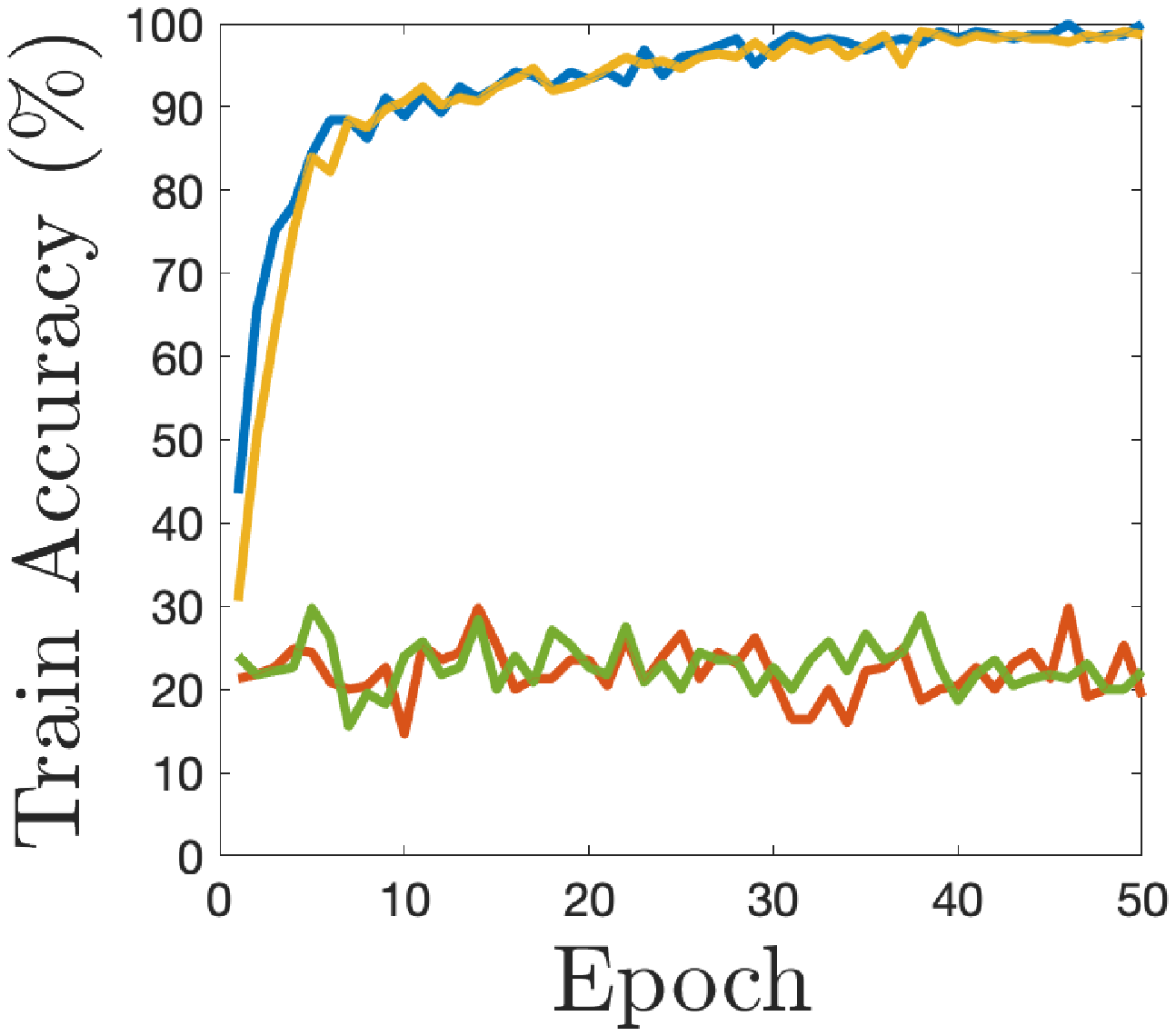}  
%   \label{fig:multi-task}
\end{minipage}
  \vspace{-0.2cm}
\caption{Training}
\label{fig: har train 29 Byzantine agents}
\end{subfigure}
\begin{subfigure}{0.495\textwidth}
  \centering
  \begin{minipage}{.48\textwidth}
  \centering
    \includegraphics[width=1\linewidth, trim=1cm 0cm 0cm 1cm]{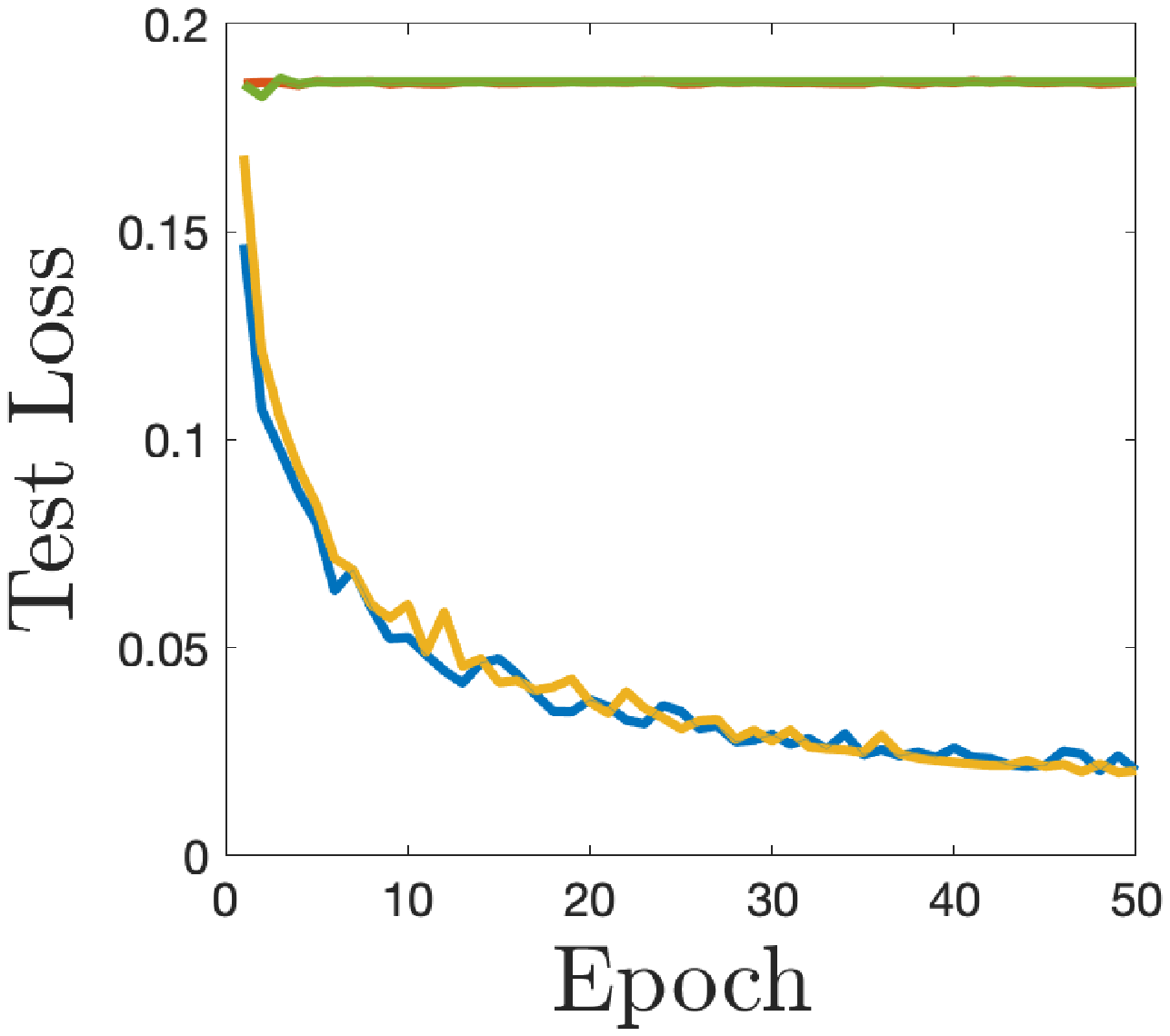}  
  \end{minipage}
    \begin{minipage}{.48\textwidth}
  \centering
    \includegraphics[width=1\linewidth, trim=1cm 0cm 0cm 1cm]{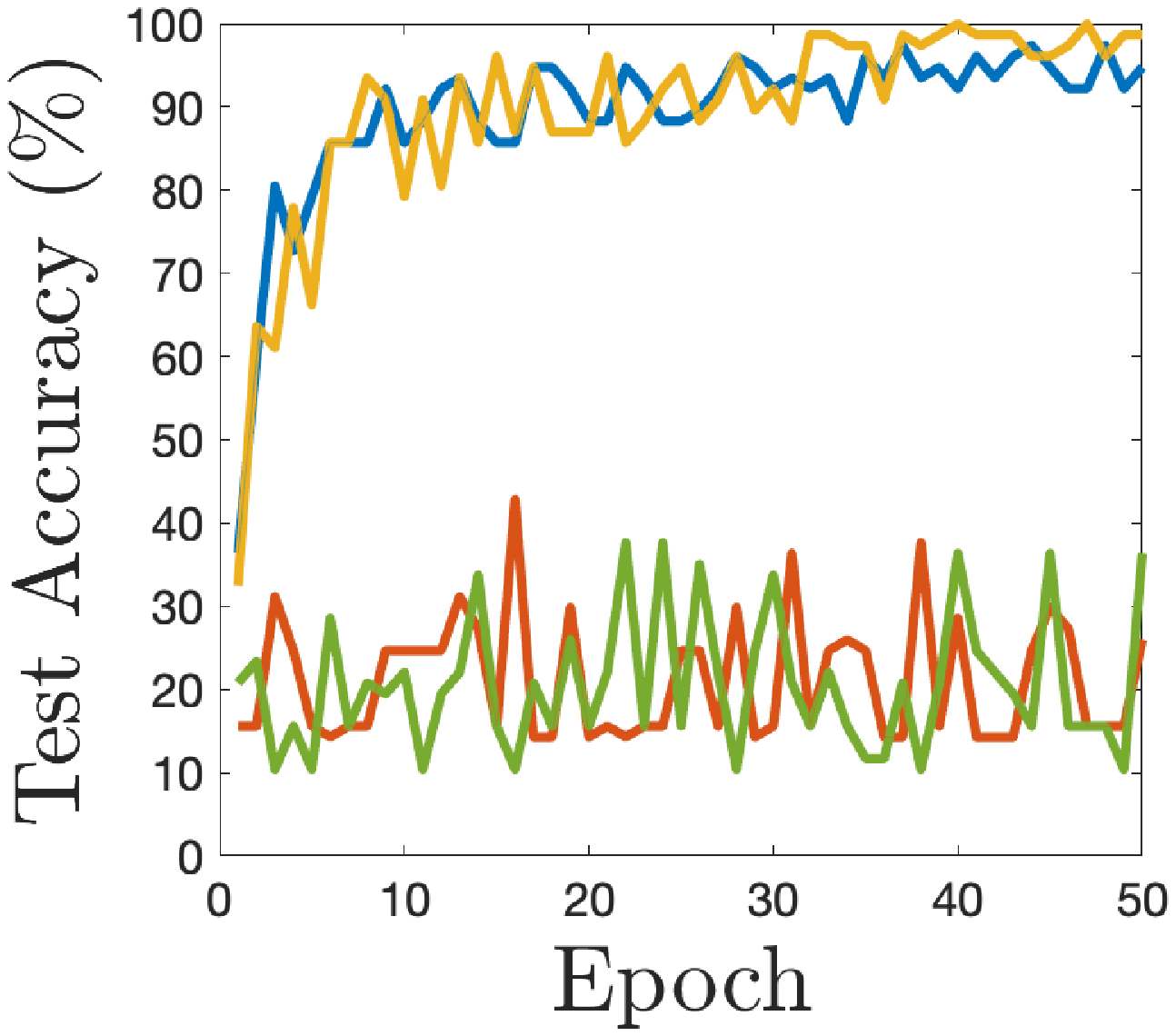}  
  \end{minipage}
  \vspace{-0.2cm}
\caption{Testing}
\label{fig: har test 29 Byzantine agents}
\end{subfigure}
% \begin{subfigure}{0.325\textwidth}
%   \centering
%   \begin{minipage}{.48\textwidth}
%   \centering
%     \includegraphics[width=1\linewidth, trim=1cm 0cm 0cm 1cm]{figure/HAR_range_loss_test_attack_99.eps}  
%   \end{minipage}
%   % include first image
%   \begin{minipage}{.48\textwidth}
%   \centering
%   % include first image
%   \includegraphics[width=1\linewidth, trim=1cm 0cm 0cm 1cm]{figure/HAR_range_acc_test_attack_99.eps}  
% %   \label{fig:multi-task}
% \end{minipage}
% \caption{29 Byzantine agents}
% \end{subfigure}
\caption{Human Action Recognition: average training/testing loss and  accuracy for normal agents with 29 Byzantine agents.}
\end{figure}

\figref{fig: har test 29 Byzantine agents} shows the average \emph{testing} loss and classification accuracy of the normal agent when 29 out of 30 agents are Byzantine  (the only normal agent has access to the entire training data).
\figref{fig: har train} and \figref{fig: har train 29 Byzantine agents} show the mean and range of the average \emph{training} loss and classification accuracy of the normal agents in the case of no attack, with 10 random selected Byzantine agents, and with 29 Byzantine agents.
In all the examples,  for both training and testing, we observe that the loss-based weight assignment rule \eqref{eq: filtering weight} outperforms the other rules as well as the non-cooperative case, with respect to the mean and range of the average loss and accuracy, which validates the result indicated by \eqref{eq: resilient convergence  with improved learning error}.
Even in the extreme case in which there is only one normal agent in the network and all of its neighbors are Byzantine, the loss-based weight assignment rule \eqref{eq: filtering weight} has the same performance as the non-cooperative case, showing its resilience to an arbitrary number of Byzantine agents.

% \subsection{Supplementary simulation results of Section \ref{sec: human action recognition}}

\subsection{Simulation details and supplementary results of Digit Classification}
\begin{figure}[H]
 \centering
\begin{subfigure}{.45\textwidth}
  \centering
  % include first image
  \includegraphics[width=0.95\linewidth, trim=0cm 0cm 0cm 0cm]{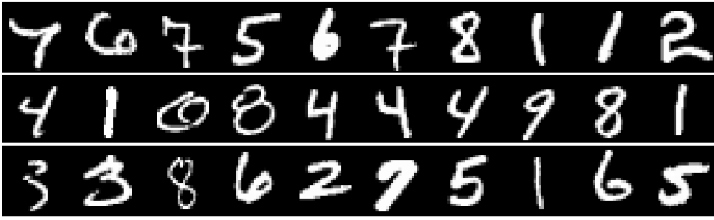}  
  \caption{MNIST}
  \label{fig:mnist}
\end{subfigure}
\begin{subfigure}{.45\textwidth}
  \centering
  % include second image
  \includegraphics[width=0.93\linewidth, trim=0cm 0cm 0cm 0cm]{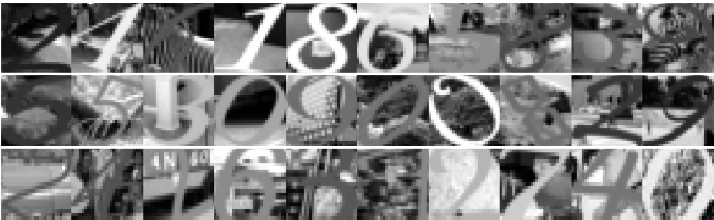}  
  \caption{Synthetic digits}
  \label{fig:synthetic digits}
\end{subfigure}
\caption{Examples of the digit classification dataset}
\label{fig: Examples of the Digit classification dataset}
\end{figure}

% We consider a network of ten agents  performing the digit classification tasks.
% Five of the ten agents are access to the MNIST dataset\footnote{http://yann.lecun.com/exdb/mnist} \cite{lecun-mnisthandwrittendigit-2010}, and the other five are access to the synthetic digits dataset\footnote{https://www.kaggle.com/prasunroy/synthetic-digits} \cite{roy2018effects} that is composed by generated images of digits embedded on random backgrounds.
% All the images are preprocessed to be $28 \times 28 \times 1$ grayscale images.
% We model each agent as a separate task and use a complete graph to  model the network topology among the ten agents.
% We classify the digits $0-9$ using the preprocessed images.
The preprocessed examples of the two datasets are given in \figref{fig: Examples of the Digit classification dataset}.
% We refer to the 5 agents for the MNIST dataset as group 1 and the other 5 agents for the synthetic digits dataset as group 2. However, agents do not know which of its neighbors are performing the same task as itself (or which of the neighbors are access to the same dataset as itself).
% For each group, we consider agents are access to uneven sizes of training data.
% For each agent, we randomly pass $200-2000$ training data and $400$ testing data from the corresponding dataset for each epoch. 
% Since some of the agents are access to limited training data, they are prone to overfitting or learn slowly if without cooperation.
The details of  the CNN architecture is given in Table \ref{tab: CNN architecture of Section 6.3}.
For each group, we consider that agents \textcolor{red}{have} access to uneven sizes of training data.
Specifically, for each agent, we randomly feed  $200-2000$ training data and $400$ testing data from the corresponding dataset for each epoch. 
We use mini-batch gradient descent with batch size of 64.
We tune the step-sizes and forgetting factors from the interval $(0,1)$ and find the best empirical performance by setting them to be
$\mu_k = 0.001$ and $\nu_k = 0.05$ for every normal agent.
$\varphi_{lk}^{-1}$ and $\phi_{lk}^{-1}$ are initialized to be zero for all $l \in \mathcal{N}_k$.
Byzantine agents are designed to send a model with very small noisy elements  for each dimension from the interval $[0,0.1]$ at each iteration.

\begin{figure}[H]
\centering
\begin{subfigure}{0.495\textwidth}
  \centering
  \begin{minipage}{.48\textwidth}
  \centering
    \includegraphics[width=1\linewidth, trim=1cm 0cm 0cm 1cm]{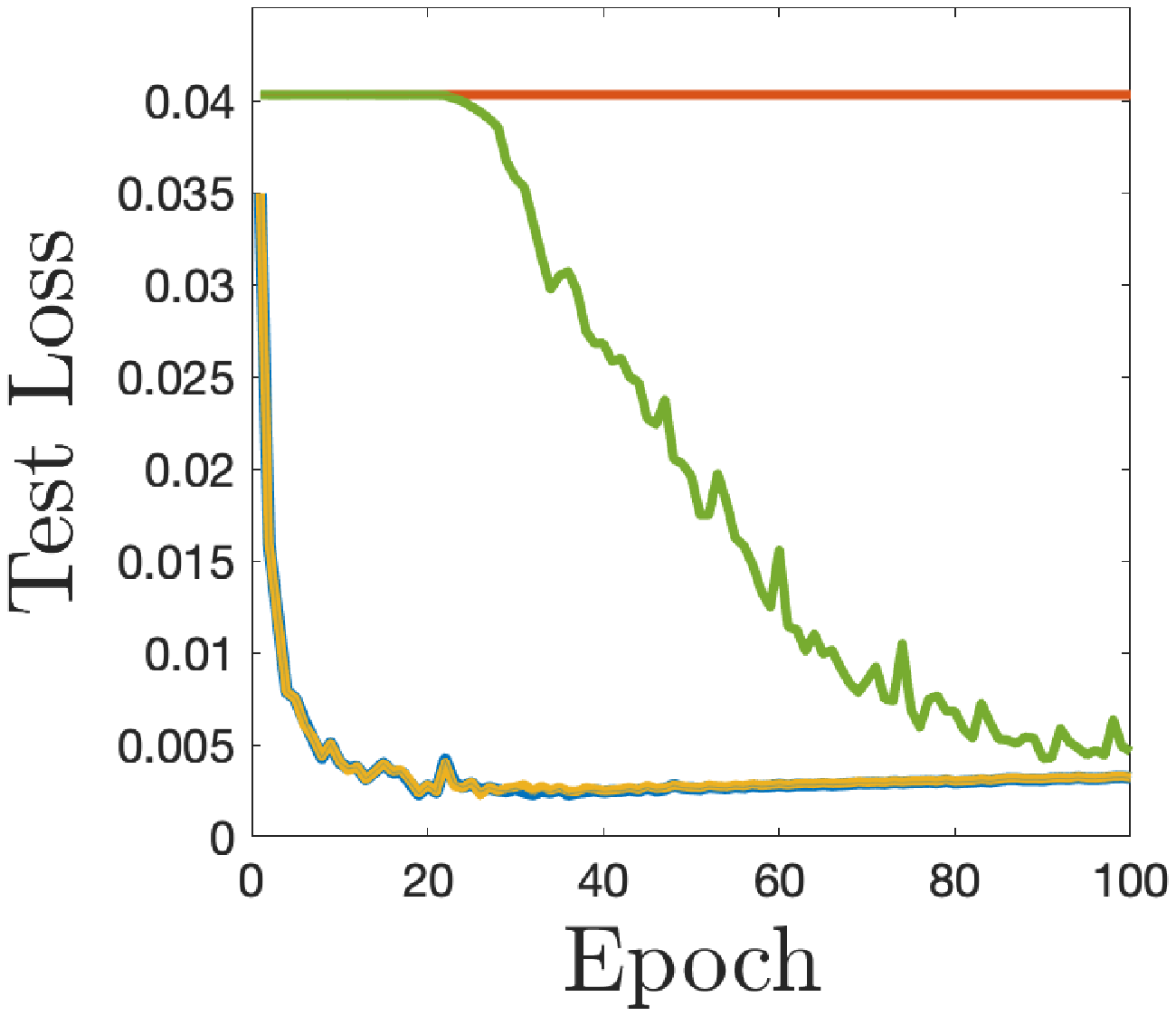}  
  \end{minipage}
  % include first image
  \begin{minipage}{.48\textwidth}
  \centering
  % include first image
  \includegraphics[width=1\linewidth, trim=1cm 0cm 0cm 1cm]{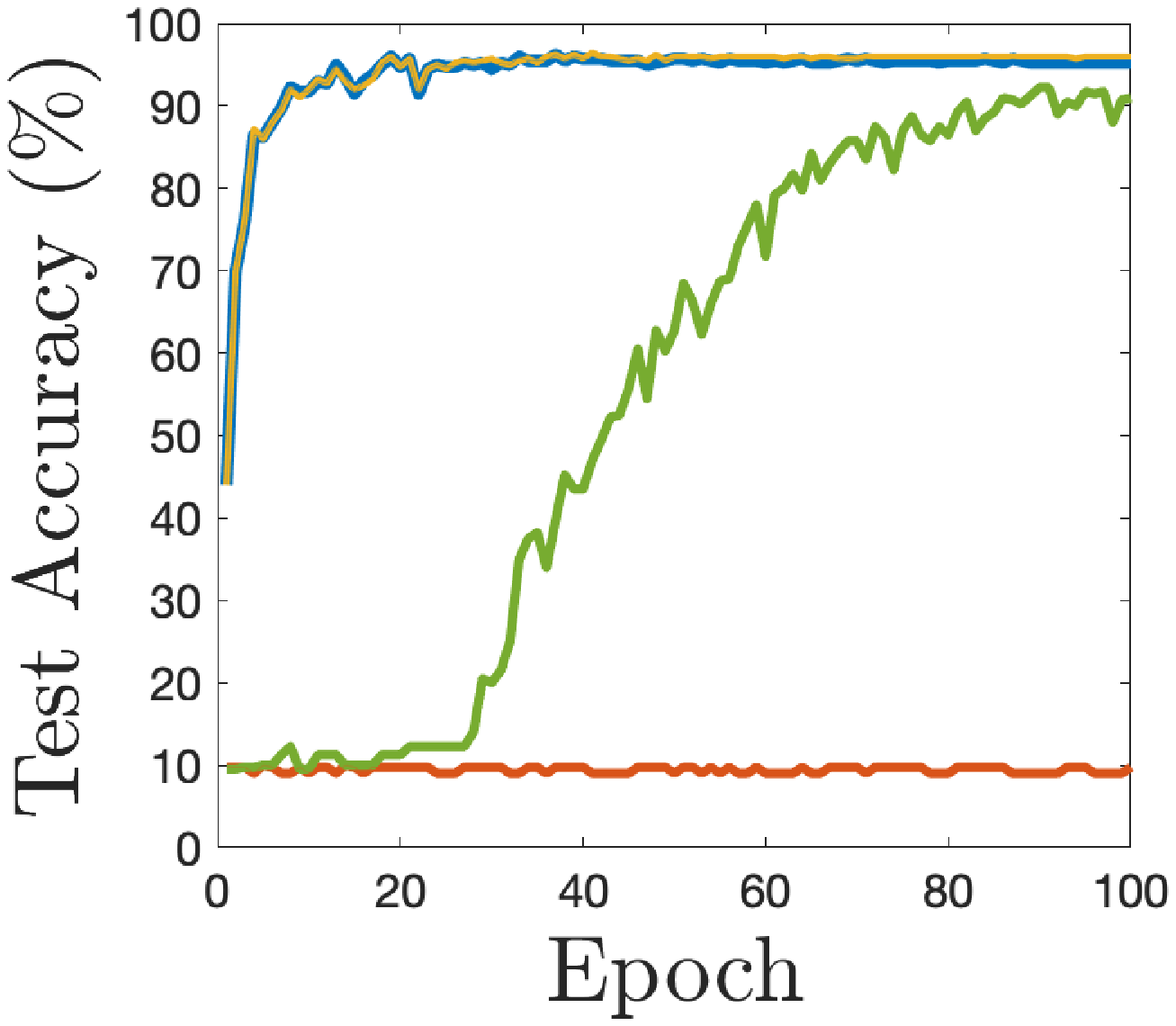}  
%   \label{fig:multi-task}
\end{minipage}
\caption{Group 1}
\label{fig: MNist test 8 Byzantine agents}
\end{subfigure}
\begin{subfigure}{0.495\textwidth}
  \centering
  \begin{minipage}{.48\textwidth}
  \centering
    \includegraphics[width=1\linewidth, trim=1cm 0cm 0cm 1cm]{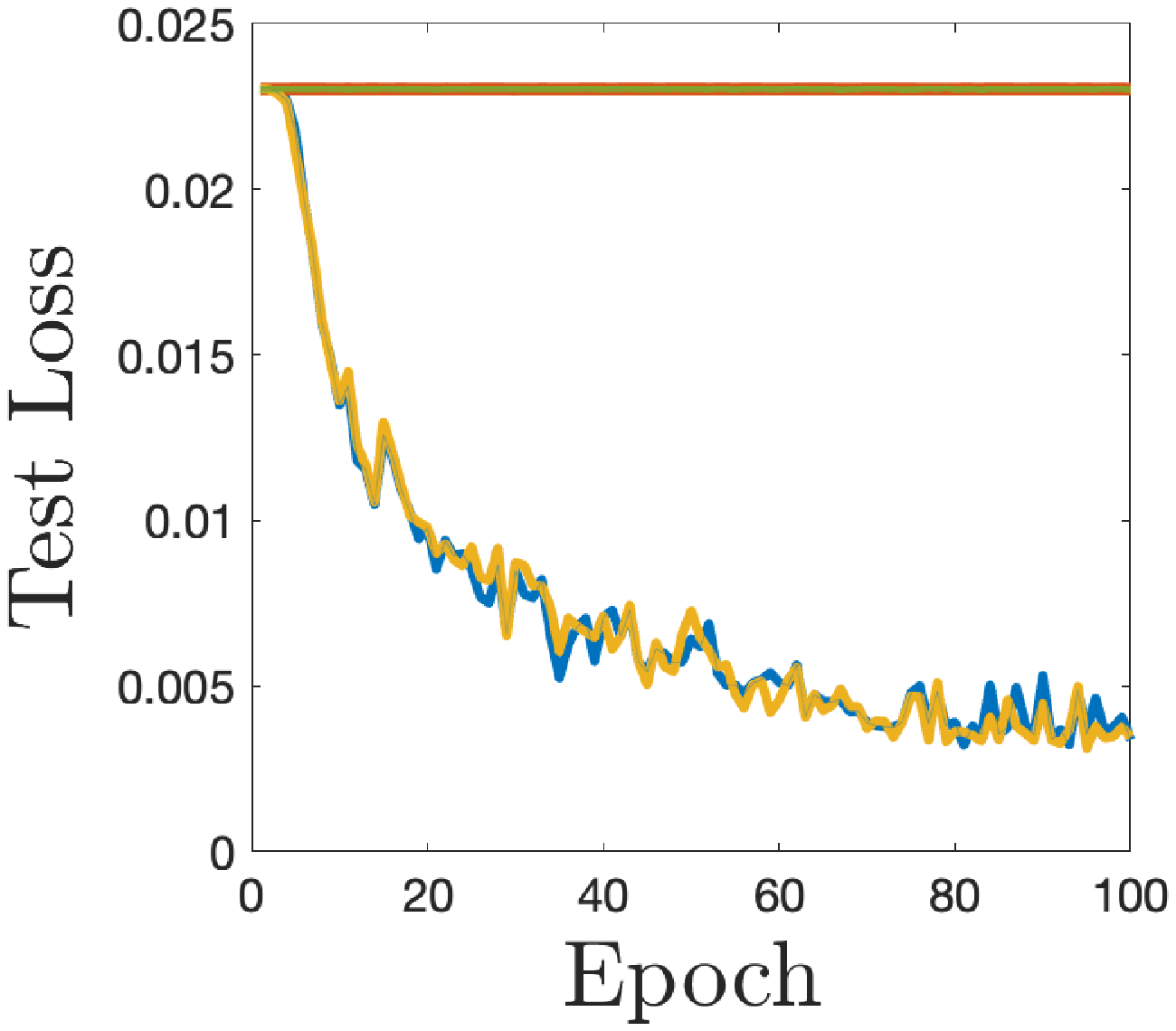}  
  \end{minipage}
  % include first image
  \begin{minipage}{.48\textwidth}
  \centering
  % include first image
  \includegraphics[width=1\linewidth, trim=1cm 0cm 0cm 1cm]{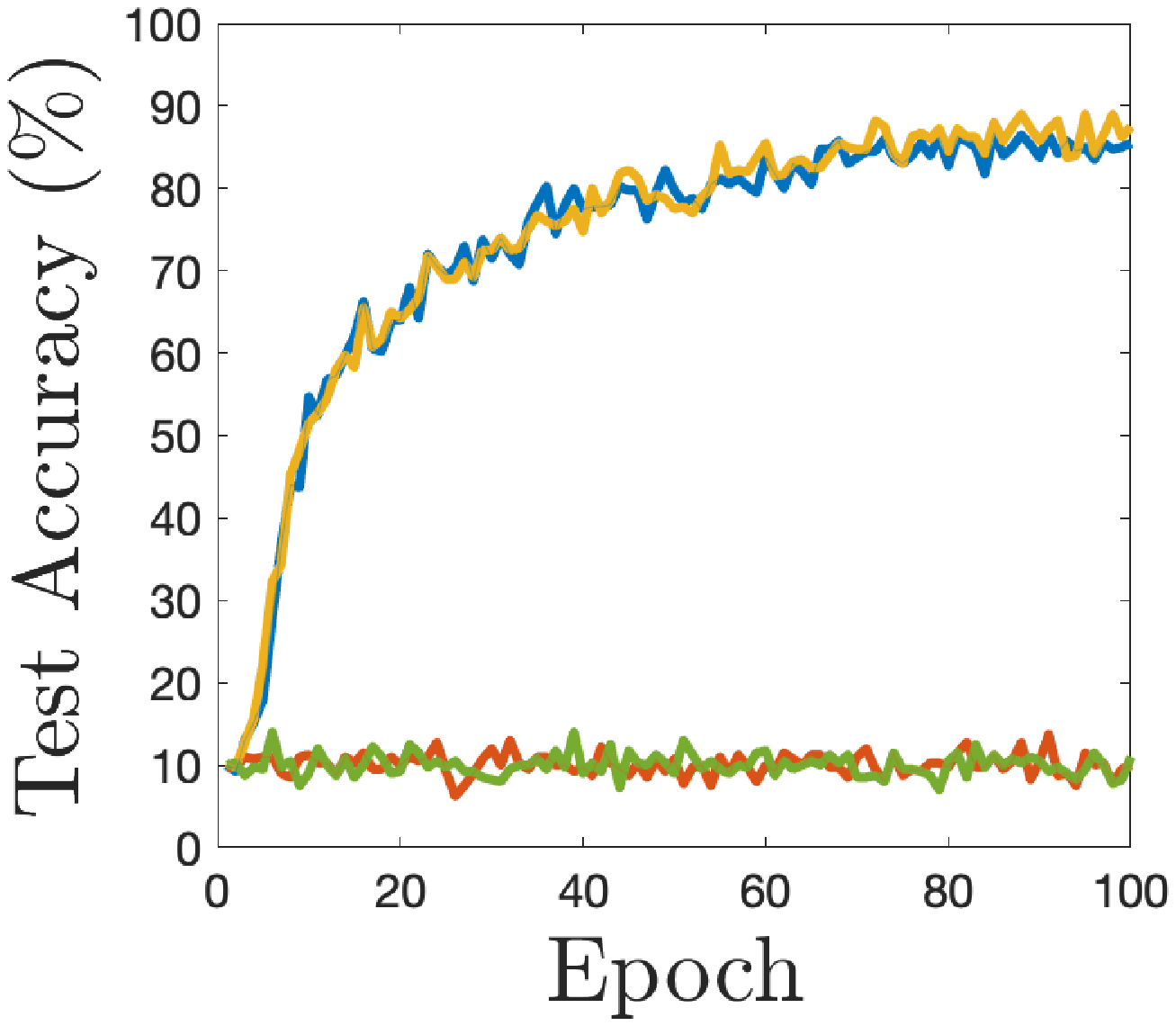}  
%   \label{fig:multi-task}
\end{minipage}
\caption{Group 2}
\label{fig: synthetic test 8 Byzantine agents}
\end{subfigure}
\caption{Digit Classification: average testing loss and  accuracy for normal agents, with 8 Byzantine agents (four for each group).}
\end{figure}

Since the performance of agents in the two groups diverges, we plot the results separately for the two groups.
\figref{fig: MNist test 8 Byzantine agents} and \figref{fig: synthetic test 8 Byzantine agents} show the average \emph{testing} loss and classification accuracy of the normal agents in group 1 and group 2,
when 8 out of 10  agents (four for each group) are Byzantine (the only normal agent in each group has access to 2000 training data). 

\begin{figure}[H]
\centering
\begin{subfigure}{0.495\textwidth}
  \centering
  \begin{minipage}{.48\textwidth}
  \centering
    \includegraphics[width=1\linewidth, trim=1cm 0cm 0cm 1cm]{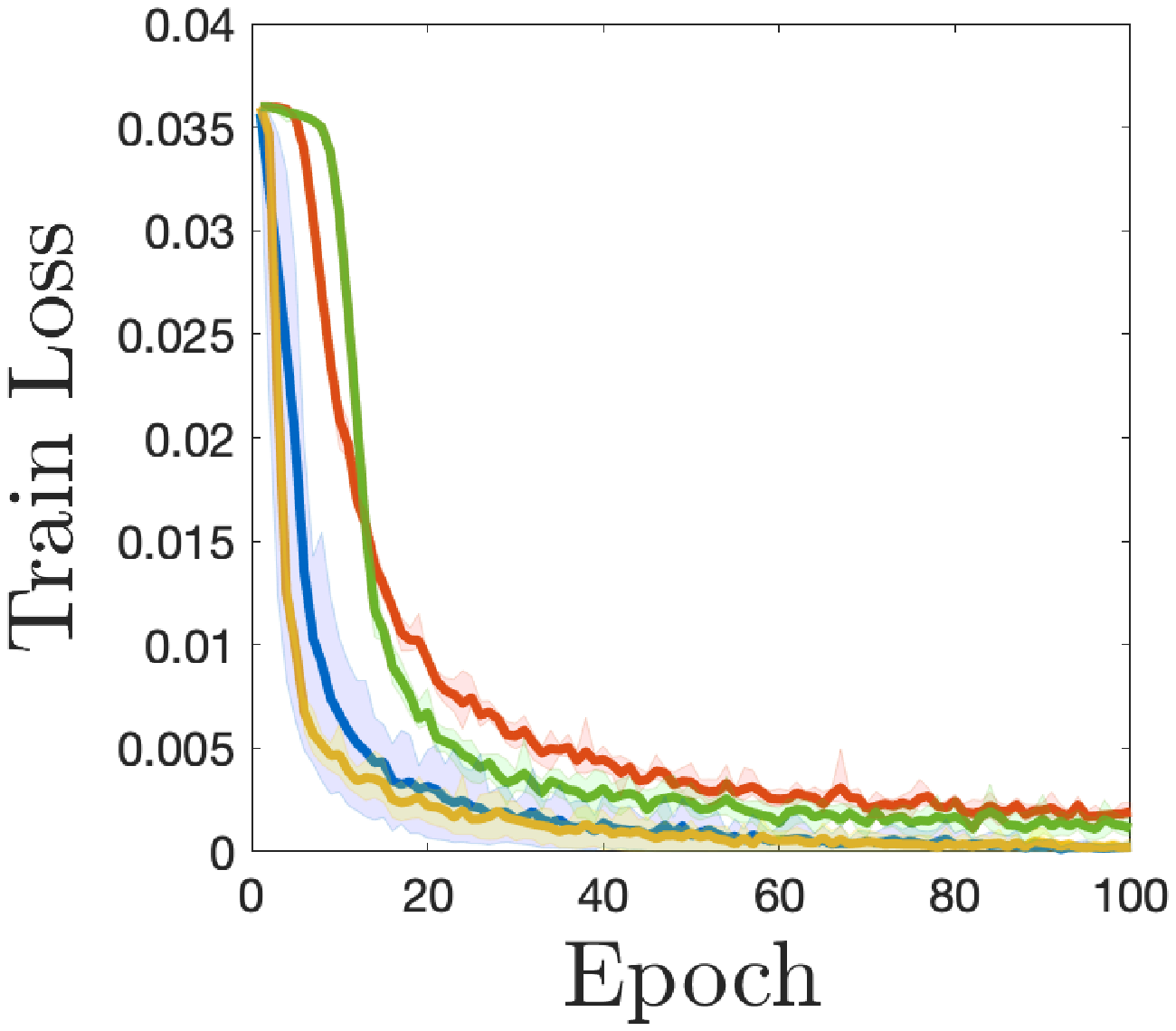}  
  \end{minipage}
  % include first image
  \begin{minipage}{.48\textwidth}
  \centering
  % include first image
  \includegraphics[width=1\linewidth, trim=1cm 0cm 0cm 1cm]{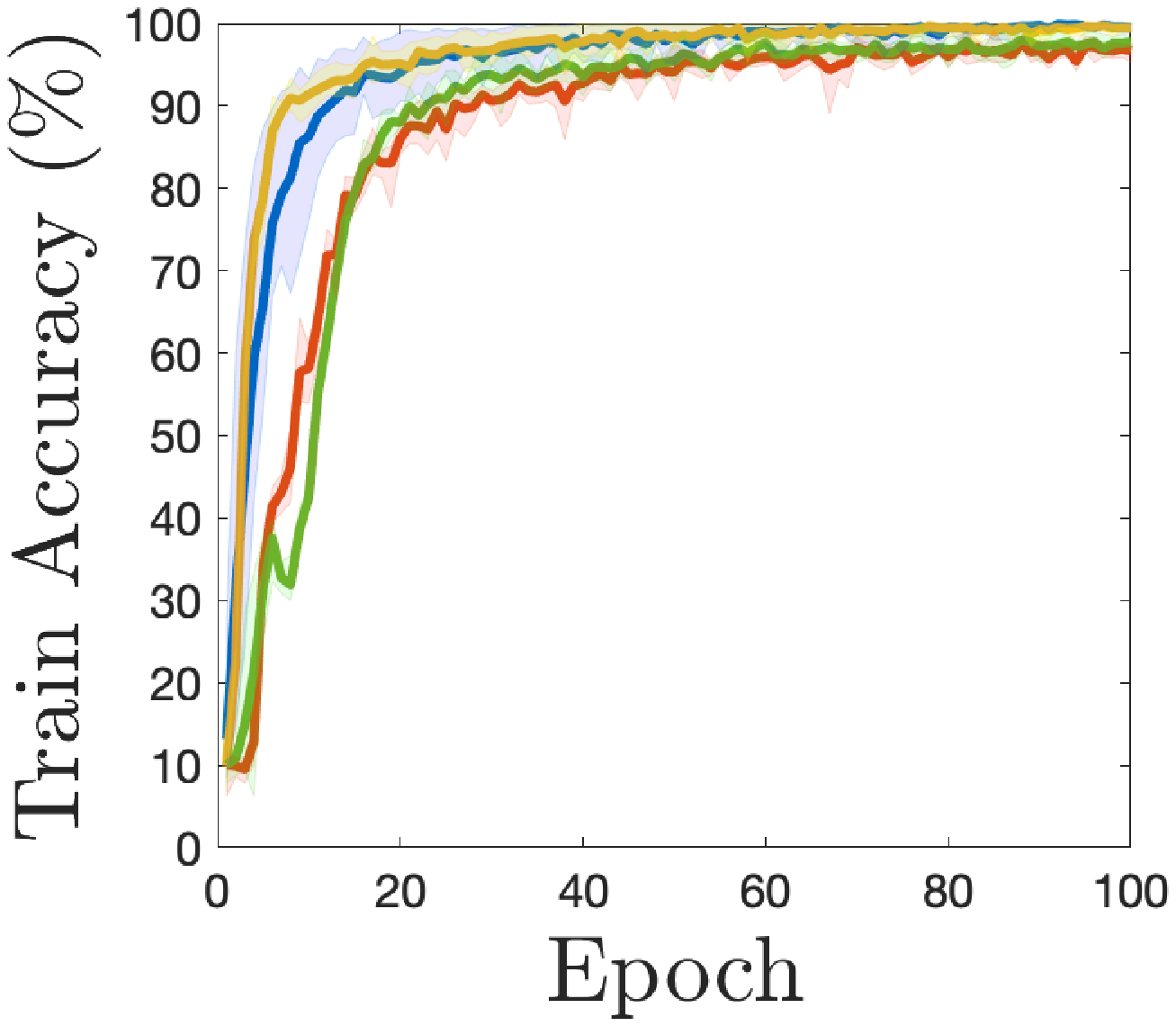}  
%   \label{fig:multi-task}
\end{minipage}
\caption{No attack}
\end{subfigure}
\begin{subfigure}{0.495\textwidth}
  \centering
  \begin{minipage}{.48\textwidth}
  \centering
    \includegraphics[width=1\linewidth, trim=1cm 0cm 0cm 1cm]{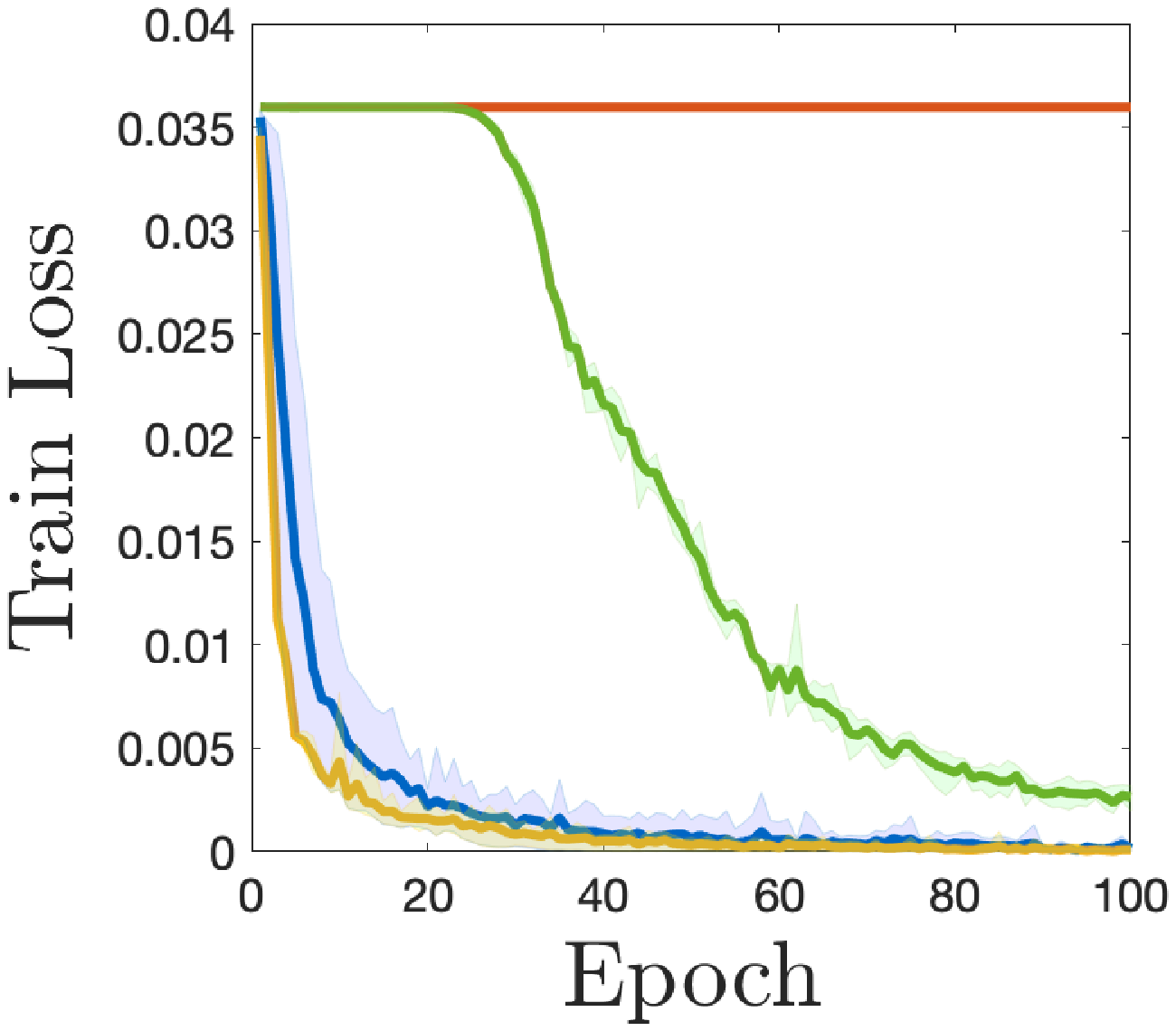}  
  \end{minipage}
  % include first image
  \begin{minipage}{.48\textwidth}
  \centering
  % include first image
  \includegraphics[width=1\linewidth, trim=1cm 0cm 0cm 1cm]{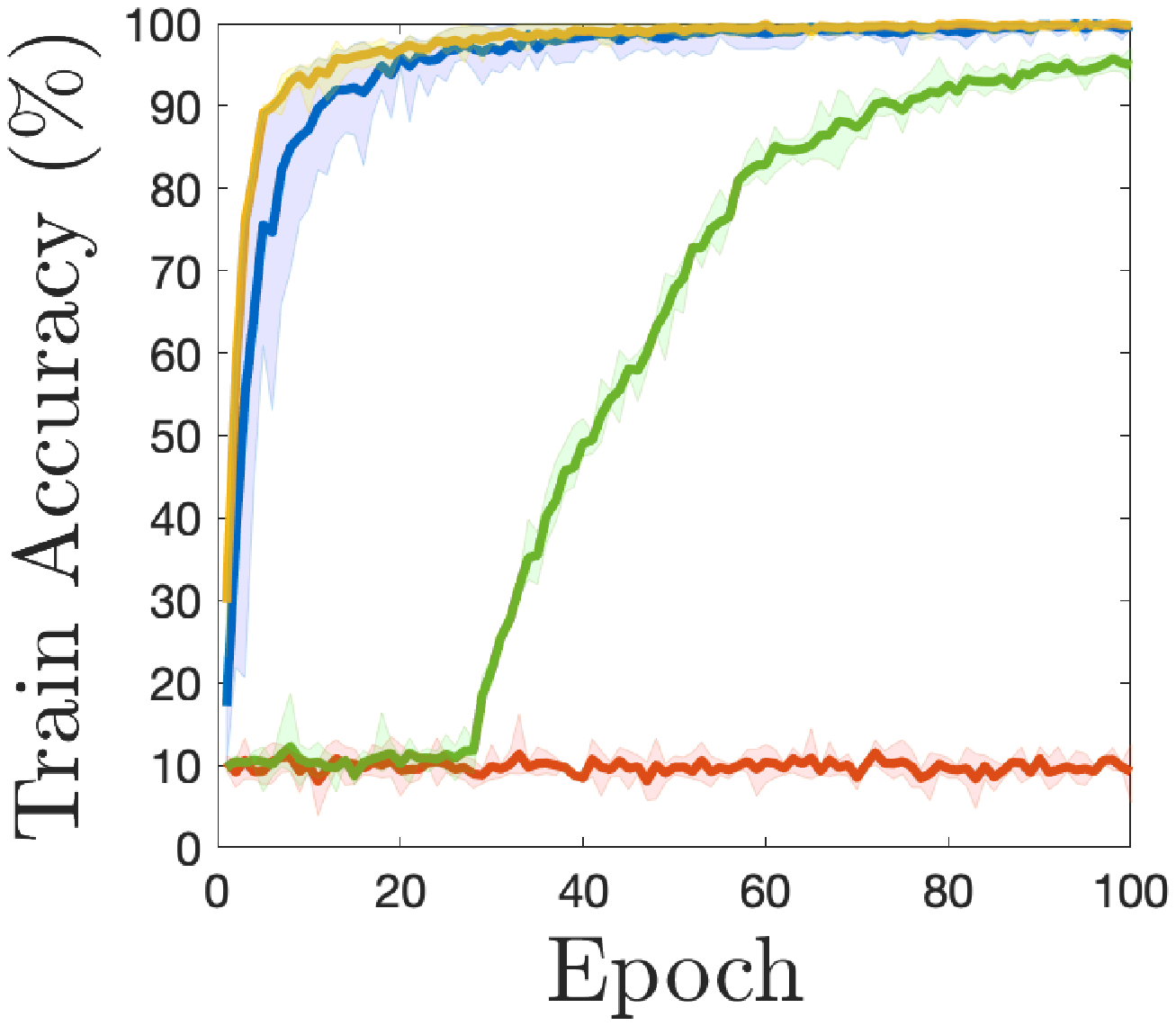}  
%   \label{fig:multi-task}
\end{minipage}
\caption{2 Byzantine agents}
\end{subfigure}
\caption{Digit Classification: average training loss and  accuracy for normal agents in group 1.}
\label{fig: MNist train}
\end{figure}

\begin{figure}[H]
\centering
% \begin{minipage}{1\textwidth}
% \centering
%     \includegraphics[width=0.7\linewidth, trim=0cm -0.4cm -1cm 0cm]{figure/TL_legend_new.jpg}
% \end{minipage}\\
% \vspace{0.1cm}
\begin{subfigure}{0.495\textwidth}
  \centering
  \begin{minipage}{.48\textwidth}
  \centering
    \includegraphics[width=1\linewidth, trim=1cm 0cm 0cm 1cm]{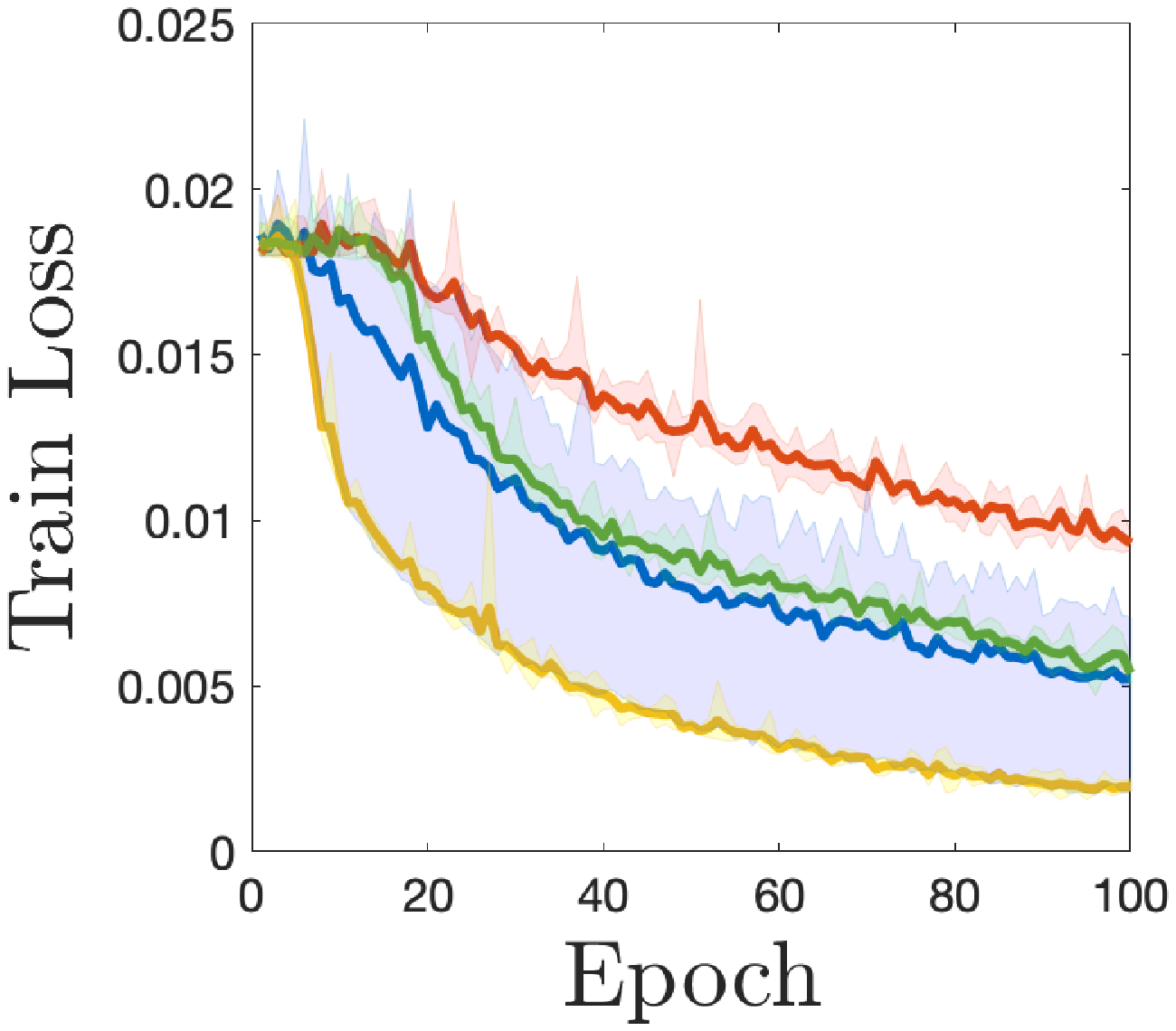}  
  \end{minipage}
  % include first image
  \begin{minipage}{.48\textwidth}
  \centering
  % include first image
  \includegraphics[width=1\linewidth, trim=1cm 0cm 0cm 1cm]{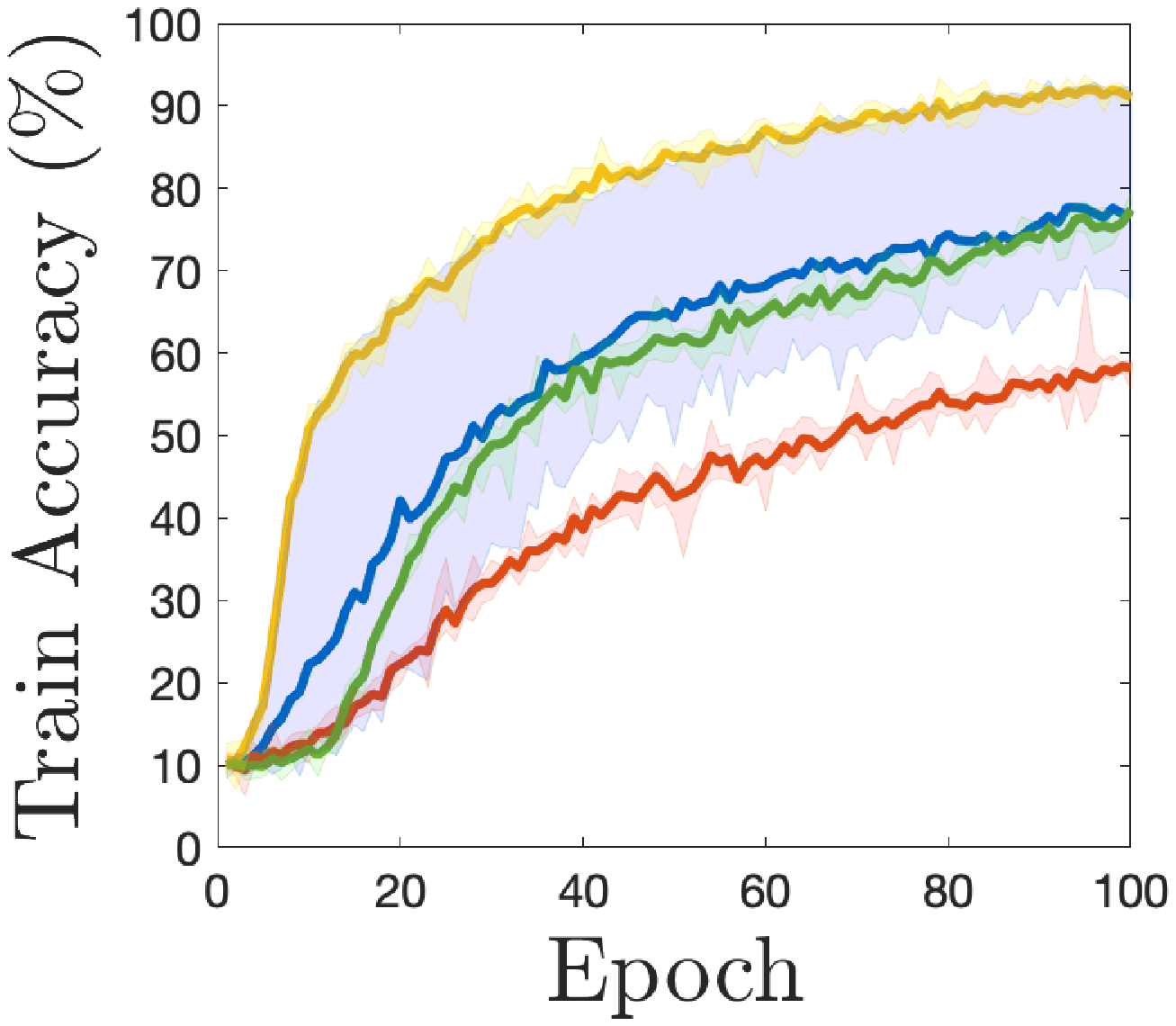}  
%   \label{fig:multi-task}
\end{minipage}
\caption{No attack}
\end{subfigure}
\begin{subfigure}{0.495\textwidth}
  \centering
  \begin{minipage}{.48\textwidth}
  \centering
    \includegraphics[width=1\linewidth, trim=1cm 0cm 0cm 1cm]{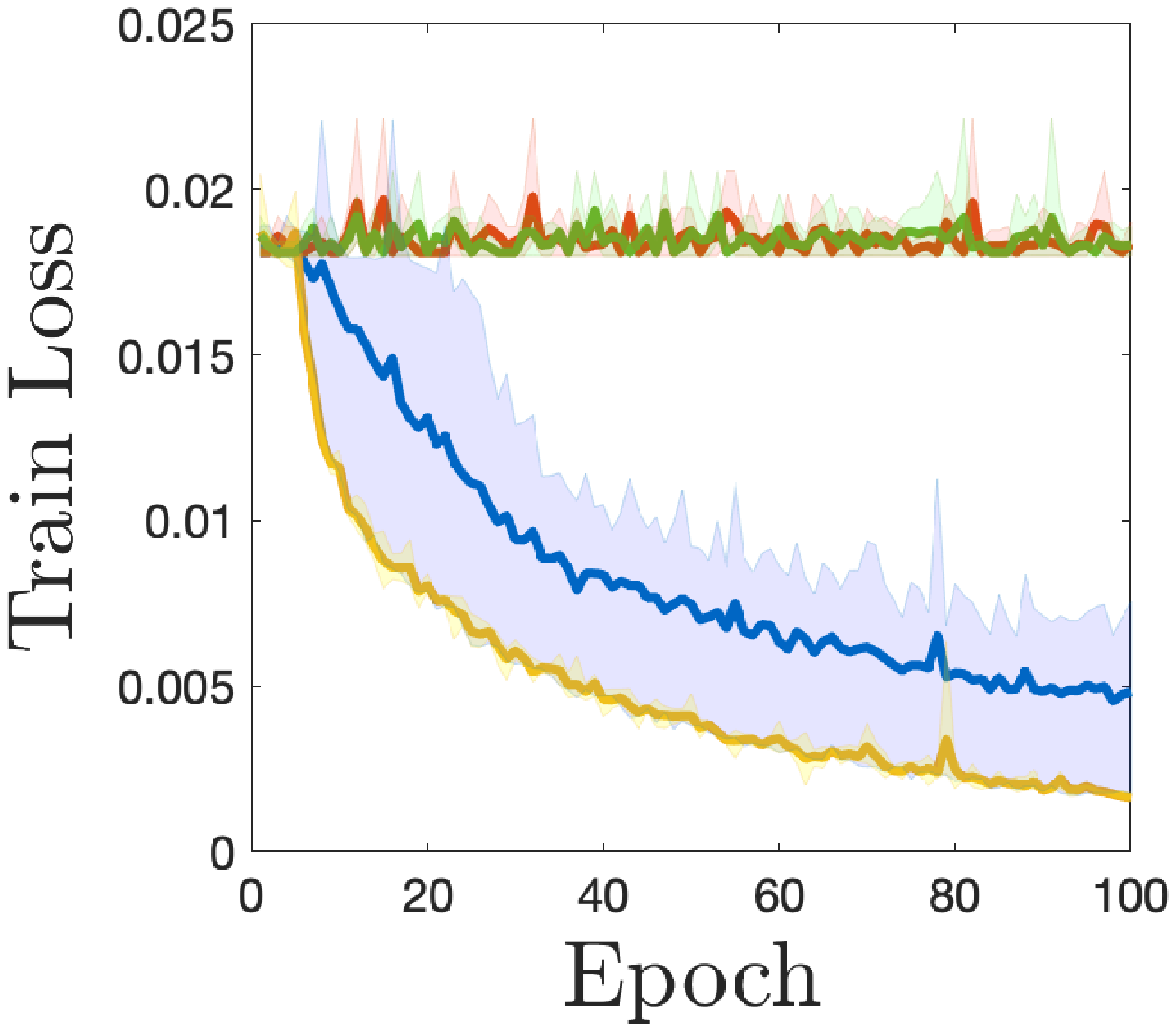}  
  \end{minipage}
  % include first image
  \begin{minipage}{.48\textwidth}
  \centering
  % include first image
  \includegraphics[width=1\linewidth, trim=1cm 0cm 0cm 1cm]{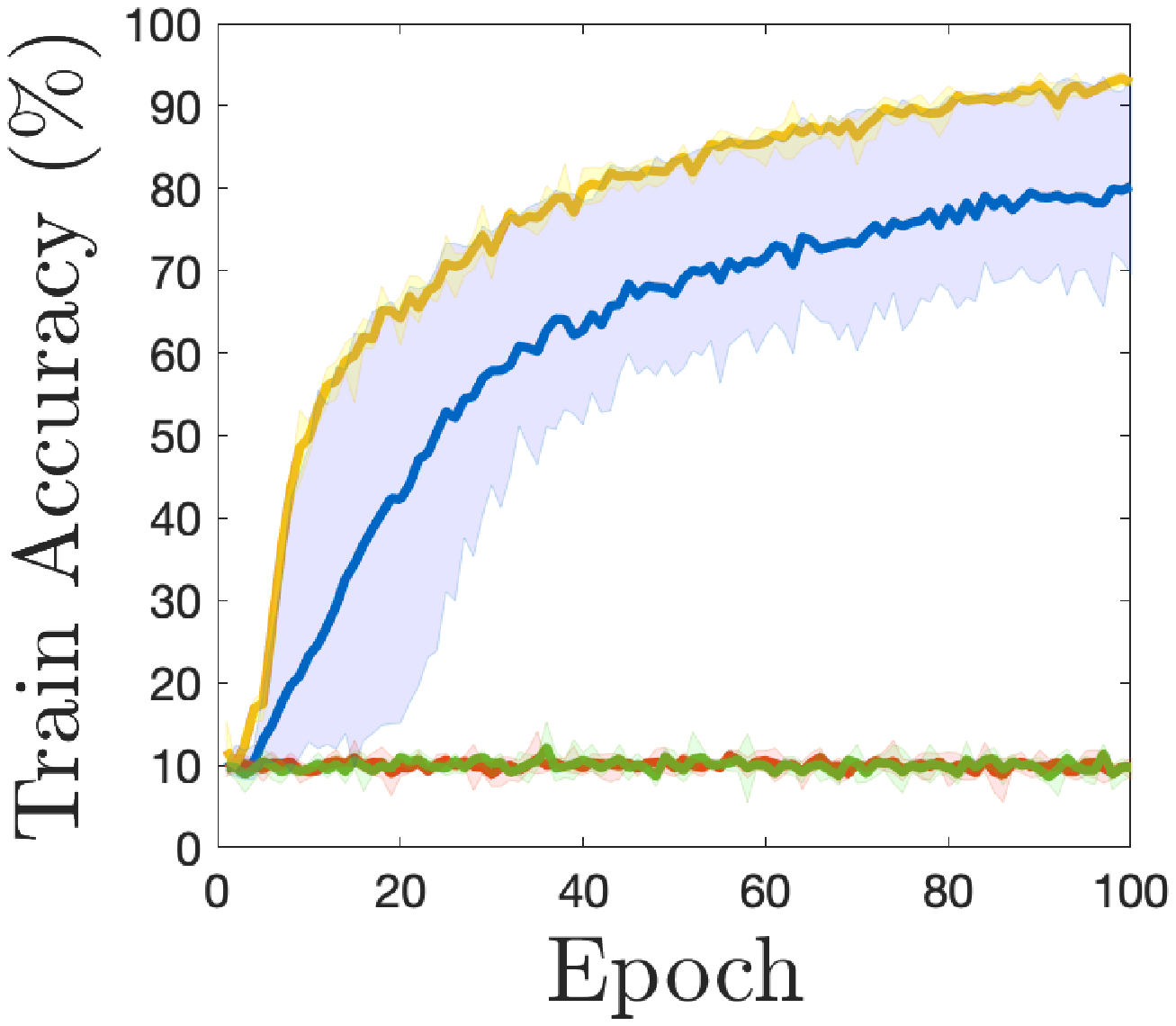}  
%   \label{fig:multi-task}
\end{minipage}
\caption{2 Byzantine agents}
\end{subfigure}
% \begin{subfigure}{0.325\textwidth}
%   \centering
%   \begin{minipage}{.48\textwidth}
%   \centering
%     \includegraphics[width=1\linewidth, trim=1cm 0cm 0cm 1cm]{figure/synthetic_range_loss_train_attack_8.eps}  
%   \end{minipage}
%   % include first image
%   \begin{minipage}{.48\textwidth}
%   \centering
%   % include first image
%   \includegraphics[width=1\linewidth, trim=1cm 0cm 0cm 1cm]{figure/synthetic_range_acc_train_attack_8.eps}  
% %   \label{fig:multi-task}
% \end{minipage}
% \caption{8 Byzantine agents}
% \end{subfigure}
\caption{Digit Classification: average training loss and  accuracy for normal agents in group 2.}
\label{fig: synthetic train}
\end{figure}

\begin{figure}[H]
\centering
% \begin{minipage}{1\textwidth}
% \centering
%     \includegraphics[width=0.7\linewidth, trim=0cm -0.4cm -1cm 0cm]{figure/TL_legend_new.jpg}
% \end{minipage}\\
% \vspace{0.1cm}
\begin{subfigure}{0.495\textwidth}
  \centering
  \begin{minipage}{.48\textwidth}
  \centering
    \includegraphics[width=1\linewidth, trim=1cm 0cm 0cm 1cm]{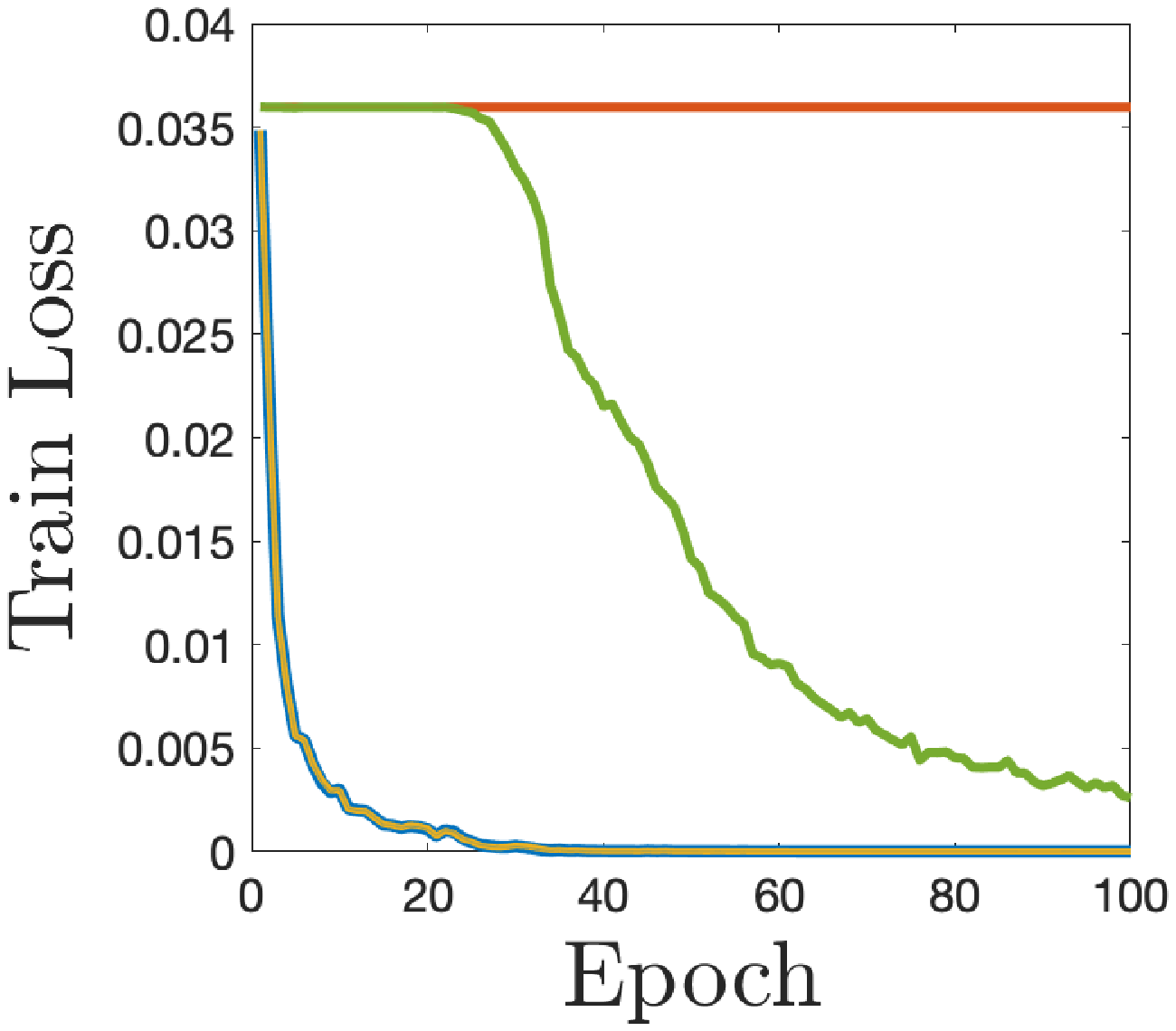}  
  \end{minipage}
  % include first image
  \begin{minipage}{.48\textwidth}
  \centering
  % include first image
  \includegraphics[width=1\linewidth, trim=1cm 0cm 0cm 1cm]{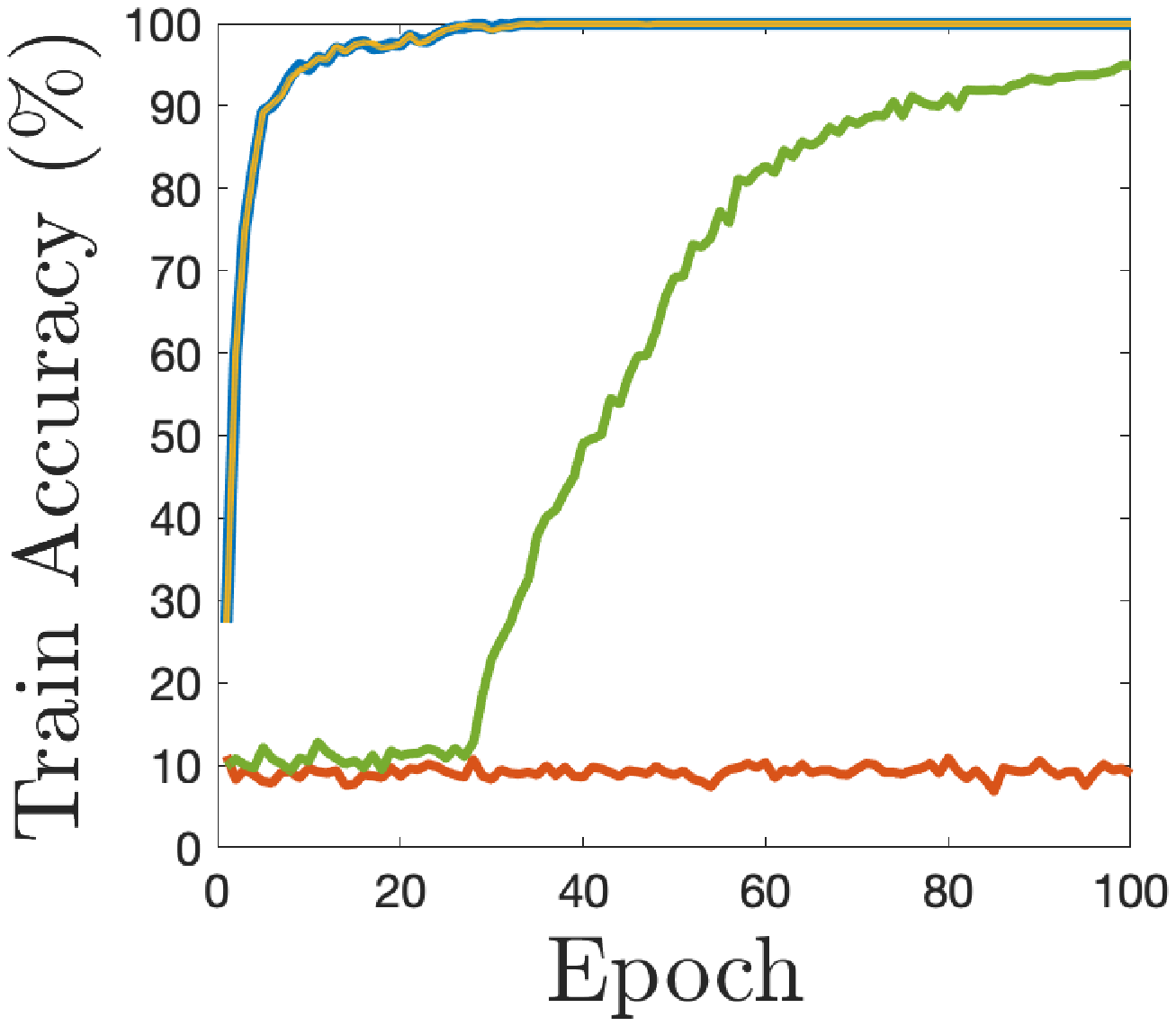}  
%   \label{fig:multi-task}
\end{minipage}
\caption{Group 1}
\label{fig: MNist train 8 Byzantine agents}
\end{subfigure}
\begin{subfigure}{0.495\textwidth}
  \centering
  \begin{minipage}{.48\textwidth}
  \centering
    \includegraphics[width=1\linewidth, trim=1cm 0cm 0cm 1cm]{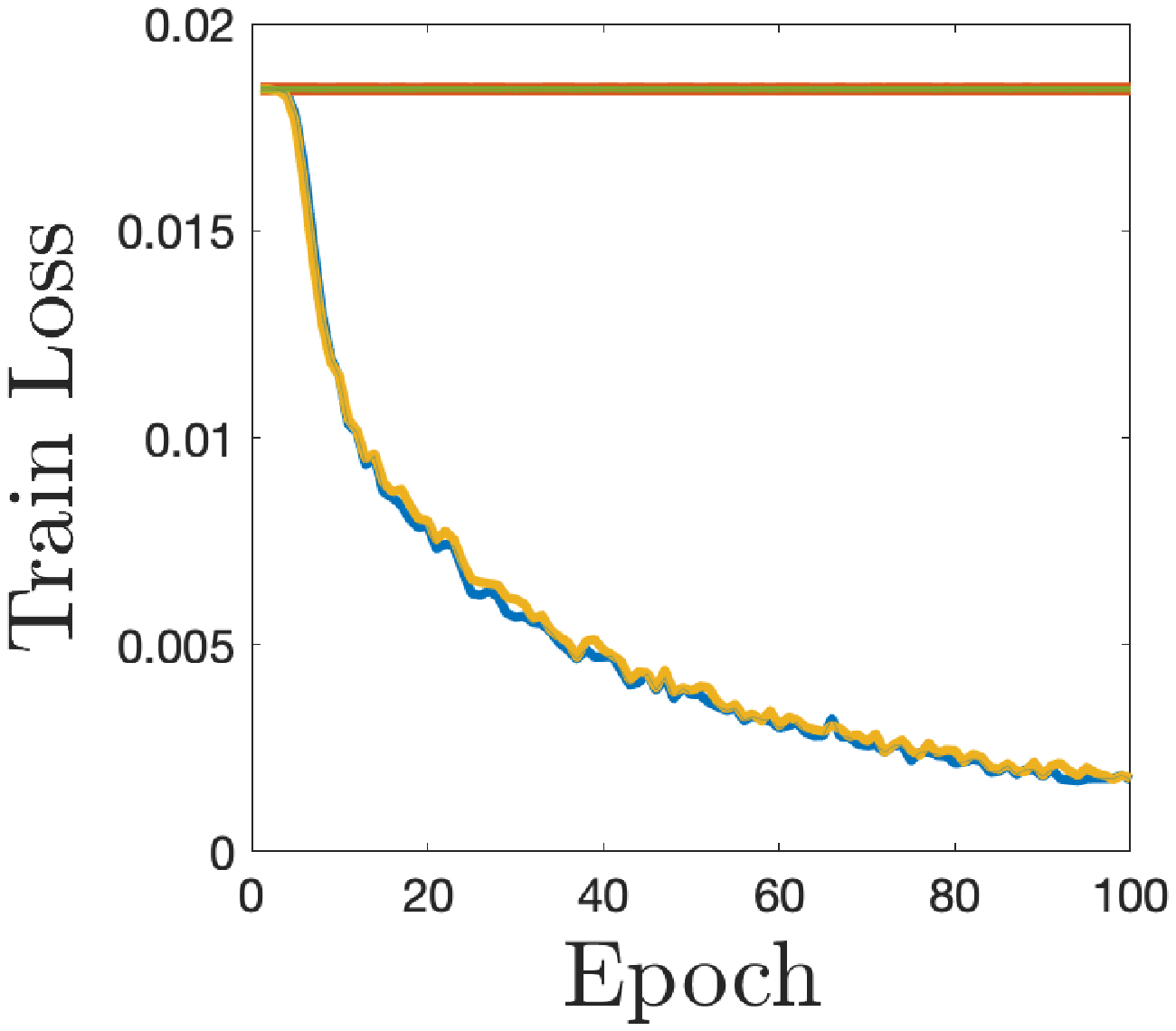}  
  \end{minipage}
  % include first image
  \begin{minipage}{.48\textwidth}
  \centering
  % include first image
  \includegraphics[width=1\linewidth, trim=1cm 0cm 0cm 1cm]{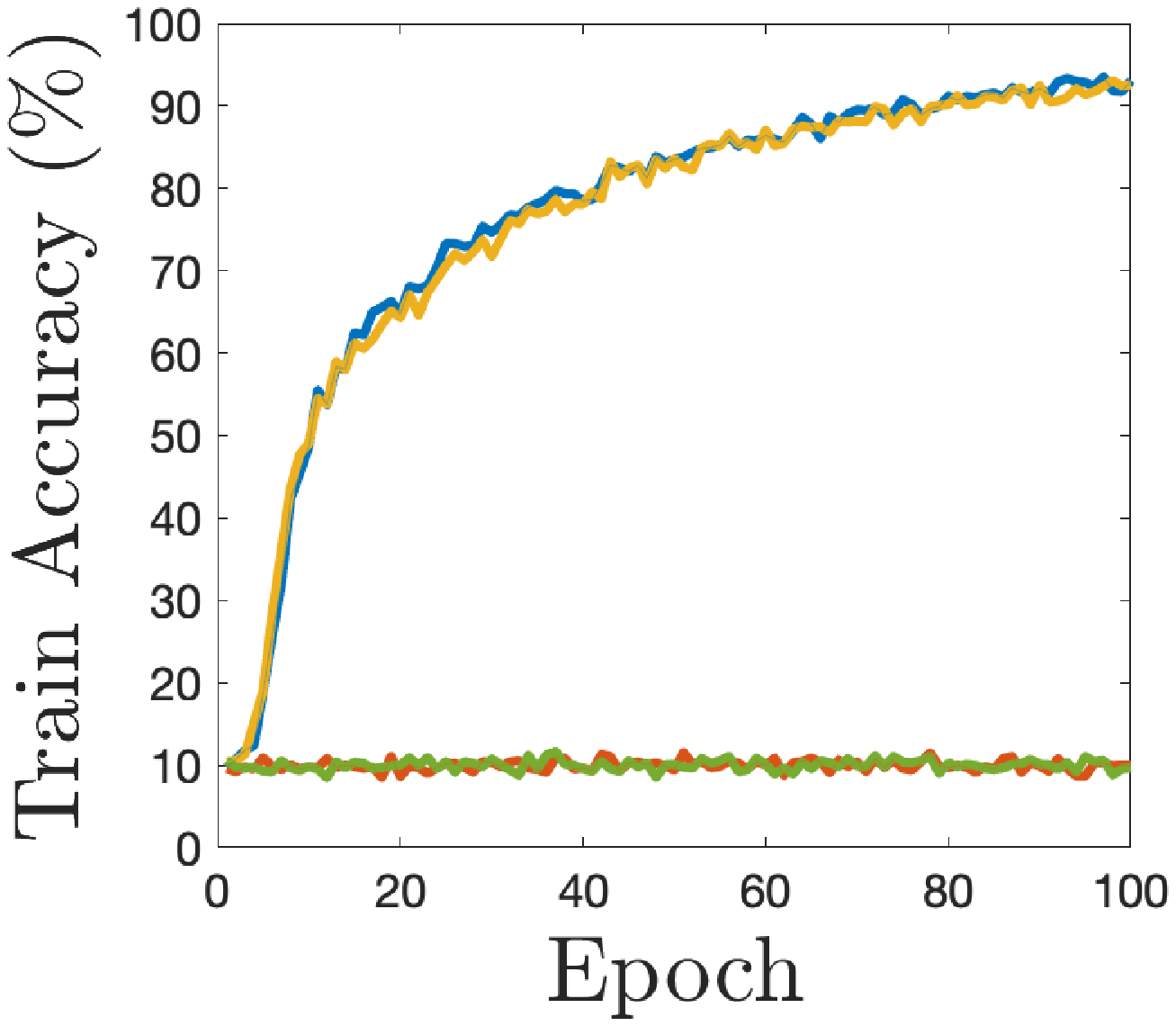}  
%   \label{fig:multi-task}
\end{minipage}
\caption{Group 2}
\label{fig: synthetic train 8 Byzantine agents}
\end{subfigure}
\caption{Digit Classification: average training loss and  accuracy for normal agents, with 8 Byzantine agents (four for each group).}
\end{figure}

\figref{fig: MNist train} and \figref{fig: MNist train 8 Byzantine agents} show the mean and range of the average \emph{training} loss and classification accuracy of the normal agents in group 1, in the case of no attack, with 2 Byzantine agents, and with 8 Byzantine agents, \textcolor{red}{which are selected randomly}.
\figref{fig: synthetic train} and \figref{fig: synthetic train 8 Byzantine agents} show the mean and range of the average \emph{training} loss and classification accuracy of the normal agents in group 2, in the case of no attack, with 2 Byzantine agents, and with 8 Byzantine agents \textcolor{red}{(again selected randomly)}.
In all the examples, for both training and testing, we observe that the loss-based weight assignment rule \eqref{eq: filtering weight} outperforms the other rules as well as the non-cooperative case, with respect to the mean and range of the average loss and accuracy, thereby validating the result indicated by \eqref{eq: resilient convergence  with improved learning error}.
Even in the extreme case in which there is only one normal agent in each group and all of the other agents are Byzantine, the loss-based weight assignment rule \eqref{eq: filtering weight} has the same performance as the  non-cooperative case, showing its resilience to an arbitrary number of Byzantine agents.

% It is obvious from the results that agents in group 1 have  better learning performance than agents in group 2 since the synthetic digit classification is more challenging than the MNIST digit classification.
% Yet using the loss-based weights, agents are able to correctly identify the neighbors estimating the same task and use the cooperation to benefit their own learning performance. 
% The loss-based weights still outperform the other combination rules in all the cases, with respect to the mean and range of the average loss and accuracy. And even when 8 of the 10 agents are Byzantine (only one normal agent for each group), the loss-based weights can still resiliently converge and performs close to the non-cooperative case.

\textcolor{red}{Comparing} the results between \textcolor{red}{groups} 1 and 2 reveals that cooperation is most beneficial when there is a substantial divergence in agents' learning performances.
% It can also be observed by a comparison between the results of group 1 and  group 2 
% that the cooperation greatly 
% improves the overall learning performance when  some of the agents do not have enough data to build refined models.
Given limited training data, agents in group 1 are able to build refined models.
\textcolor{red}{It} is harder for agents receiving less training data in group 2 to achieve a high learning performance \textcolor{red}{as} the synthetic digit classification is a more  challenging task than the MNIST digit classification.
Using the weight assignment rule \eqref{eq: filtering weight}, those agents receiving less data (and therefore, struggling to learn a good model), are
able to benefit from the cooperation with the neighbors having learned a refined model. 
At the same time, agents exhibiting high learning performance will not be negatively affected by such cooperation.

% \subsection{Examples of the dataset in Section \ref{sec: Digit classification}}

% \subsection{CNN architecture of Section \ref{sec: Digit classification}}

\begin{table}[H]%
% \scriptsize
\small
\caption{CNN architecture of Digit Classification}
\label{tab: CNN architecture of Section 6.3}
\vspace{0.3cm}
\centering %
\begin{tabular}{ R{2.2cm} R{3cm} R{2cm} }
\toprule 
 Layer (type)   &   Output Shape & Param \# \\
    \midrule
            Conv2d-1   &        [-1, 32, 28, 28]     &         320\\
              ReLU-2     &       [-1, 32, 28, 28]     &           0\\
         MaxPool2d-3     &       [-1, 32, 14, 14]     &           0\\
            Conv2d-4     &       [-1, 64, 14, 14]     &      18,496\\
              ReLU-5    &        [-1, 64, 14, 14]     &           0\\
         MaxPool2d-6    &          [-1, 64, 7, 7]    &            0\\
            Conv2d-7    &          [-1, 64, 7, 7]    &       36,928\\
              ReLU-8    &          [-1, 64, 7, 7]    &            0\\
         MaxPool2d-9     &         [-1, 64, 3, 3]    &            0\\
           Linear-10     &              [-1, 128]    &       73,856\\
             ReLU-11     &              [-1, 128]    &            0\\
           Linear-12     &               [-1, 10]     &       1,290\\
    \bottomrule
\end{tabular}
\end{table}

\vspace{-0.3cm}

\vspace{-0.3cm}

% \subsection{Supplementary simulation results of Section \ref{sec: Digit classification}}

\vspace{-0.3cm}

\vspace{-0.3cm}

\vspace{-0.3cm}

